\documentclass[runningheads]{llncs}
\usepackage{graphicx}

\usepackage{tikz}
\usepackage{comment}
\usepackage{amsmath,amssymb} 
\usepackage{color}

\usepackage{xcolor}
\usepackage{booktabs}
\usepackage{amsfonts}
\usepackage{adjustbox}
\usepackage{dashrule}
\usepackage{mathtools}
\usepackage{makecell}
\usepackage{colortbl}
\usepackage{array}
\usepackage{times}
\usepackage{dsfont}
\usepackage{ctable}
\usepackage{cite}
\usepackage{wrapfig}
\usepackage{floatrow}
\usepackage{float}
\floatstyle{plaintop}
\restylefloat{table}

\definecolor{blueblack}{RGB}{0, 108, 173}
\definecolor{taborange}{RGB}{235, 127, 14}
\definecolor{tabgreen}{RGB}{30, 160, 30}
\definecolor{tabpurple}{RGB}{128, 103, 189}
\definecolor{tabblue}{RGB}{31, 119, 180}
\definecolor{tabred}{RGB}{214, 39, 40}
\definecolor{tabpink}{RGB}{227, 119, 194}
\definecolor{tabgray}{RGB}{127, 127, 127}

\usepackage{subcaption}
\usepackage[usestackEOL]{stackengine}
\usepackage[accsupp]{axessibility}  

\usepackage[pagebackref,breaklinks,colorlinks]{hyperref}

\usepackage[capitalize]{cleveref}

\newcommand{\ie}{\emph{i.e}.}
\newcommand{\eg}{\emph{e.g}.}

\newcommand{\etal}{\emph{et al}.}

\newcommand{\startcompact}[1]{\par\vspace{-1em}\begin{#1}%
\allowdisplaybreaks\ignorespaces}
\newcommand{\stopcompact}[1]{\end{#1}\ignorespaces}

\newcommand{\A}{\mathbf{A}}

\newcommand{\x}{\mathbf{x}}
\newcommand{\y}{\mathbf{y}}

\newcommand{\p}{\mathbf{p}}

\renewcommand{\b}{\mathbf{b}}

\newcommand{\X}{\mathbf{X}}

\newcommand{\W}{\mathbf{W}}

\newcommand{\D}{\mathbf{D}}

\newtheorem{prop}{Proposition}

\begin{document}
\pagestyle{headings}
\mainmatter
\def\ECCVSubNumber{7771}  

\title{Trading Positional Complexity vs. Deepness \\ in Coordinate Networks}

\titlerunning{Trading deepness vs. complexity}
%
\author{Jianqiao Zheng \thanks{Project page at \href{https://osiriszjq.github.io/complex_encoding}{https://osiriszjq.github.io/complex\_encoding}}  \and Sameera Ramasinghe \inst{*} \and
Xueqian Li\and
Simon Lucey}
\authorrunning{Zheng et al.}
\institute{Australian Institute for Machine Learning\\
University of Adelaide\\
\email{jianqiao.zheng@adelaide.edu.au}}

\maketitle

\begin{abstract}
It is well noted that coordinate-based MLPs benefit---in terms of preserving high-frequency information---through the encoding of coordinate positions as an array of Fourier features. 
Hitherto, the rationale for the effectiveness of these \emph{positional encodings} has been mainly studied through a Fourier lens. 
In this paper, we strive to broaden this understanding by showing that alternative non-Fourier embedding functions can indeed be used for positional encoding. Moreover, we show that their performance is entirely determined by a trade-off between the stable rank of the embedded matrix and the distance preservation between embedded coordinates. We further establish that the now ubiquitous Fourier feature mapping of position is a special case that fulfills these conditions.  Consequently, we present a more general theory to analyze positional encoding in terms of shifted basis functions. 
In addition, we argue that employing a more complex positional encoding---that scales exponentially with the number of modes---requires only a linear (rather than deep) coordinate function to achieve comparable performance. 
Counter-intuitively, we demonstrate that trading positional embedding complexity for network deepness is orders of magnitude faster than current state-of-the-art; despite the additional embedding complexity.
To this end, we develop the necessary theoretical formulae and empirically verify that our theoretical claims hold in practice.
\end{abstract}

\section{Introduction}
Positional encoding is an umbrella term used for representing the coordinates of a structured object as a finite-dimensional embedding. Such embeddings are fast becoming critical instruments in modern language models~\cite{gehring2017convolutional, vaswani2017attention, devlin2018bert, lewis2019bart, yang2019xlnet, bao2020unilmv2} and vision tasks that involve encoding a signal (\eg, 2D image, 3D object, etc.) as weights of a neural network~\cite{mildenhall2020nerf, zhong2019reconstructing, park2020deformable,li2020neural, ost2020neural, gafni2020dynamic, martin2020nerf, barron2021mip}. Of specific interest in this paper is the use of positional encodings when being used to enhance the performance of~\emph{coordinate-MLPs}. Coordinate-MLPs are fully connected networks, trained to learn the structure of an object as a continuous function, with coordinates as inputs. However, the major drawback of training coordinate-MLPs with raw input coordinates is their sub-optimal performance in learning high-frequency content~\cite{rahaman2019spectral}.

As a remedy, recent studies empirically confirmed that projecting the coordinates to a higher dimensional space using sine and cosine functions of different frequencies (\ie, Fourier frequency mapping) allows coordinate-MLPs to learn high-frequency information more effectively~\cite{mildenhall2020nerf, zhong2019reconstructing}. This observation was recently characterized theoretically by Tancik~\etal~\cite{tancik2020fourier}, showing that the above projection permits tuning the spectrum of the neural tangent kernel (NTK) of the corresponding MLP, thereby enabling the network to learn high-frequency information. Despite impressive empirical results, encoding position through Fourier frequency mapping entails some unenviable attributes. First, prior research substantiates the belief that the performance of the Fourier feature mapping is sensitive to the choice of frequencies. Leading methods for frequency selection, however, employ a stochastic strategy (\ie, random sampling) which can become volatile as one attempts to keep to a minimum the number of sampled frequencies. Second, viewing positional encoding solely through a Fourier lens obfuscates some of the fundamental principles behind its effectiveness. These concerns have heightened the need for an extended analysis of positional encoding.

This paper aims to overcome the aforesaid limitations by developing an alternative and more comprehensive understanding of positional encoding. The foremost benefit of our work is allowing non-Fourier embedding functions to be used in the positional encoding. Specifically, we show that positional encoding can be accomplished via systematic sampling of shifted continuous basis functions, where the shifts are determined by the coordinate positions. In comparison to the ambiguous frequency sampling in Fourier feature mapping, we derive a more interpretable relationship between the sampling density and the behavior of the embedding scheme. In particular, we discover that the effectiveness of the proposed embedding scheme primarily relies on two factors: (i) the approximate matrix rank of the embedded representation across positions, and (ii) the distance preservation between the embedded coordinates. Distance preservation measures the extent to which the inner product between the shifted functions correlates with the Euclidean distance between the corresponding coordinates. Intuitively, a higher approximate matrix rank causes better memorization of the training data, while the distance preservation correlates with generalization. Remarkably, we establish that any given continuous function can be used for positional encoding---as performance is simply determined by the trade-off between the aforementioned two factors. Further, we assert that the effectiveness and shortcomings of Fourier feature mapping can also be analyzed in the context of this newly developed framework. 
We also propose a complex positional encoding to relax the expressibility of the coordinate network into a single linear layer, which largely speedups the instance-based optimization.
An essential idea here is the separation of the coordinates. For a simple 1D signal, the input is only embedded in one direction. As for 2D natural images and 3D video sequences, the coordinates are still separable, which enables us to use the Kronecker product to gather input embedding in every single direction.
With signals that have non-separable coordinates, we add a blending matrix to linearly interpolate to get the final embedding.
In summary, the contribution of this paper is four-fold:

\begin{itemize}
    \item We expand the current understanding of positional encoding and show that it can be formulated as a systematic sampling scheme of shifted continuous basis functions. Compared to the popular Fourier frequency mapping, our formulation is more interpretative in nature and less restrictive. 
    \item We develop theoretical formulae to show that the performance of the encoding is governed by the approximate rank of the embedding matrix (sampled at different positions) and the distance preservation between the embedded coordinates. We further solidify this new insight using empirical evaluations. 
    \item As a practical example, we employ a Gaussian signal as the embedding function and show that it can deliver on-par performance with the Fourier frequency mapping. Most importantly, we demonstrate that the Gaussian embedding is more efficient in terms of the embedding dimension while being less volatile. 
    \item We show that trading a complex positional encoding for a deep network allows us to encode high-frequency features with a substantial speedup by circumventing the heavy computation for a simple positional encoding combined with a deep neural network. Promising empirical reconstruction performance is obtained on 1D, 2D, and 3D signals using our proposed embedding function in conjunction with coordinate networks. 
\end{itemize}

\begin{figure}[ht]
    \centering
    \subfloat[\centering]{{\includegraphics[width=0.42\textwidth]{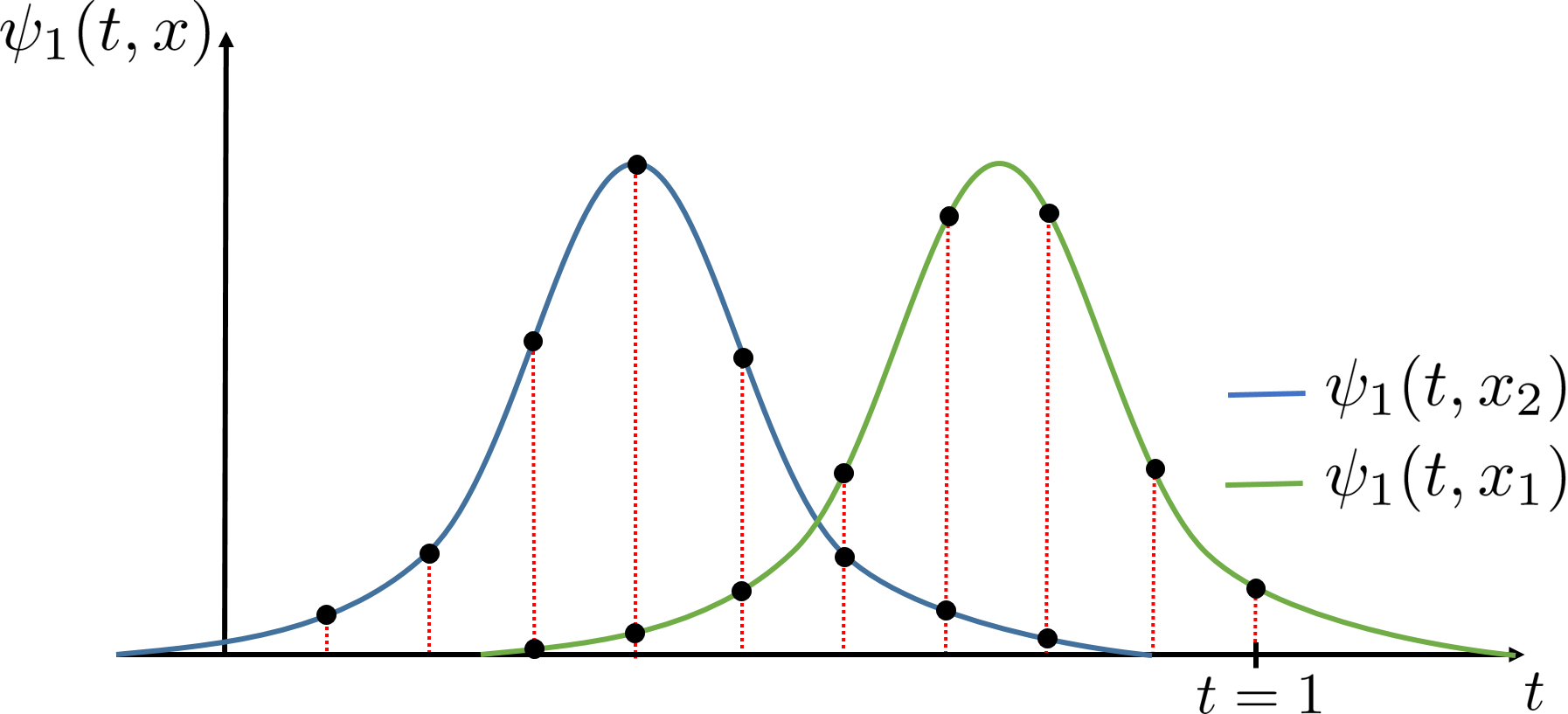} }}%
    \qquad
    \subfloat[\centering]{{\includegraphics[width=0.42\textwidth]{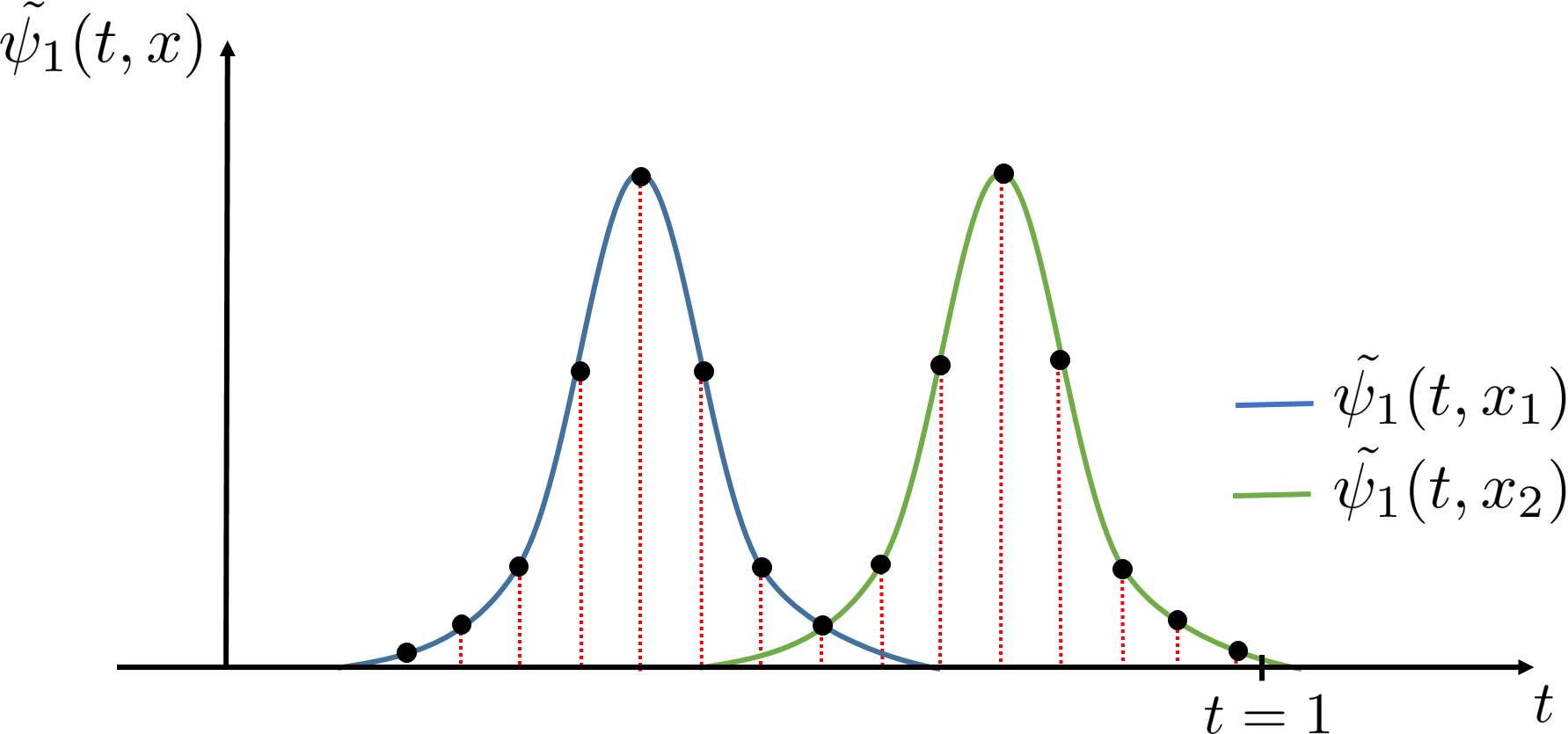} }}%
    \qquad
    \subfloat[\centering]{{\includegraphics[width=0.42\textwidth]{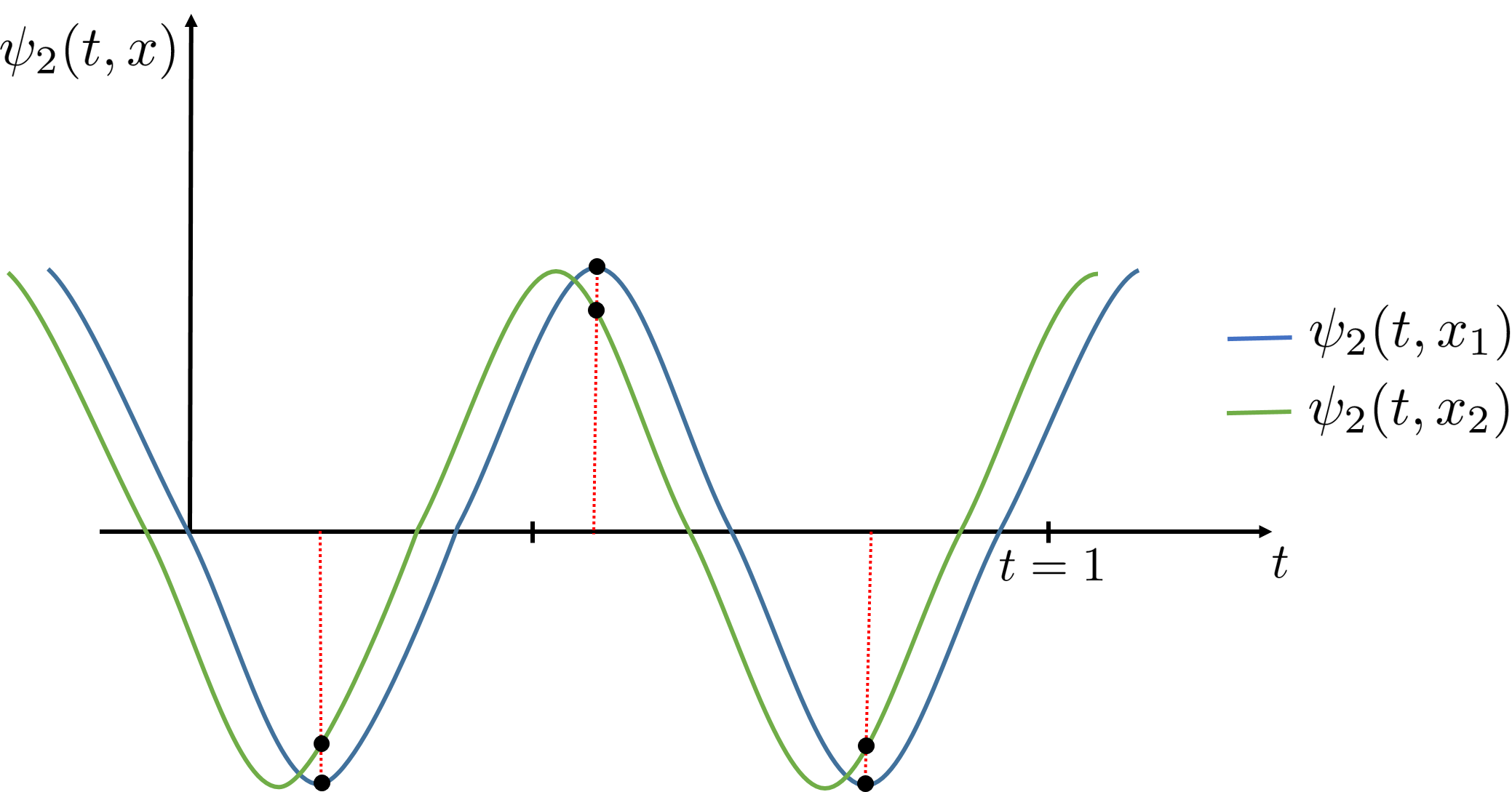} }}%
     \qquad
    \subfloat[\centering]{{\includegraphics[width=0.42\textwidth]{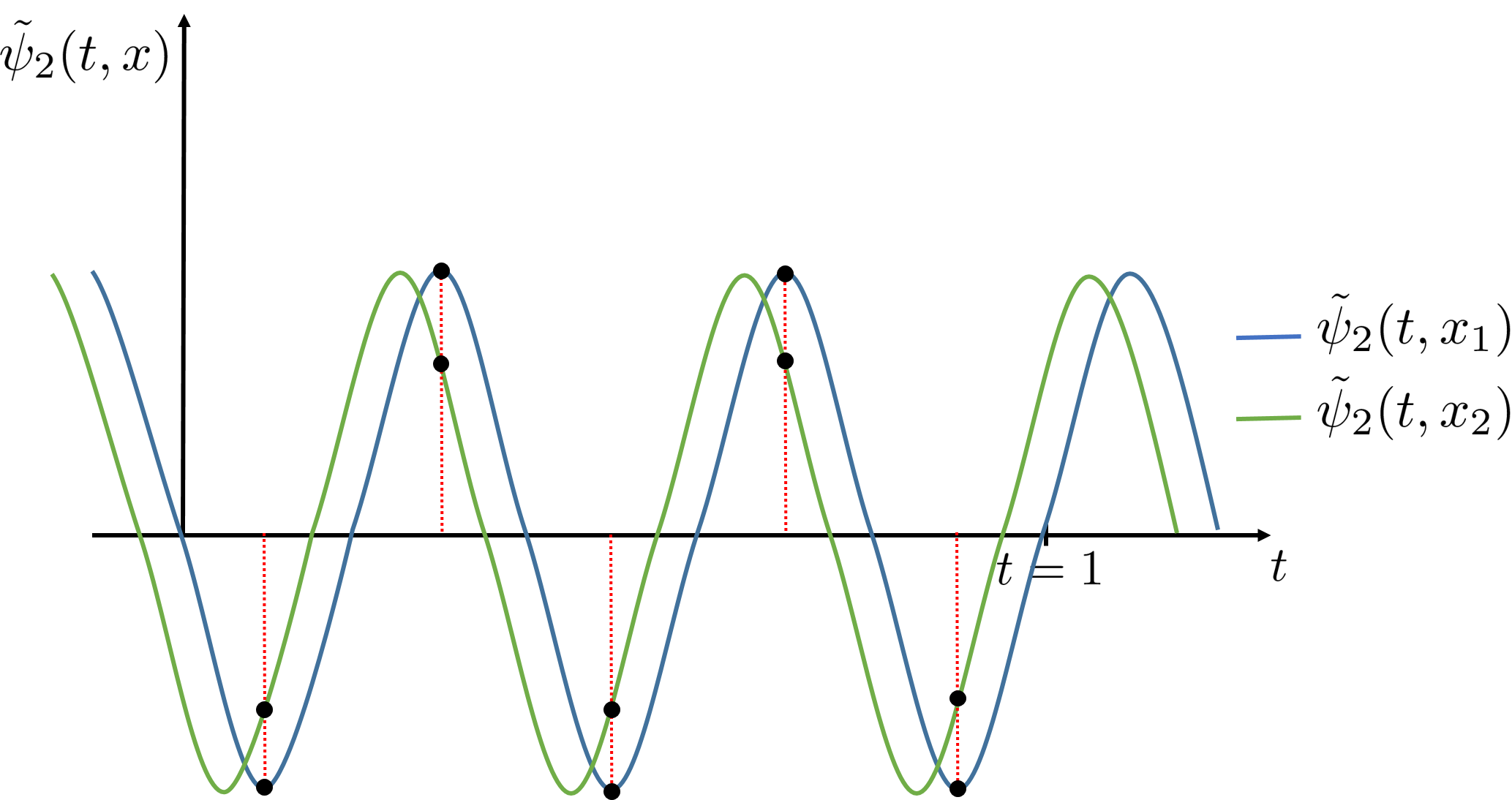} }}%
    \caption{Overview of the proposed positional encoding scheme. Positions are encoded as equidistant samples from shifted basis functions (embedders). The shifts are determined by the corresponding coordinate positions we are wanting to embed. In (a) and (b),  $x_1$ and $x_2$ are encoded as samples from shifted Gaussians with a higher and a lower standard deviation, respectively. Note that we need a higher number of samples for (b) due to higher bandwidth (see~\cref{sec:methodology}). In (c) and (d), $x_1$ and $x_2$ are encoded with sinusoidal signals with a different frequencies. Note that although different sampling rates are employed for (c) and (d), the same two values are repeated across the samples. Hence, sampling more than twice is redundant.}%
    \label{fig:example}%
\end{figure}

\section{Related works}
\label{sec:related_work}
Positional encoding became a popular topic among the machine learning community after the seminal work on Transformers by Vaswani~\etal~\cite{vaswani2017attention}  Since the attention mechanism used in the Transformers is position-insensitive, they employed a sinusoidal signal to encode the positions before feeding them to the higher blocks. A contemporary work by Gehring~\etal~\cite{gehring2017convolutional} also proposed a convolutional seq2seq model, adapting a positional encoding mechanism. Since then, using positional encoding in language models became a common trend~\cite{shaw2018self, dai1901attentive, raffel2019exploring, he2020deberta, kitaev2018constituency}. Notably, Wang~\etal~\cite{wang2019encoding} extended the embedding space from real numbers to complex values. Another critical aspect of their work is replacing the pre-defined encoding mechanism with a learnable one. There have also been other exciting attempts to improve positional encoding, such as extending the sequential positional encoding to tree-based positional encoding~\cite{shiv2019novel}, untying the correlations between words and positions while embedding coordinates~\cite{ke2020rethinking}, and modeling positional encoding using dynamical systems~\cite{liu2020learning}.

In parallel, positional encoding is also gaining attention in computer vision, specifically with coordinate-MLPs. Coordinate-MLPs provide an efficient method to encode objects such as images~\cite{nguyen2015deep, stanley2007compositional}  and 3D scenes~\cite{niemeyer2020differentiable, saito2019pifu, sitzmann2019scene} as their weights. Remarkably, Mildenhall~\etal~\cite{mildenhall2020nerf} and Zhong~\etal~\cite{zhong2019reconstructing} found that encoding coordinates with sinusoidal signals allow coordinate-MLPs to learn high frequency content better.  One of the earliest roots of this approach can perhaps be traced to the work by Rahimi and Recht~\cite{rahimi2007random}, where they used random Fourier features to approximate an arbitrary stationary kernel function by applying
Bochner’s theorem. More recently, Tancik~\etal~\cite{tancik2020fourier}, leveraging the NTK theory~\cite{arora2019fine, bietti2019inductive, du2018gradient, jacot2018neural,lee2019wide}, recently added theoretical rigor to this particular practice by showing that such embeddings enable tuning the spectrum of the NTK of the corresponding MLP. In contrast, the goal of this paper is to show that one does not have to be limited to the Fourier embedding for positional encoding. We demonstrate that alternative functions can be used for positional encoding while gaining similar or better performance compared to Fourier embedding.

\section{Positional encoding: a theoretical walk-through}
\label{sec:methodology}
This section contains an exposition of the machinery and fundamentals necessary to understand the proposed framework. We begin our analysis by considering a simple linear learner since rigorous characterization of a linear learner is convenient compared to a non-linear model. Therefore, we study a linear learner and empirically show that the gathered insights are extendable to the non-linear models.

First, we show that the capacity to memorize a given set of training data entirely depends on the (approximate) rank of the embedding matrix. Next, we establish that for generalization, the rank should be upper-bounded against the number of coordinates,~\ie, the embedding function should be bandlimited~\footnote{We assume that in regression, the smoothness of a model is implicitly related to generalization.}. We incur a crucial insight here that positional encoding essentially portrays a trade-off between memorization and generalization. Afterward, we discuss the importance of distance preservation between embedded coordinates and its relationship to bandlimited embedding functions. Finally, we consider several possible embedder functions and analyze their behavior using the developed tools.

\subsection{Rank of the embedded representation}
\label{sec:rank}
Let $\x {=} [x_1, x_2, {\cdots}, x_N]^T$  be a vector of 1D coordinates, in which $x_i {\in} [0,C]$. And let $\y {=} [y_1, y_2, {\cdots}, y_n]^T$ be the corresponding outputs of a function $f{:}\mathbb{R} {\to} \mathbb{R}$. Our goal is to find a $d$ dimensional embedding $\Psi{:}\mathbb{R} {\to} \mathbb{R}^d$ for these positions, so that a linear model can be employed to learn the mapping $f$ as,
\begin{equation}
\label{equ:linear_learner}
    \textbf{w}^T\Psi(\x) + b \approx f(\cdot)\:,
\end{equation}
where $\textbf{w} {\in} \mathbb{R}^{d}$ and $b {\in} \mathbb{R}$ are the learnable weights and the bias, respectively. Then, it is straightforward to show that for the perfect reconstruction of \emph{any} given $\y$ using~\cref{equ:linear_learner}, the following condition should be satisfied: 
\begin{equation}
\label{equ:rank_condition}
    \mathrm{Rank}\left\{\left[\Psi(x_1) \, \Psi(x_2) \, \dots \, \Psi(x_N) \right]\right\}= N\:.
\end{equation}
Thus, we establish the following Proposition:
\begin{prop}
Consider a set of coordinates $\x {=} [x_1, x_2, {\cdots}, x_N]^T$, corresponding outputs $\y {=} [y_1, y_2, {\cdots}, y_N]^T$, and a $d$ dimensional embedding $\Psi{:}\mathbb{R} {\to} \mathbb{R}^d$. Under perfect convergence, the  sufficient condition for a linear model for perfectly memorizing the mapping between $\x$ and $\y$ is for $\X {=} [\Psi(x_1), \Psi(x_2), {\dots}, \Psi(x_N)]^T$ to have full rank.
\end{prop}

\subsection{Bandlimited embedders}
One possible way of enforcing the condition in~\cref{equ:rank_condition} is to define an embedding scheme where the rank of the embedded matrix strictly monotonically increases with $N$ (for a sufficiently large $d$). As depicted in~\cref{sec:rank}, this would ensure that the model can memorize the training data and therefore perfectly reconstruct $\y$. However, memorization alone does not yield a good model. On the contrary, we also need our model to be generalizable to unseen coordinates.

To this end, let us define elements of $\Psi(\cdot)$ as sampled values from a function $\psi{:}\mathbb{R}^2 {\to} \mathbb{R}$ such that for a given $x$,
\begin{equation}
\label{equ:embedding_matrix}
    \Psi(x) = [\psi(0,x), \psi(s,x), \dots, \psi((d-1)s,x)  ]^T\:,
\end{equation}
where $s {=} Cd^{-1}$ is the sampling interval. We shall refer to $\psi(\cdot)$ as the \emph{embedder}. As discussed above, for better generalization, we need, 
\begin{equation}
\label{equ:band_limited}
    \psi(t,x) \approx \sum_{b = 0}^{B} \alpha_{b} \beta_{b}(t)\:,
\end{equation}
where~$\alpha_b$ and~$\beta_b(t)$ are weights and shifted basis functions, respectively, that can approximately estimate~$\psi(t,x)$ at any arbitrary position~$x$. We refer to such embedders as~\emph{bandlimited embedders} with a bandwidth $B$. This is equivalent to saying that the embedding matrix has a bounded rank, \ie, the rank cannot increase arbitrarily with $N$. 
The intuition here is that if $B$ is too small, the model will demonstrate poor memorization and overly smooth generalization. On the other hand, if $B$ is extremely high, the model is capable of perfect memorization but poor generalization. Therefore we conclude that for ideal performance, the embedder should be chosen carefully, such that it is both bandlimited and has a sufficient rank. As we shall discuss the bandwidth~$B$ can also act as a guide for the minimal value of~$d$. 

\subsection{Distance preservation}
Intuitively, the embedded coordinates should preserve the distance between the original coordinates, irrespective of the absolute position. The embedded distance (or similarity) $\mathrm{D(\cdot,\cdot)}$ between two coordinates $(x_1,x_2)$ can be measured via the inner product $\mathrm{D}(x_1,x_2) {=} \int_{0}^{1} \psi(t,x_1)\psi(t,x_2)dt$. For ideal distance preservation we need,
\begin{equation}
   \lVert {x_1 - x_2} \rVert \propto \mathrm{D}(x_1,x_2)\:.
\end{equation}

Interestingly, this property is also implicitly related to the limited bandwidth requirement. Note that in practice, we employ sampled embedders to construct $\Psi$ as shown in~\cref{equ:embedding_matrix}. Hence, the dot product between the sampled $\psi(t, x_1)$ and $\psi(t, x_2)$ should be able to approximate $\mathrm{D}$ as,
\begin{equation}
\label{equ:distance_preservation}
    \mathrm{D}(x_1,x_2) = \int_{0}^{C} \psi(t,x_1)\psi(t,x_2)dt \approx \sum_{d = 0}^{d-1}\psi(s\cdot d,x_1)\psi(s \cdot d,x_2)\:,
\end{equation}
which is possible, if and only if, $\psi$ is bandlimited. In that case, $d {=} B$ is sufficient where $B$ is the bandwidth of $\psi$ (by Nyquist sampling theory). In practice, we choose $C{=}1$.

\begin{remark}
The embedder should be bandlimited for better generalization (equivalently, the rank of the embedded matrix should be upper-bounded). Further, the ideal embedder should essentially face a trade-off between memorization and generalization. Here, memorization correlates with the rank of the embedded matrix, while generalization relates to the distance preservation between the embedded coordinates.
\end{remark}

\section{Analysis of possible embedders}
\label{sec:analysis}
\vspace{-0.2cm}
Although our derivations in~\cref{sec:methodology} are generic, it is imperative to carefully choose a specific form of $\psi(\cdot, \cdot)$, such that properties of candidate embedders can be conveniently analyzed. Hence, we define embedders in terms of shifted basis functions,~\ie, $\psi(t,x){=} \psi(t{-}x)$. Such a definition permits us to examine embedders in a unified manner, as we shall see below.

Moreover, the rank of a matrix can be extremely noisy in practice. Typically, we need to heuristically set an appropriate threshold to the singular values, leading to unstable calculations. Therefore, we use the stable rank~\cite{rudelson2007sampling} instead of the rank in all our experiments. In particular, the stable rank is a more stable surrogate for the rank, and is defined as $\frac{\|\A\|_{F}^2}{\|\A\|_{2}^2}$, where $\A$ is the matrix, $\| {\cdot} \|_F$ is the Frobenius norm, and $\| {\cdot} \|_2$ is the matrix norm. From here onwards, we will use the terms rank, approximate rank, and stable rank interchangeably.


\noindent\textbf{Impulse embedder.}\:\: One simple way to satisfy the condition of~\cref{equ:rank_condition} for an arbitrary large ${N}$ is to define $\psi(t,x) {=} \delta (t{-}x)$, where $\delta({\cdot})$ is the impulse function. 
Note that using an impulse embedder essentially converts the embedding matrix to a set of one-hot encodings. With the impulse embedder, we can perfectly memorize a given set of data points, as the embedded matrix has full rank. The obvious drawback, however, is that the bandwidth of the impulse embedder is infinite,~\ie, assuming a continuous domain, $d$ needs to reach infinity to learn outputs for all possible positions. Hence, the distance preservation is hampered, and consequently, the learned model lacks generalization.

\noindent\textbf{Rectangle embedder.}\:\: As an approximation of impulse function (unit pulse), rectangular function $rect(x){=}1$ when $|x|{<}\frac{1}{2}$ and $rect(x){=}0$ when $|x|{>}\frac{1}{2}$. We can define $\psi(x){=}rect\left(\frac{x{-}t}{d}\right)$, where $d$ is the width of the impulse. Immediately we know the stable rank of rectangle embedder is $\min\left(N,\frac{1}{d}\right)$, where $N$ is the number of sampled coordinates, the distance function $D(x_1,x_2){=}tri(x_1{-}x_2)$, where $tri(\cdot)$ is triangular function. A physical way to understand rectangle embedder is nearest neighbour regression.

\begin{figure}[t]
    \centering
    \subfloat[\centering]{\includegraphics[width=0.46\textwidth]{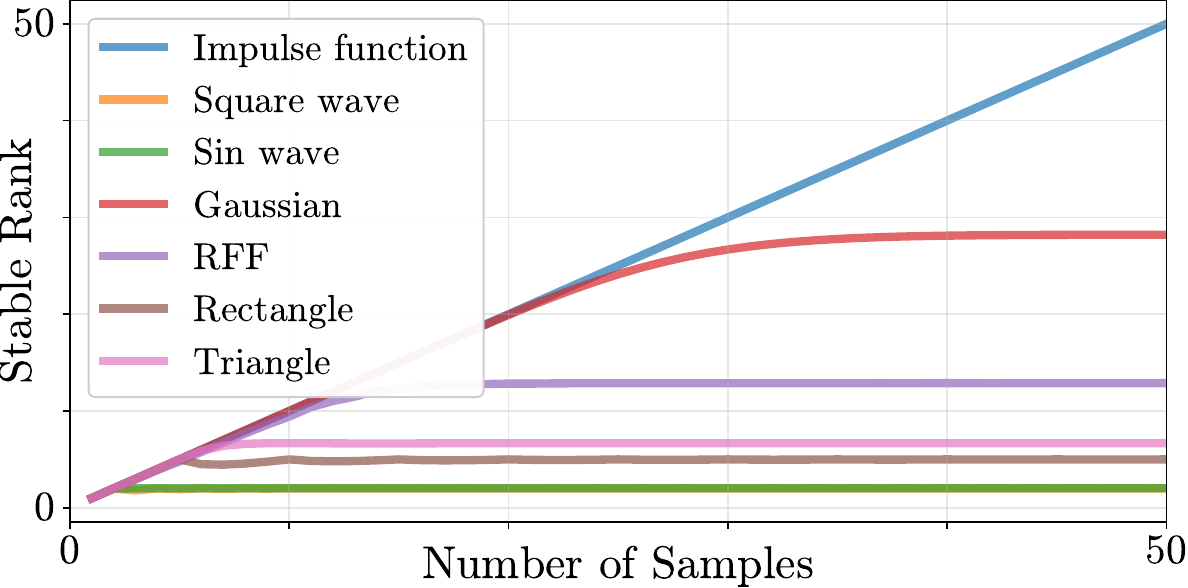}}%
    \qquad
    \subfloat[\centering]{\includegraphics[width=0.46\textwidth]{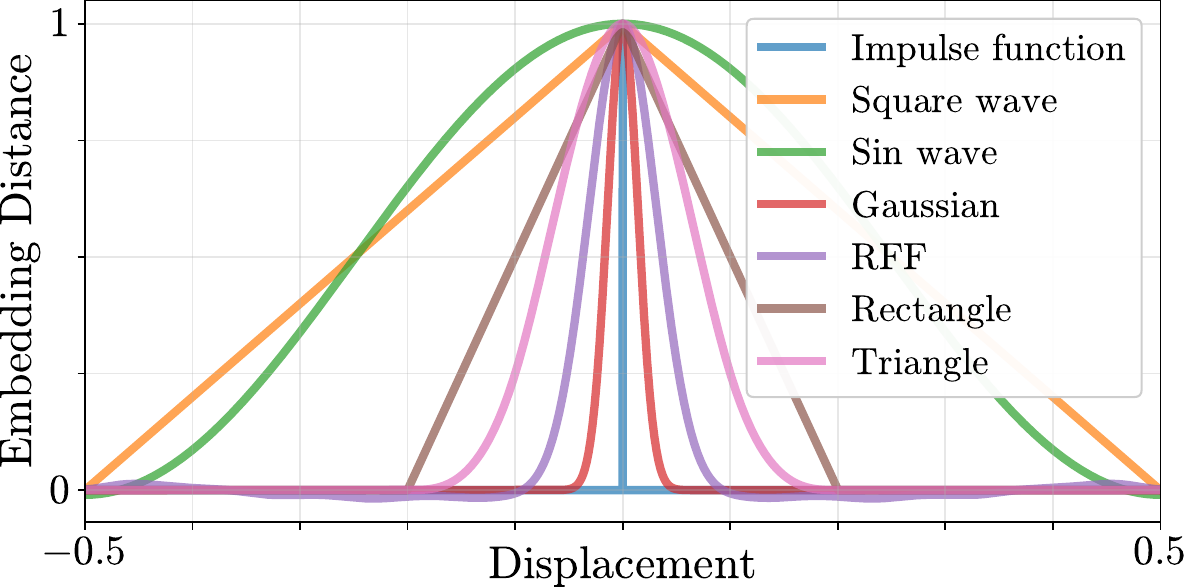}}%
    \caption{Quantitative comparison of (a) the stable rank and (b) the distance preservation of different embedders and RFFs. As expected, the stable rank of the impulse embedder strictly increases with the number of sampled points, causing poor distance preservation. The stable rank of the sine embedder is upper-bounded at $2$. The stable ranks of the square embedder and the sine embedder almost overlap. However, if the sample numbers are extremely high (not shown in the figure), their stable ranks begin to deviate. Similarly, the square embedder demonstrates perfect distance preservation, and the sine embedder is a close competitor. In contrast, the Gaussian embedder and the RFF showcase mid-range upper bounds for the stable rank and adequate distance preservation, advocating a much better trade-off between memorization and generalization.}%
    \label{fig:embeddercomparison}%
\end{figure}

\noindent\textbf{Triangle embedder.}\:\: A better choice to approximate impulse function may be triangular function, which is defined as $tri(x){=}\max(1{-}|x|,0)$. Thus the embedder is defined as $\psi(x){=}tri\left(\frac{x{-}t}{0.5d}\right)$. Here the factor $0.5$ makes the width of the triangle to be $d$. The stable rank of triangular embedder is $\min\left(N,\frac{4/3}{d}\right)$. When $d$ is the same, triangle embedder has a higher stable rank than rectangle embedder. The distance function of triangular embedder is $D(x_1,x_2){=}\frac{1}{4}\max
\left(d{-}|x_1{-}x_2|,0\right)^2$. This distance function looks really like Gaussian function, as illustrated in~\cref{fig:embeddercomparison}. A physical way to understand triangle embedder is linear interpolation.

\noindent\textbf{Sine embedder.}\:\: Consider $\psi(t,x) {=} \mathrm{sin}(f(t{-}x))$ for an arbitrary fixed function $f$. Since $\mathrm{sin}(f(t{-}x)) {=} \mathrm{sin}( ft)\mathrm{cos}(fx) {-} \mathrm{cos}(ft)\mathrm{sin}(fx)$, elements of any row of the embedding matrix can be written as a linear combination of the corresponding $\mathrm{sin}(ft)$ and $\mathrm{cos}(ft)$. Thus, the rank of the embedding matrix is upper-bounded at $2$. Consequently, the expressiveness of the encoding is limited, leading to poor memorization and overly smooth generalization (interpolation) at unseen coordinates.

\noindent\textbf{Square embedder.}\:\: Let us denote a square wave with unit amplitude and period $2\pi$ as $\mathrm{sgn}(\mathrm{sin}(t))$, where $\mathrm{sgn}$ is the sign function. Then, define $\psi(t,x) {=} \mathrm{sgn}(\mathrm{sin}(t{-}x))$. It is easy to deduce that the embedded distance $D(x_1, x_2) {=} 1 {-} 2\lVert x_1 {-} x_2 \rVert, \forall |x| {\leq} 1$ which implies perfect distance preservation. The drawback, however, is that the square wave is not bandlimited. Thus, it cannot approximate the inner product $\int \psi(t,x)\psi(t,x')$ using a finite set of samples  as in~\cref{equ:distance_preservation}. However, an interesting attribute of the square wave is that it can be decomposed into a series of sine waves with odd-integer harmonic frequencies as $\mathrm{sgn}(\mathrm{sin}(t)) {=} \frac{4}{\pi} \big[ \mathrm{sin}(t) {+} \frac{1}{3}\mathrm{sin}(3t) {+} \frac{1}{5}\mathrm{sin}(5t) {+} \frac{1}{7}\mathrm{sin}(7t) {+} \dots \big]$. 
In other words, its highest energy (from a signal processing perspective) is contained in a sinusoidal with the same frequency. Thus, the square wave can be \emph{almost} approximated by a sinusoidal signal. In fact, the square wave and the sinusoidal demonstrate similar properties in terms of the stable rank and the distance preservation (see~\cref{fig:embeddercomparison}).  
\begin{wrapfigure}[13]{r}{0.45\textwidth}
\centering
\includegraphics[width=0.45\textwidth]{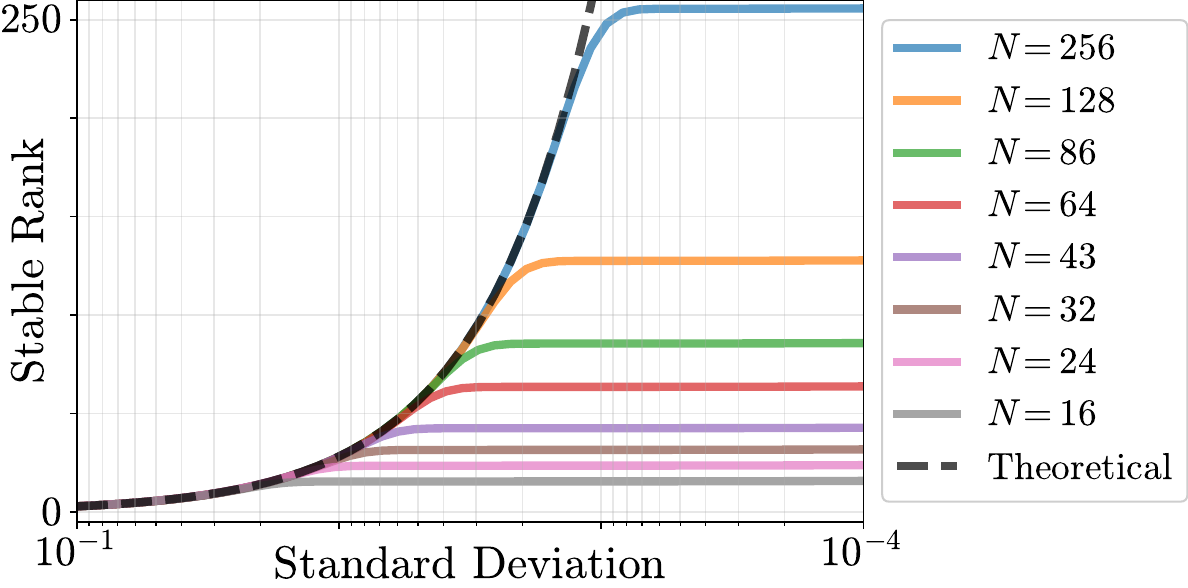}
\caption{\label{fig:gaussian_sr} Stable rank of the Gaussian embedder vs the standard deviation for different number of samples. The dash line is the theoretical stable rank $\frac{1}{2\sqrt{\pi}\sigma}$}
\end{wrapfigure}

\noindent\textbf{Gaussian embedder.}\:\: We define the Gaussian embedder as  $\psi(t,x) {=} \exp \left({-}\frac{\|t{-}x\|^2}{2\sigma^2}\right)$ where $\sigma$ is the standard deviation. The Gaussian embedder is also approximately bandlimited like the square embedder. However, the Gaussian embedder has a higher upper bound for the stable rank that can be controlled by $\sigma$. More precisely, when the embedding dimension is large enough, the stable rank of the Gaussian embedding matrix and the embedded distance between coordinates can be obtained analytically as shown below.

\begin{prop}
\label{prop:gaussian}
Let the Gaussian embedder be denoted as $\psi(t,x) {=} \exp \left({-}\frac{\|t{-}x\|^2}{2\sigma^2}\right)$. With a sufficient embedding dimension, the stable rank of the embedding matrix obtained using the Gaussian embedder is $\min\left(N, \frac{1}{2\sqrt{\pi}\sigma}\right)$ where $N$ is the number of embedded coordinates. Under the same conditions, the embedded distance between two coordinates $x_1$ and $x_2$ is $\mathrm{D}(x_1,x_2) {=} \exp \left({-}\frac{\|x_1{-}x_2\|^2}{4\sigma^2}\right)$.
\end{prop}
(see~\cref{fig:gaussian_sr} for an experimental illustration). It is clear from~\cref{prop:gaussian} that as the number of sampled positions goes up, the stable rank of the Gaussian embedding matrix will linearly increase until it reaches its upper bound. 
Finally,~\cref{fig:embeddercomparison} empirically validates the theoretically discussed properties of different embedders.

\subsection{Connection to the Random Fourier Features}
\label{sec:connectiontorff}

The prominent way of employing Fourier frequency mapping is via Random Fourier Features (RFF) mapping~\cite{tancik2020fourier}, where the frequencies are randomly sampled from a Gaussian distribution with a certain standard deviation $\sigma$. In this Section, we show that RFF mapping can be analyzed through the lens of our theoretical framework discussed thus far. To this end, we first establish the following proposition:
\begin{prop}
\label{prop:rff}
Let the RFF embedding be denoted as $\gamma(x){=}[\cos(2\pi\mathbf{b}x),\sin(2\pi\mathbf{b}x)]$, where $\mathbf{b}$ are sampled from a Gaussian distribution. When the embedding dimension is large enough, the stable rank of RFF will be $\min\left(N, \sqrt{2\pi}\sigma\right)$, where $N$ is the numnber of embedded coordinates. Under the same conditions, the embedded distance between two coordinates $x_1$ and $x_2$ is $\mathrm{D}(x_1,x_2) {=} \sum_j \cos{2\pi b_j(x_1{-}x_2)}$.
\end{prop}

\begin{figure}[t]
\centering
\begin{subfigure}{0.42\textwidth}
\includegraphics[width=\linewidth]{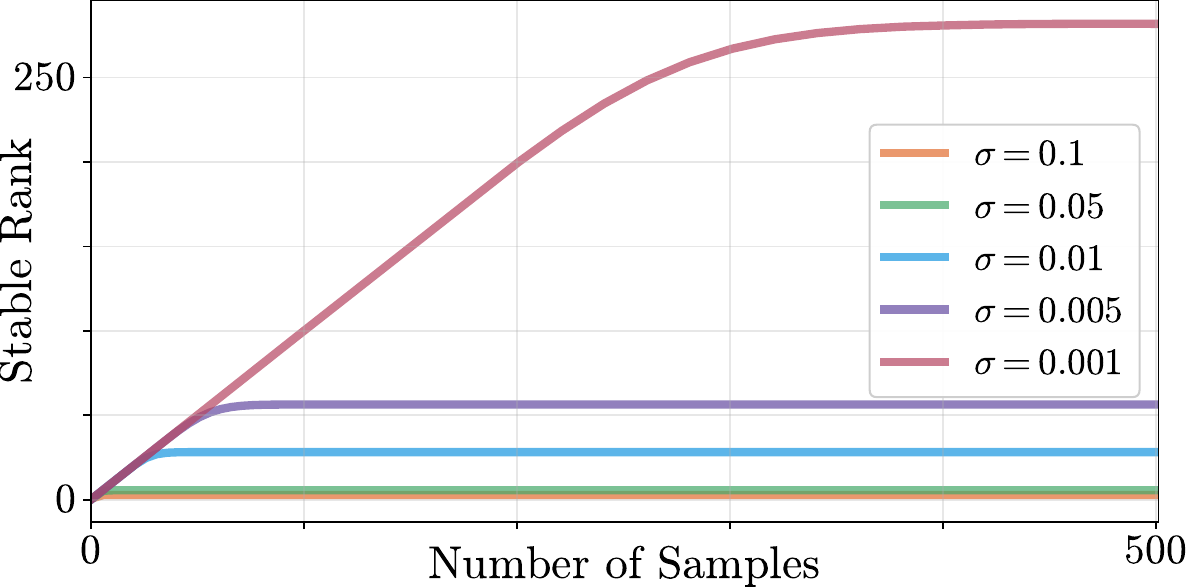} 
\end{subfigure} 
\begin{subfigure}{0.42\textwidth}
\includegraphics[width=\linewidth]{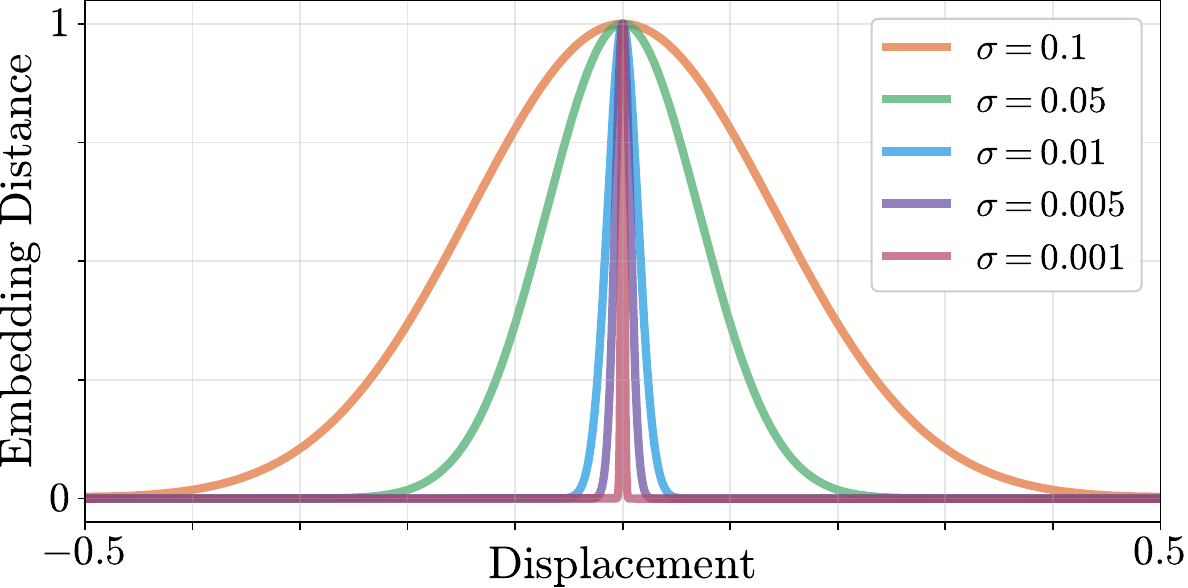} 
\end{subfigure}
\begin{subfigure}{0.42\textwidth}
\includegraphics[width=\linewidth]{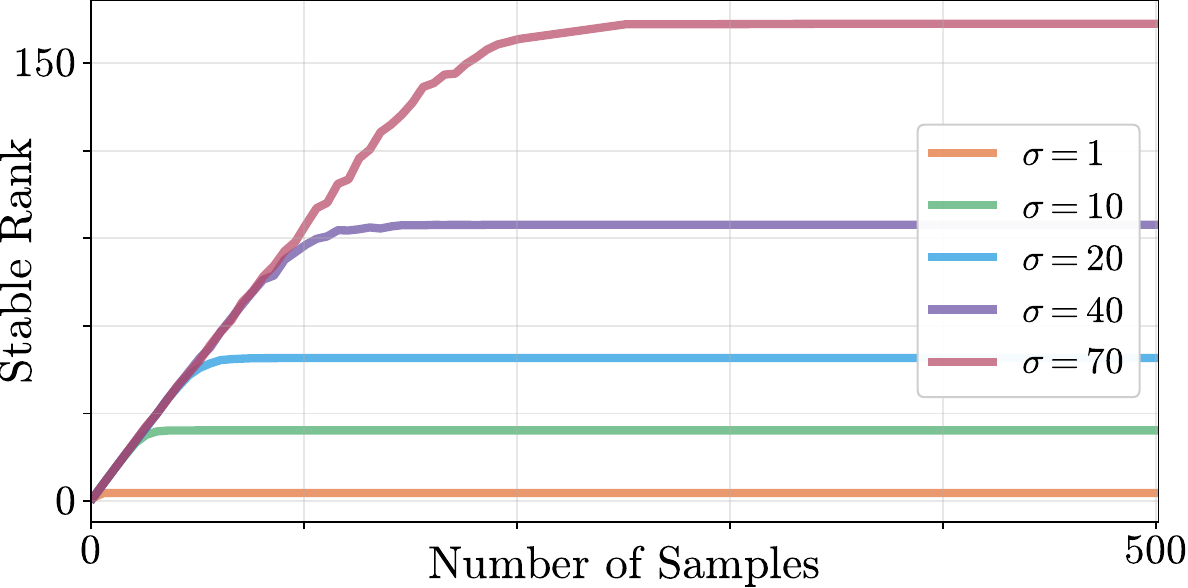} 
\end{subfigure} 
\begin{subfigure}{0.42\textwidth}
\includegraphics[width=\linewidth]{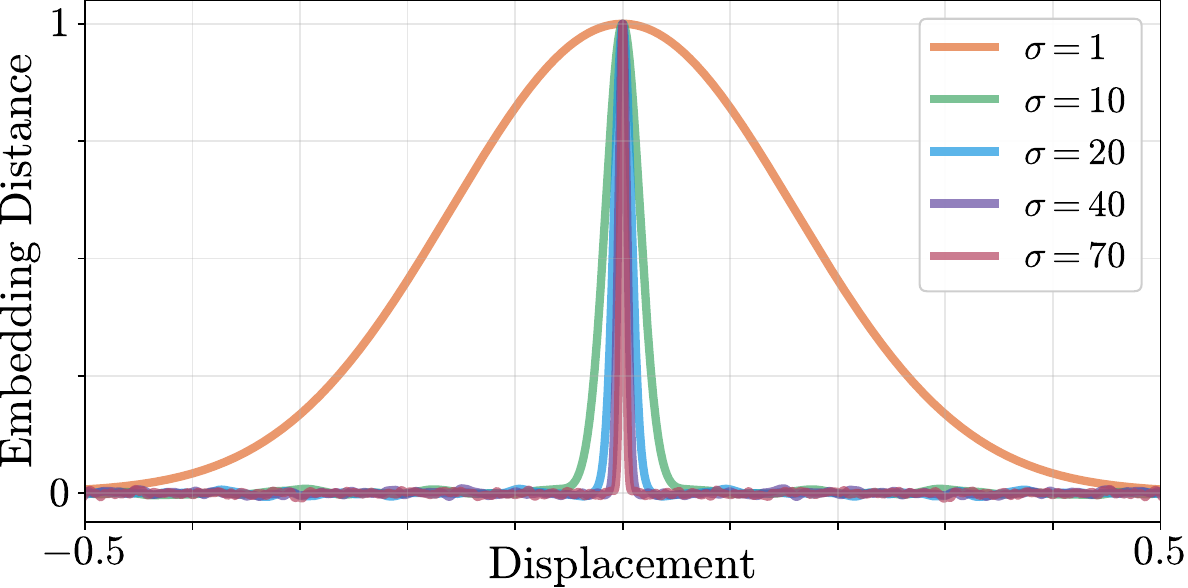} 
\end{subfigure}
\caption{The stable rank (left column) and distance preservation (right column) of Gaussian embedder (top row) and RFF (bottom row) across different standard deviations.}
\label{fig:Gauss_vs_RFF}
\end{figure}

As shown in~\cref{fig:Gauss_vs_RFF}, the stable rank of RFF increases linearly with the number of samples until it gets saturated at $\sqrt{2\pi}\sigma$. This indicates a relationship between RFF and Gaussian embedder. Let $\sigma_g$ and $\sigma_f$ be the standard deviations of Gaussian embedder and RFF. When their stable ranks are equal, $\frac{1}{2\sqrt{\pi}\sigma_g}{=}\sqrt{2\pi}\sigma_f$ (from~\cref{prop:gaussian},~\ref{prop:rff}). This implies that when $\sigma_g\sigma_f{=}\frac{1}{2\sqrt{2}\pi}$, these two embedders are equivalent in terms of the stable rank and distance preservation (see~\cref{fig:Gauss_vs_RFF} when $\sigma_g{=}0.01$ and $\sigma_f{=}0.1$).

A common observation with RFFs is that when $\sigma_f$ is too low, the reconstruction is overly smooth and if $\sigma_f$ is too high, it gives noisy interpolation~\cite{tancik2020fourier}. This observation directly correlates to our theory. In~\cref{fig:Gauss_vs_RFF}, as the standard deviation increases, the stable rank increases and distance preservation decreases. Similarly, When the standard deviation is too low, the stable rank decreases while distance preservation increases.

\begin{figure}[t]
\centering
\includegraphics[width=1.0\textwidth]{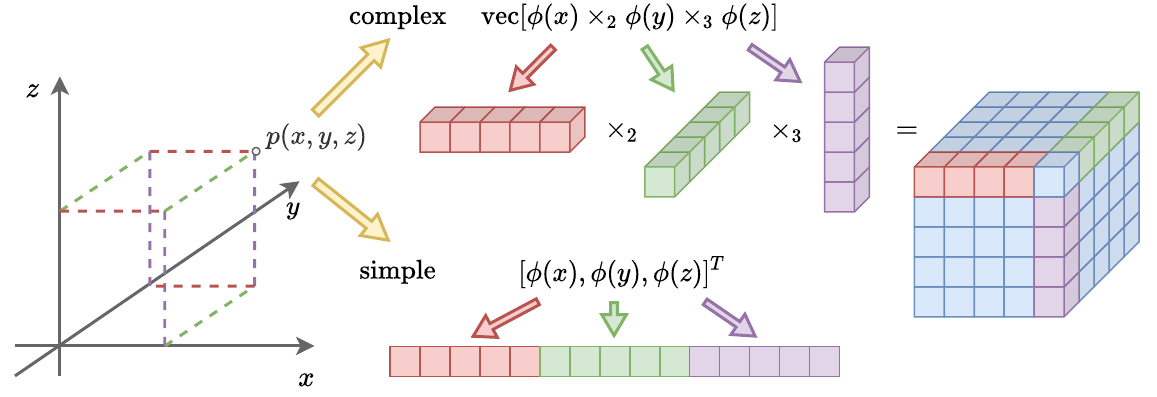}
\caption{Illustration of different methods to extend 1D encoding of length $K$ to $N$-dimensional case. Simple encoding is the widely used method currently, which is only the concatenation of embeddings in each dimension to get a $NK$ size encoding. We propose an alternative embedding method: conduct mode-$n$ product to get the $N$ dimensional cube, which we define as the complex embedding.}
\label{fig:complexvssimple}
\end{figure}

\section{Simplicity vs complexity in positional encoding}
\vspace{-0.2cm}
Thus far, we established that the positional encoding can achieved by  sampling shifted basis functions, and the well-known RFF-embedding can also be analyzed through this lens. However, the analysis so far  focused only on 1D coordinates. In this section, we shall investigate how to extend these positional embedding schemes to high-dimensional signals,~\eg, images and 3D signals.

\subsection{2D simple encoding}
\label{sec:2D-simple}
Suppose $\psi(\cdot)$ is a an arbitrary positional encoding function. We define simple positional encoding as the  concatenation of the 1D encoding in each dimension: $\Psi(x,y) = [\psi(x),\psi(y)]$. Then, with a linear model we have,
\begin{equation}
    I(x,y) \approx \mathbf{w}^T\Psi(x,y) =  \mathbf{w}_x^T\psi(x) + \mathbf{w}_y ^T \psi(y)\:.
\end{equation}
The above formula can be written in the matrix form as,
\startcompact{small}
\begin{equation}
    \begin{bmatrix}
    I(1,1) & {\dots} & I(N,1) \\
    {\vdots} & {\ddots} & {\vdots} \\
    I(1,N) & {\dots} & I(N,N)
    \end{bmatrix} {\approx }
    \underbrace{\begin{bmatrix}
    \mathbf{w}_x^T \psi(x_1) & {\dots} & \mathbf{w}_x^T \psi(x_N) \\
    {\vdots} & {\ddots} & {\vdots} \\
    \mathbf{w}_x^T \psi(x_1) & {\dots} & \mathbf{w}_x^T \psi(x_N)
    \end{bmatrix}}_{\begin{matrix} \mathbf{A} \end{matrix}} {+}
    \underbrace{\begin{bmatrix}
    \mathbf{w}_y^T \psi(y_1) & {\dots} & \mathbf{w}_y^T \psi(y_1) \\
    {\vdots} & {\ddots} & {\vdots} \\
    \mathbf{w}_y^T \psi(y_N) & {\dots} & \mathbf{w}_y^T \psi(y_N)
    \end{bmatrix}}_{\begin{matrix} \mathbf{B} \end{matrix}} \:.
\end{equation}
\stopcompact{small}
Clearly, $\mathbf{A}$ and $\mathbf{B}$ are rank $1$ matrices. Therefore, a linear network can only reconstruct a 2D image signal with at most rank $2$. 
This drawback can be addressed in most practical cases using deeper non-linear MLPs, since the rank of the representations can be increased with multiple layers.

\subsection{$2$D complex encoding}
\label{sec:2D-complex}
As opposed to simple encoding, we propose an alternative method for positional encoding in higher dimensions using the Kronecker product. With this approach, we can obtain a higher rank for the positional embedding matrix. For example, consider 2D inputs. Then, we can obtain the complex encoding as $\Psi(\x,\y) {=} \Psi(\x)\otimes\Psi(\y)$, where $\Psi(\x)$ and $\Psi(\y)$ are 1D encodings along each dimension. Also, the following relationship holds:    
\begin{equation}
    \mathrm{Rank}(\Psi(\x)\otimes\Psi(\y))= \mathrm{Rank}(\Psi(\x))\mathrm{Rank}(\Psi(\y))\:.
\end{equation}
However, the drawback is also obvious. The embedding dimension is squared, which takes significantly more memory and computational cost. However, we propose an elegant workaround for this problem given that the points are sampled on a regular grid,~\ie, when the coordinates are separable, using the following property of the Kronecker product,
\begin{equation}
    \mbox{vec}(\mathrm{S})^T \approx\mathrm{vec}(\mathbf{W})^T (\Psi(\x)\otimes\Psi(\y))=\mathrm{vec}(\Psi(\y)^T\mathbf{W}\Psi(\x))^T\:,
    \label{eq:kronecker_trick}
\end{equation}
where $\mathrm{S}_{i,j}{=}I(x_i,y_j)$. For instance, suppose we have ${N^2}$ number of 2D separable points where the feature length is $K$ for each dimension. The naive Kronecker product leads to $O(N^2K^2)$ computational complexity and $O(N^2K^2)$ memory complexity. Using~\cref{eq:kronecker_trick}, we reduce it dramatically to  $O(NK(N+K))$ computational complexity and $O(2NK+K^2)$ memory complexity. The key advantage of the complex encoding mechanism is that although it leads to a larger encoding matrix, the ability to achieve full rank allows us to use a single linear layer instead of an MLP, which reduces the computational cost of the neural network substantially. In addition, this enables us to obtain a closed-form solution instead of using stochastic gradient descent, leading to dramatically faster optimization.
\noindent More precisely, we need to solve,
\begin{equation}
    \arg\min_{\mathbf{W}} \| \mathrm{vec}(\mathrm{S}) -(\Psi(\x)\otimes\Psi(\y))^T\mathrm{vec}(\mathbf{W})\|_2^2\:,
\end{equation}
where the solution can be obtained analytically as,
\begin{equation}
    \mathbf{W} = (\Psi(\y)\Psi(\y)^T)^{-1}\Psi(\y)\mathrm{S}\Psi(\x)^T(\Psi(\x)\Psi(\x)^T)^{-1}\:.
\end{equation}
When the coordinates are not separable, we can still take advantage of ~\cref{eq:kronecker_trick} by adding a sparse blending matrix $B$ as,
\begin{equation}
    \mathrm{vec}(I(\x,\y))^T \approx\mathrm{vec}(\mathbf{W})^T (\Psi(\x)\otimes\Psi(\y))B=\mathrm{vec}(\Psi(\y)^T\mathbf{W}\Psi(\x))^TB \:.
\end{equation}
This procedure is equivalent to evaluating virtual points (sampled on a regular grid) and interpolating to arbitrary coordinates using interpolation weights of $B$. The nature of the interpolation depends on the basis function used for positional encoding. Suppose $x_0$ and $x_1$ are two grid points and $x_0{\leq} x{\leq} x_1$. We want $\psi(x){\approx} \alpha_0\psi(x_0){+}\alpha_1\psi(x_1)$. It can be solved by
\vspace{-0.12cm}
\begin{equation}
    \arg\min_{\alpha}\|\psi(x)- \begin{bmatrix}\psi(x_0)&\psi(x_1)\end{bmatrix}\alpha\|_2^2\:,
\end{equation}
where $\alpha{=}\begin{bmatrix}\alpha_0&\alpha_1\end{bmatrix}^T$. With the definition $D(x)$ in~\cref{sec:rank}, consider $d {=} x_1{-}x_0$, which is the grid interval. Then, $x {=} x_0 {+} \beta d$, where $0{\leq}\beta{\leq}1$, and we have
\begin{equation}
   \alpha = \frac{1}{D^2(0)-D^2(d)}\begin{bmatrix}D(0)&-D(d)\\-D(d)&D(0)\end{bmatrix}\begin{bmatrix}D(\beta d)\\D((1-\beta) d)\end{bmatrix}\:.
\end{equation}
Please refer to the supplementary material for more details.

\begin{figure}[t]
\centering
    \subfloat[\centering]{\includegraphics[width=0.6\textwidth]{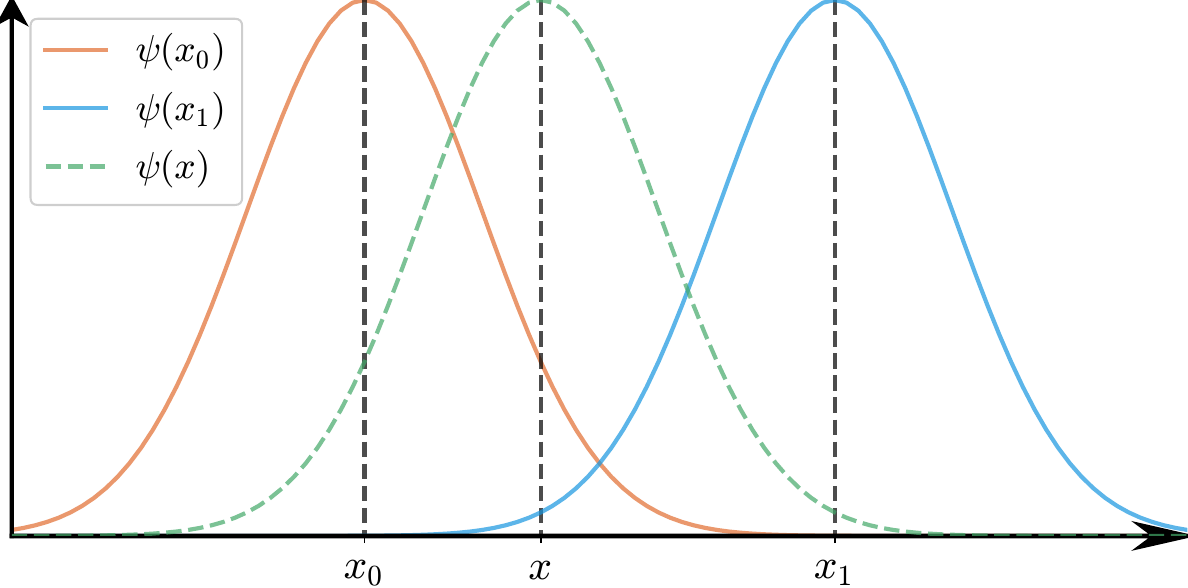}}%
    \qquad
    \subfloat[\centering]{\includegraphics[width=0.3\textwidth]{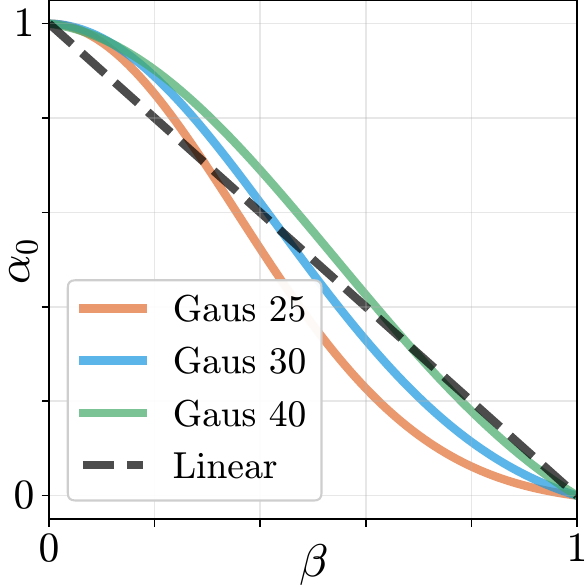}}%
\caption{\textbf{Relationship between $\alpha_0$ and $\beta$}. (a) is an example positional encoding at $x_0$, $x_1$ and $x$. x-axis is along the feature dimension and y-axis is the encoding value. Here, we want to express $\psi(x)$ as a linear combination of $\psi(x_0)$ and $\psi(x_1)$. (b) is the relationship between the interpolation weight $\alpha_0$ and the position ratio $\beta$. Instead of simple linear interpolation (dash line), the actual relation depends on the encoding function.}
\label{fig:ab-relation}
\end{figure}

\subsection{High dimensional encoding}
\label{sec:HD}
 \vspace{-0.15cm}
Both simple and complex encoding methods discussed in the previous sections can be extended easily to arbitrarily high dimensions. Suppose $\x {=} [x_1, x_2, {\cdots}, x_D]^T$ are the coordinates of a $D$ dimensional point, $\psi{:}\mathbb{R}{\to}\mathbb{R}^{K}$ is the 1D encoder and $\Psi^{(D)}(\cdot)$ is the $D$ dimensional encoding function. Then, the simple encoding is
\vspace{-0.2cm}
\begin{equation}
    \Psi^{(D)}(\x)=[\psi(x_1),\psi(x_2), \cdots, \psi(x_D)]^T\:.
\end{equation}
Similarly, the complex encoding can be obtained via the Kronecker product between each encoding as,
\vspace{-0.2cm}
\begin{equation}
    \Psi^{(D)}(\x)=\mbox{vec}([\psi(x_1)\times_2\psi(x_2)\times_3 \cdots\psi(x_3)\cdots\times_D \psi(x_D)])\:.
\end{equation}
Then, we can again extend the workaround we used in Eq.~\ref{eq:kronecker_trick} to multiple dimensions as,
\vspace{-0.2cm}
\begin{equation}
    \mbox{vec}(\W)^T\Psi^{(D)}(\x)= {\W}{\times_1}{\psi(x_1)}{\times_2}{\psi(x_2)}{\cdots}{\times_D}{\psi(x_D)}\:.
\end{equation}
Fig.~\ref{fig:complexvssimple} graphically illustrates the simple and complex positional encoding schemes.

\begin{table}[t]
    \setlength{\tabcolsep}{9pt}
    \centering
    \caption[]{Performance of natural 2D image reconstruction without random sampling (separable coordinates).
    \tikz\draw[taborange,fill=taborange] (0,0) circle (.5ex); are simple positional encodings. \tikz\draw[tabpurple,fill=tabpurple] (0,0) circle (.5ex); are complex positional encodings with gradient descent. \tikz\draw[tabgreen,fill=tabgreen] (0,0) circle (.5ex); are complex positional encodings with closed-form solution. We did not find a result for the complex encoding LogF, since it is ill-conditioned (rank deficient).}
     \vspace{-1em}
    \begin{adjustbox}{width=0.67\linewidth}
    \begin{tabular}{lccccc}
    \toprule
    & \thead{PSNR} & \thead{No. of params (memory)} & \thead{Time (s)}
    \\ \midrule
    \tikz\draw[taborange,fill=taborange] (0,0) circle (.5ex);~LinF
    & $26.12\pm3.99$ & $329,475 \:(1.32M)$ & $22.59$ \\
    \tikz\draw[taborange,fill=taborange] (0,0) circle (.5ex);~LogF
    & $26.11\pm4.04$    & $329,475 \:(1.32M)$ & $22.67$ \\
    \tikz\draw[taborange,fill=taborange] (0,0) circle (.5ex);~RFF~\cite{tancik2020fourier} 
    & $\mathbf{26.58\pm4.18} $ & $329,475 \:(1.32M)$ & $22.62$ \\
    \tikz\draw[taborange,fill=taborange] (0,0) circle (.5ex);~Tri
    & $25.96\pm3.88$ & $329,475 \:(1.32M)$ & $22.98$ \\
    \tikz\draw[taborange,fill=taborange] (0,0) circle (.5ex);~Gau
    & $26.17\pm4.17$    & $329,475 \:(1.32M)$ & $23.12$ \\
    \cmidrule{1-4}
    \tikz\draw[tabpurple,fill=tabpurple] (0,0) circle (.5ex);~LinF
    & $19.73\pm2.49$ & $196,608 \:(0.79M)$ & $1.93$ \\
    \tikz\draw[tabpurple,fill=tabpurple] (0,0) circle (.5ex);~LogF
    & $23.55\pm2.99$ & $196,608 \:(0.79M)$ & $1.85$ \\
    \tikz\draw[tabpurple,fill=tabpurple] (0,0) circle (.5ex);~RFF~\cite{tancik2020fourier} 
    & $24.34\pm3.24$ & $196,608 \:(0.79M)$ & $1.55$ \\
    \tikz\draw[tabpurple,fill=tabpurple] (0,0) circle (.5ex);~Tri
    & $26.65\pm3.57$ & $196,608 \:(0.79M)$ & $1.48$ \\
    \tikz\draw[tabpurple,fill=tabpurple] (0,0) circle (.5ex);~Gau
    & $\mathbf{26.69\pm3.74} $ & $196,608 \:(0.79M)$ & $2.01$ \\
    \cmidrule{1-4}
    \tikz\draw[tabgreen,fill=tabgreen] (0,0) circle (.5ex);~Tri
    & $26.36\pm3.39$ & $196,608 \:(0.79M)$ & $0.03$ \\
    \tikz\draw[tabgreen,fill=tabgreen] (0,0) circle (.5ex);~Gau
    & $\mathbf{26.69\pm3.76} $ & $196,608 \:(0.79M)$ & $0.18$ \\
    \arrayrulecolor{black}\bottomrule
    \end{tabular}
    \end{adjustbox}
    \label{tab:2d-separable}
\end{table}

\section{Experiments}
In this Section, we empirically confirm the advantages of using the proposed embedding procedure and verify that the theoretically predicted properties in the previous sections hold in practice. To this end, five encoding methods are compared: linearly sampled frequency (LinF), log-linearly sampled frequency (LogF), RFF, shifted Gaussian (Gau), and shifted triangle encoder (Tri).

\subsection{1D: Rank of input \& depth of Network}
In this experiment, we randomly sample 16 columns and 16 rows from $512{\times}512$ natural
\begin{figure}
\centering
\includegraphics[width=0.5\textwidth]{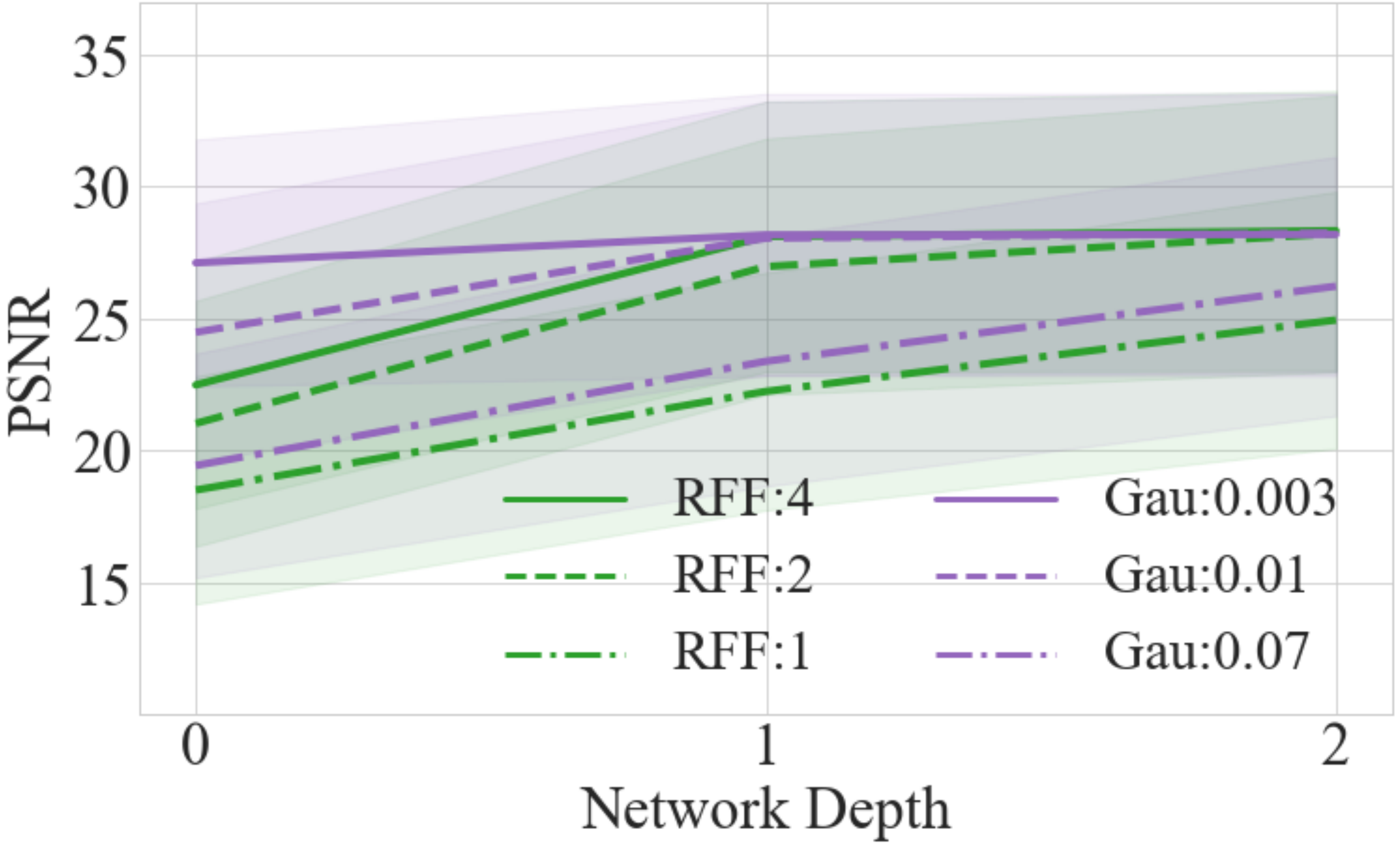}
\caption{\label{fig:rff_gauss} The performance of the reconstruction depends on the deepness of the network and the rank of the input embedding. }
\end{figure}
images from the image dataset in~\cite{tancik2020fourier}.
And we used 256 equally spaced points for training and the rest 256 points for testing. 
The encoding length is set to be 256, the same as the number of training points. Parameters were chosen carefully for each encoder 
to show the relationship of the encoding rank and the depth of the network.
The depth of the network changes from 0 (linear layer) to 2 with a fixed width of 256. 
From~\cref{fig:Gauss_vs_RFF}, we already know that the rank of the encoding matrix drops when $\sigma$ of Gau increases or $\sigma$ of RFF decreases.
The result in~\cref{fig:rff_gauss} shows that when the rank is high, a linear network (0 depth) also works well. 
When the rank drops (\eg, when the $\sigma$ of Gau changes from 0.003 to 0.01 to 0.07), the performance of a single linear layer also drops.
Adding one layer to the linear network makes up for the performance drop while adding more layers does not help a lot. In conclusion, 
as the rank drops, a deeper network is needed for better performance.

\begin{figure}[ht]
\captionsetup[subfigure]{labelformat=empty}
\centering 

\begin{subfigure}{0.2\textwidth}
\stackinset{c}{}{t}{-1.0\baselineskip}{\textbf{Ground Truth}}{%
\includegraphics[width=\linewidth]{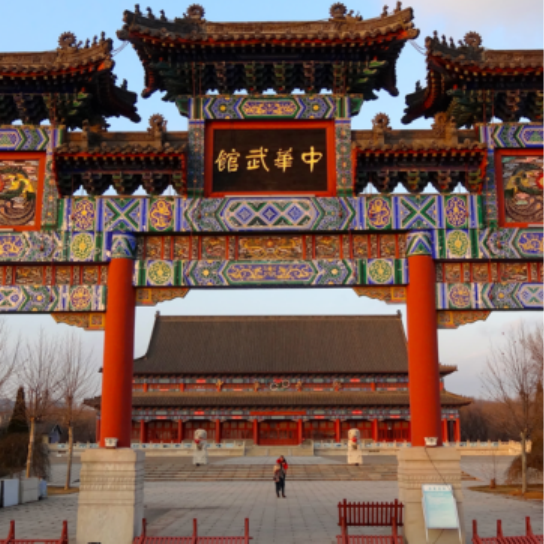}}
\vspace{-0.5\baselineskip}
\end{subfigure}

\begin{subfigure}{0.015\textwidth}\hspace{-2mm}
\rotatebox[origin=c]{90}{\textbf{LinF}}
\end{subfigure}
\begin{subfigure}{0.2\textwidth}
\stackinset{c}{}{t}{-1.0\baselineskip}{\textbf{Simple, Depth 0}}{%
\includegraphics[width=\linewidth]{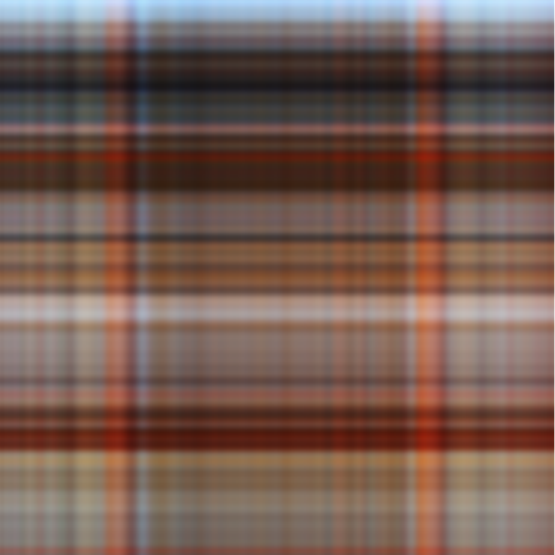}}
\vspace{-1.5\baselineskip}
\caption{PSNR: 13.40}
\end{subfigure} 
\begin{subfigure}{0.2\textwidth}
\stackinset{c}{}{t}{-1.0\baselineskip}{\textbf{Simple, Depth 4}}{%
\includegraphics[width=\linewidth]{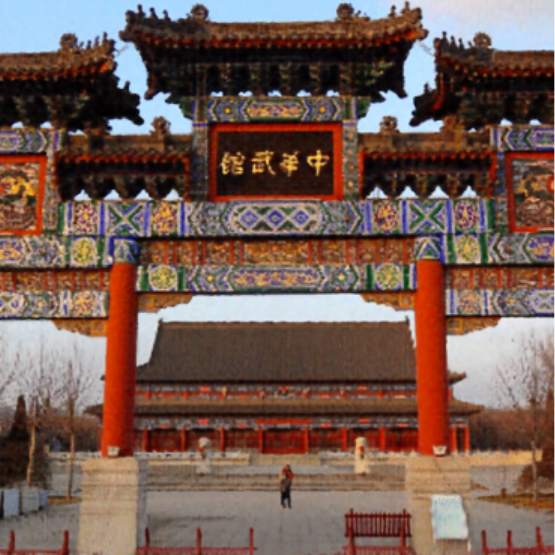} }
\vspace{-1.5\baselineskip}
\caption{PSNR: 22.03}
\end{subfigure}
\begin{subfigure}{0.2\textwidth}
\stackinset{c}{}{t}{-1.0\baselineskip}{\textbf{Complex, Depth 0}}{%
\includegraphics[width=\linewidth]{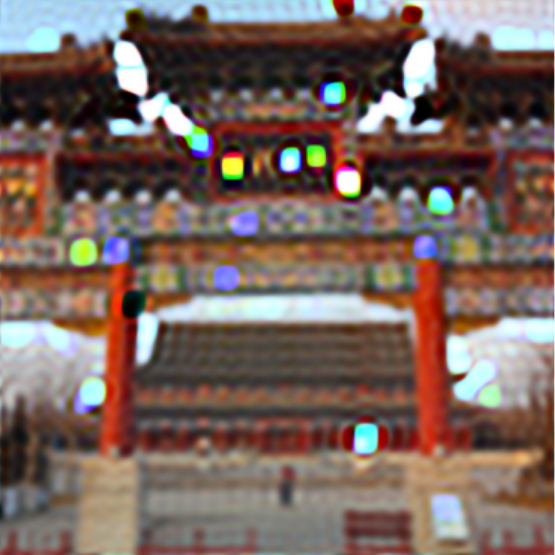}}
\vspace{-1.5\baselineskip}
\caption{PSNR: 17.29}
\end{subfigure}
\begin{subfigure}{0.2\textwidth}
\stackinset{c}{}{t}{-1.0\baselineskip}{\textbf{Complex, Depth 1}}{%
\includegraphics[width=\linewidth]{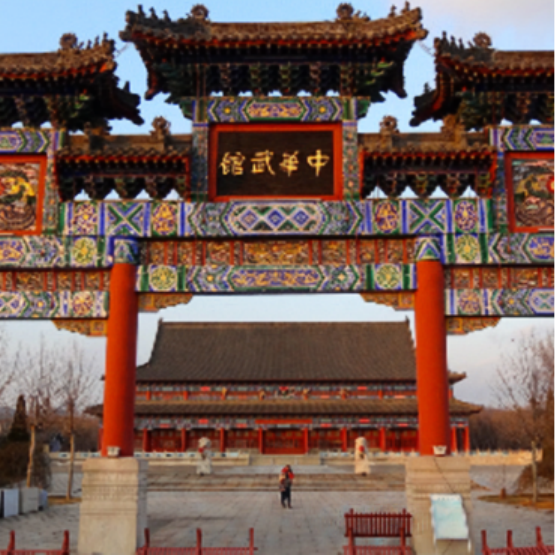}}
\vspace{-1.5\baselineskip}
\caption{PSNR: 22.80}
\end{subfigure}

\begin{subfigure}{0.015\textwidth}\hspace{-2mm}
\rotatebox[origin=c]{90}{\textbf{LogF}}
\end{subfigure}
\begin{subfigure}{0.2\textwidth}
\includegraphics[width=\linewidth]{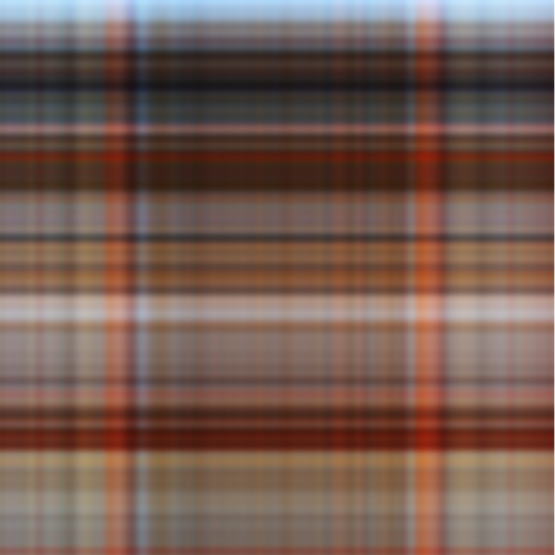} 
\vspace{-1.5\baselineskip}
\caption{PSNR: 13.40}
\end{subfigure} 
\begin{subfigure}{0.2\textwidth}
\includegraphics[width=\linewidth]{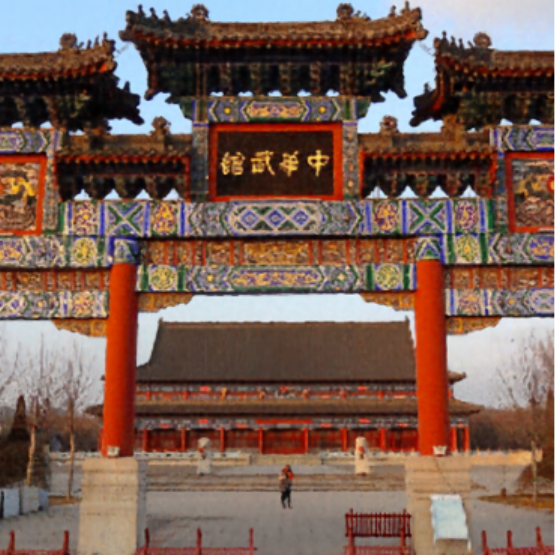} 
\vspace{-1.5\baselineskip}
\caption{PSNR: 21.80}
\end{subfigure} 
\begin{subfigure}{0.2\textwidth}
\includegraphics[width=\linewidth]{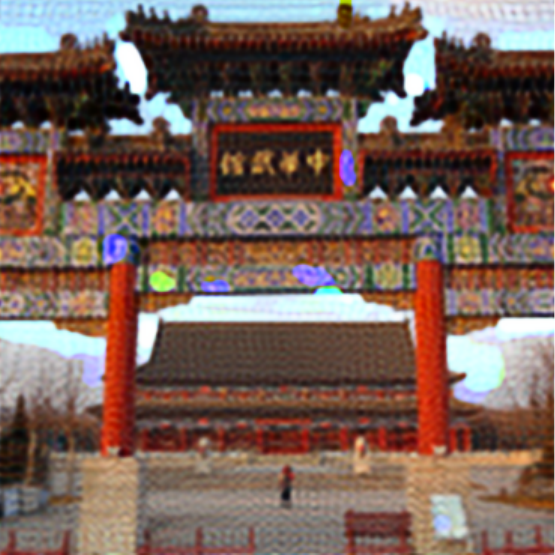} 
\vspace{-1.5\baselineskip}
\caption{PSNR: 19.97}
\end{subfigure}
\begin{subfigure}{0.2\textwidth}
\includegraphics[width=\linewidth]{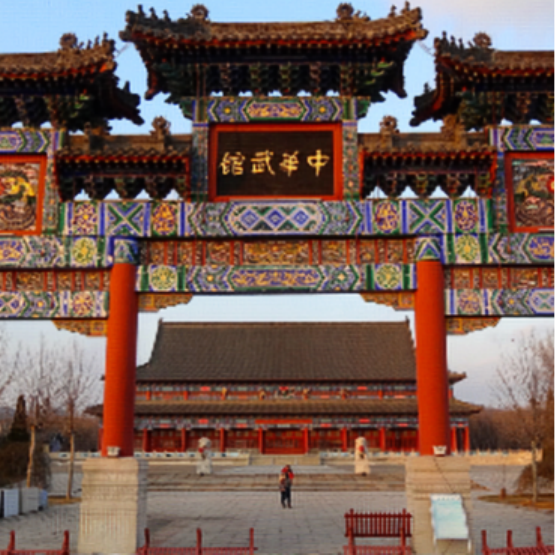} 
\vspace{-1.5\baselineskip}
\caption{PSNR: 22.71}
\end{subfigure}

\begin{subfigure}{0.015\textwidth}\hspace{-2mm}
\rotatebox[origin=c]{90}{\textbf{RFF}}
\end{subfigure}
\begin{subfigure}{0.2\textwidth}
\includegraphics[width=\linewidth]{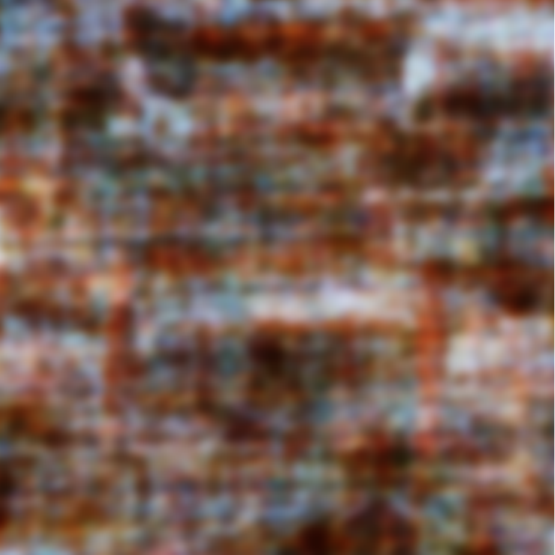} 
\vspace{-1.5\baselineskip}
\caption{PSNR: 12.93}
\end{subfigure}
\begin{subfigure}{0.2\textwidth}
\includegraphics[width=\linewidth]{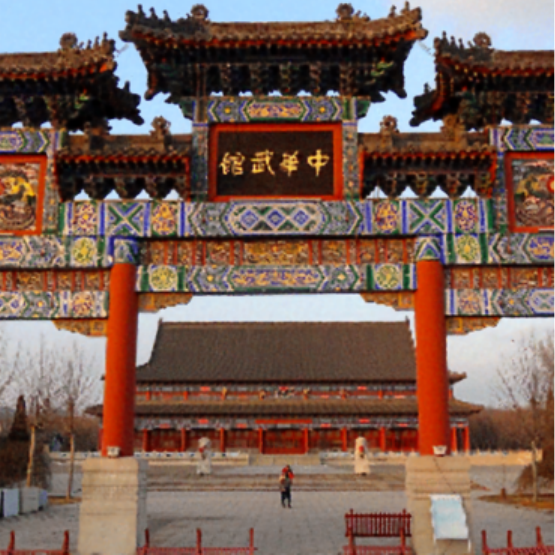} 
\vspace{-1.5\baselineskip}
\caption{PSNR: 22.38}
\end{subfigure} 
\begin{subfigure}{0.2\textwidth}
\includegraphics[width=\linewidth]{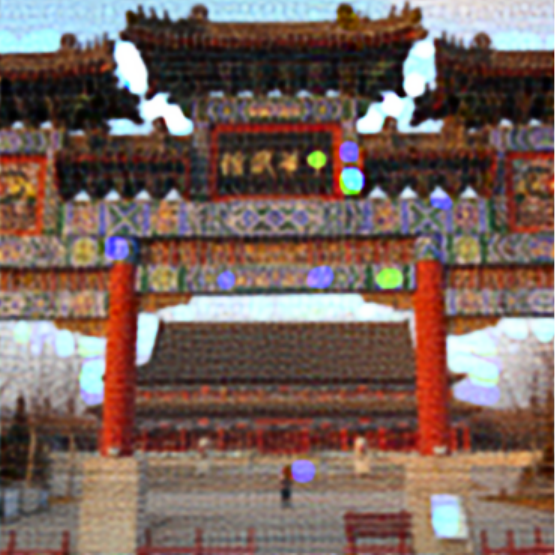} 
\vspace{-1.5\baselineskip}
\caption{PSNR: 19.37}
\end{subfigure}
\begin{subfigure}{0.2\textwidth}
\includegraphics[width=\linewidth]{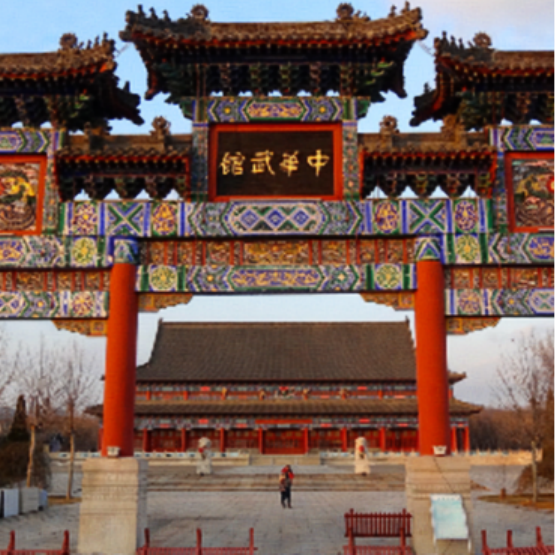} 
\vspace{-1.5\baselineskip}
\caption{PSNR: 22.79}
\end{subfigure}

\begin{subfigure}{0.015\textwidth}\hspace{-2mm}
\rotatebox[origin=c]{90}{\textbf{Tri}}
\end{subfigure}
\begin{subfigure}{0.2\textwidth}
\includegraphics[width=\linewidth]{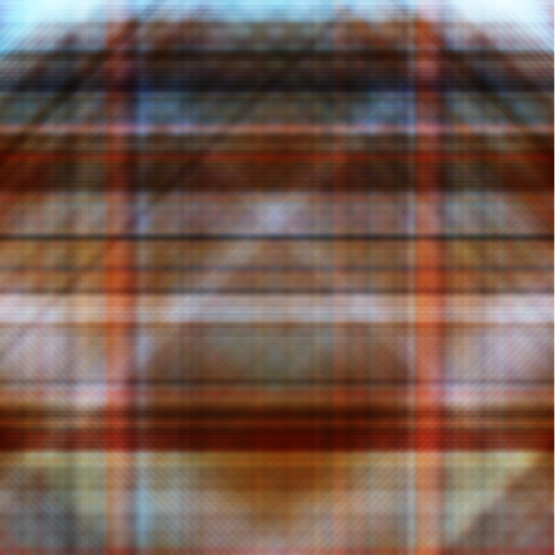} 
\vspace{-1.5\baselineskip}
\caption{PSNR: 14.07}
\end{subfigure} 
\begin{subfigure}{0.2\textwidth}
\includegraphics[width=\linewidth]{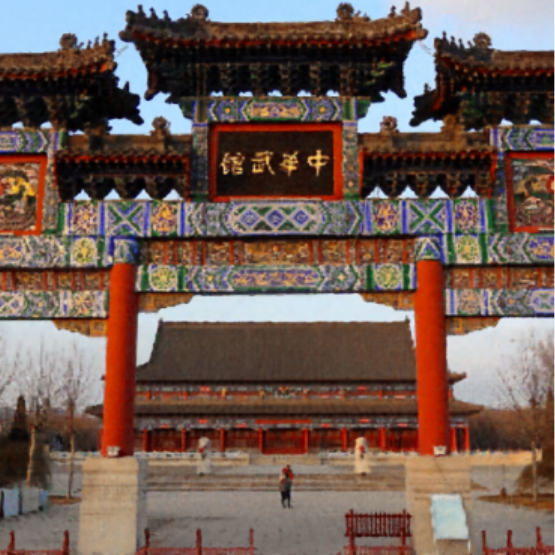} 
\vspace{-1.5\baselineskip}
\caption{PSNR: 21.75}
\end{subfigure}
\begin{subfigure}{0.2\textwidth}
\includegraphics[width=\linewidth]{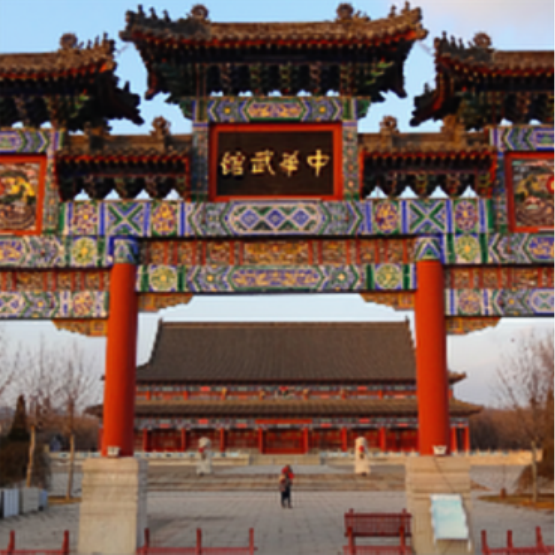} 
\vspace{-1.5\baselineskip}
\caption{PSNR: 22.77}
\end{subfigure} 
\begin{subfigure}{0.2\textwidth}
\includegraphics[width=\linewidth]{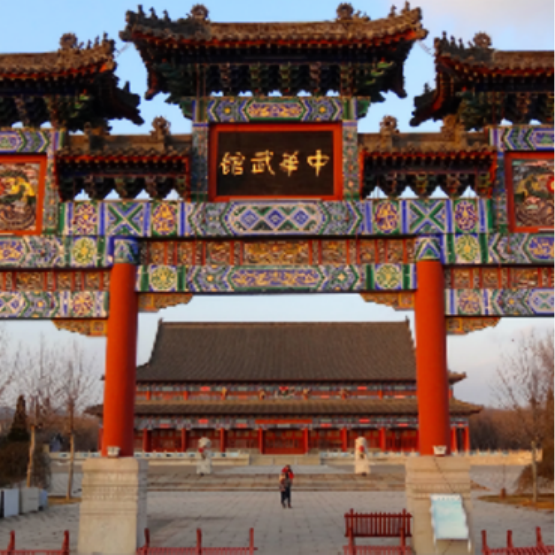} 
\vspace{-1.5\baselineskip}
\caption{PSNR: 23.05}
\end{subfigure}

\begin{subfigure}{0.015\textwidth}\hspace{-2mm}
\rotatebox[origin=c]{90}{\textbf{Gau}}
\end{subfigure}
\begin{subfigure}{0.2\textwidth}
\includegraphics[width=\linewidth]{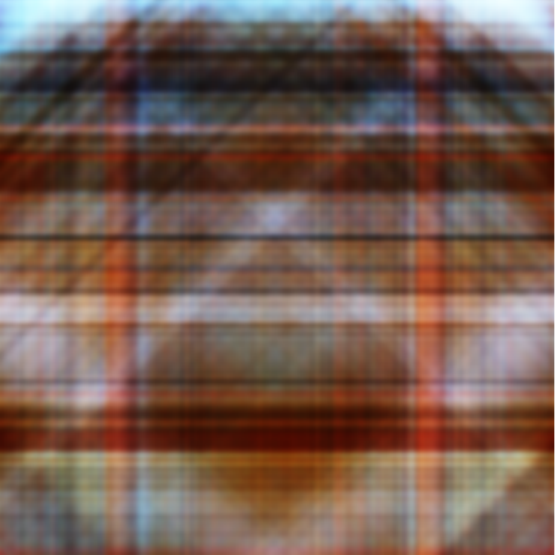}
\vspace{-1.5\baselineskip}
\caption{PSNR: 14.08}
\end{subfigure} 
\begin{subfigure}{0.2\textwidth}
\includegraphics[width=\linewidth]{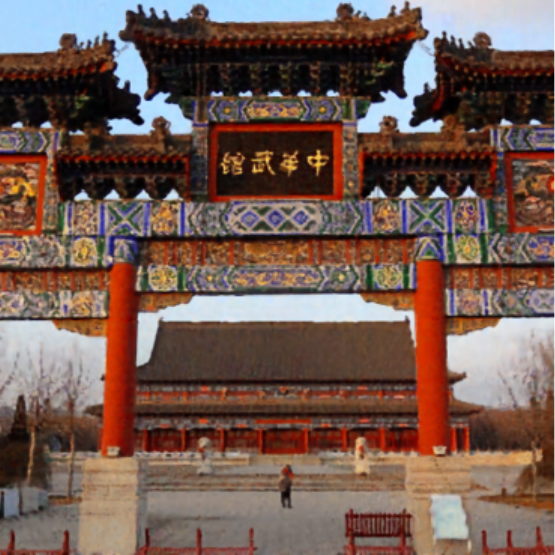} 
\vspace{-1.5\baselineskip}
\caption{PSNR: 21.91}
\end{subfigure}
\begin{subfigure}{0.2\textwidth}
\includegraphics[width=\linewidth]{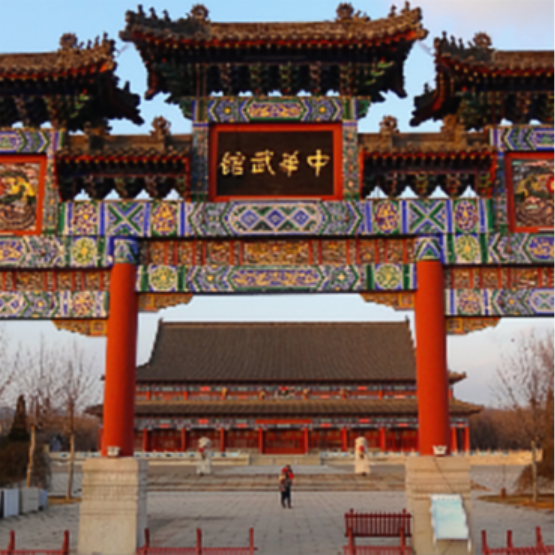}
\vspace{-1.5\baselineskip}
\caption{PSNR: 22.85}
\end{subfigure}
\begin{subfigure}{0.2\textwidth}
\includegraphics[width=\linewidth]{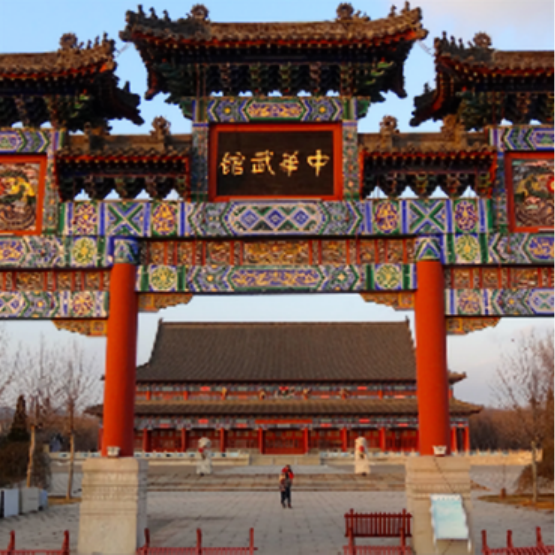}
\vspace{-1.5\baselineskip}
\caption{PSNR: 22.82}
\end{subfigure} 
\vspace{-0.2cm}
\caption{Reconstruction results of an archway using separable coordinates (regular-grid sampled training points) with different combinations of simple or complex encodings and network depths.}
\label{fig:2d_grid_0}
\end{figure}

\subsection{2D: image reconstruction}
For this experiment, we used the image dataset in~\cite{tancik2020fourier}, which contains 32 natural images of size $512{\times}512$. For simple encoding, we used a 4 layer MLP with hidden units of 256 widths, and for complex encoding, we only used a single linear layer. The results discussed below are the average metrics for 16 images where the networks are trained for 2000 epochs.

\noindent\textbf{Separable coordinates.}\:\: $256{\times}256$ grid points were evenly sampled for training and the rest were used for testing. As shown in~\cref{tab:2d-separable}, complex encoding is around $20$ times faster compared to simple encoding. 
In fact, although both the encodings were trained for 2000 epochs, we observed that complex encodings achieved good performance in a significantly lower number of epochs. 
Complex encoding can also be solved in closed-form without training, which is orders of magnitude faster than simple encoding and maintains a good performance. 
Frequency encodings (LinF, LogF, RFF) did not perform well with complex encoding since they were rank deficient.
Although complex frequency encodings did not perform as well as complex shifted encoding when combined with a single linear layer, they still outperformed simple encodings followed by a single linear layer.

\begin{table}[t]
    \setlength{\tabcolsep}{10pt}
    \centering
    \caption[]{Performance of 2D image reconstruction with randomly sampled inputs (non-separable coordinates).
    \tikz\draw[taborange,fill=taborange] (0,0) circle (.5ex); are simple positional encodings. \tikz\draw[tabpurple,fill=tabpurple] (0,0) circle (.5ex); are complex positional encodings with stochastic gradient descent using smart indexing. }
    \vspace{-1em}
    \begin{adjustbox}{width=0.7\linewidth}
    \begin{tabular}{lccccc}
    \toprule
    & \thead{PSNR} & \thead{No. of params (memory)} & \thead{Time (s)}
    \\ \midrule
    \tikz\draw[taborange,fill=taborange] (0,0) circle (.5ex);~LinF 
    & $24.58\pm3.74$ & $329,475 \:(1.32M)$ & $21.97$ \\  
    \tikz\draw[taborange,fill=taborange] (0,0) circle (.5ex);~LogF
    & $24.76\pm3.82$ & $329,475 \:(1.32M)$ & $22.13$ \\ 
    \tikz\draw[taborange,fill=taborange] (0,0) circle (.5ex);~RFF~\cite{tancik2020fourier} 
    & $\mathbf{24.90\pm3.78} $ & $329,475 \:(1.32M)$ & $22.28$ \\ 
    \tikz\draw[taborange,fill=taborange] (0,0) circle (.5ex);~Tri
    & $24.24\pm3.56$ & $329,475 \:(1.32M)$ & $22.50$ \\
    \tikz\draw[taborange,fill=taborange] (0,0) circle (.5ex);~Gau
    & $24.63\pm3.70$ & $329,475 \:(1.32M)$ & $22.86$ \\ 
    \cmidrule{1-4}
    \tikz\draw[tabpurple,fill=tabpurple] (0,0) circle (.5ex);~LinF
    & $19.64\pm2.04$ & $196,608 \:(0.79M)$ & $11.51$ \\
    \tikz\draw[tabpurple,fill=tabpurple] (0,0) circle (.5ex);~LogF
    & $22.35\pm2.82$ & $196,608 \:(0.79M)$ & $11.52$ \\
    \tikz\draw[tabpurple,fill=tabpurple] (0,0) circle (.5ex);~RFF~\cite{tancik2020fourier} 
    & $20.54\pm2.61$ & $196,608 \:(0.79M)$ & $3.08$ \\ 
    \tikz\draw[tabpurple,fill=tabpurple] (0,0) circle (.5ex);~Tri
    & $23.21\pm3.58$ & $196,608 \:(0.79M)$ & $1.90$ \\
    \tikz\draw[tabpurple,fill=tabpurple] (0,0) circle (.5ex);~Gau
    & $\mathbf{23.97\pm3.44} $ & $196,608 \:(0.79M)$ & $2.51$ \\
    \arrayrulecolor{black}\bottomrule
    \end{tabular}
    \end{adjustbox}
    \label{tab:2d-non-separable}
\end{table}

\begin{table}[!ht]
    \setlength{\tabcolsep}{9pt}
    \centering
    \caption[]{Performance of video reconstruction without random sampling (separable coordinates).
    \tikz\draw[taborange,fill=taborange] (0,0) circle (.5ex); are simple positional encodings. \tikz\draw[tabpurple,fill=tabpurple] (0,0) circle (.5ex); are complex positional encodings with gradient descent; \tikz\draw[tabgreen,fill=tabgreen] (0,0) circle (.5ex); are complex positional encodings with closed-form solution. }
     \vspace{-1em}
    \begin{adjustbox}{width=0.7\linewidth}
    \begin{tabular}{lccccc}
    \toprule
    & \thead{PSNR} & \thead{No. of params (memory)} & \thead{Time (s)}
    \\ \midrule
    \tikz\draw[taborange,fill=taborange] (0,0) circle (.5ex);~LinF
    & $21.49\pm3.29$ & $1,445,891 \:(5.78M)$ & $77.06$ \\
    \tikz\draw[taborange,fill=taborange] (0,0) circle (.5ex);~LogF
    & $21.46\pm3.09$    & $1,445,891 \:(5.78M)$ & $77.86$ \\
    \tikz\draw[taborange,fill=taborange] (0,0) circle (.5ex);~RFF~\cite{tancik2020fourier} 
    & $\mathbf{21.06\pm3.26} $    & $1,445,891 \:(5.78M)$ & $77.86$ \\
    \tikz\draw[taborange,fill=taborange] (0,0) circle (.5ex);~Tri
    & $20.86\pm3.09$    & $1,445,891 \:(5.78M)$ $78.91$ \\
    \tikz\draw[taborange,fill=taborange] (0,0) circle (.5ex);~Gau
    & $21.21\pm3.26$ & $1,445,891 \:(5.78M)$ & $79.09$ \\
    \cmidrule{1-4}
    \tikz\draw[tabpurple,fill=tabpurple] (0,0) circle (.5ex);~LinF
    & $9.76\pm3.95$ & $786,432 \:(3.15M)$ & $0.69$ \\
    \tikz\draw[tabpurple,fill=tabpurple] (0,0) circle (.5ex);~LogF
    & $9.98\pm3.84$ & $786,432 \:(3.15M)$ & $0.68$ \\
    \tikz\draw[tabpurple,fill=tabpurple] (0,0) circle (.5ex);~RFF~\cite{tancik2020fourier} 
    & $20.33\pm3.06$ & $786,432 \:(3.15M)$ & $0.69$ \\
    \tikz\draw[tabpurple,fill=tabpurple] (0,0) circle (.5ex);~Tri
    & $22.24\pm2.89$ & $786,432 \:(3.15M)$ & $0.68$ \\
    \tikz\draw[tabpurple,fill=tabpurple] (0,0) circle (.5ex);~Gau
    & $\mathbf{22.11\pm3.14} $ & $786,432 \:(3.15M)$ & $0.66$ \\
    \cmidrule{1-4}
    \tikz\draw[tabgreen,fill=tabgreen] (0,0) circle (.5ex);~RFF~\cite{tancik2020fourier} 
    & $18.05\pm2.36$ & $786,432 \:(3.15M)$ & $0.01$ \\
    \tikz\draw[tabgreen,fill=tabgreen] (0,0) circle (.5ex);~Tri
    & $22.04\pm2.95$ & $786,432 \:(3.15M)$ & $0.009$ \\
    \tikz\draw[tabgreen,fill=tabgreen] (0,0) circle (.5ex);~Gau
    & $\mathbf{21.83\pm3.33} $ & $786,432 \:(3.15M)$ & $0.008$ \\
    \arrayrulecolor{black}\bottomrule
    \end{tabular}
    \end{adjustbox}
    \label{tab:3d-separable}
\end{table}

\noindent\textbf{Non-separable coordinates.}\:\: For non-separable coordinates, the training points were randomly sampled $25\%$ of the $512{\times}512$ natural images. Complex encoding contains a blending matrix representing virtual separable coordinates, which can be pre-computed. 
As illustrated in~\cref{tab:2d-non-separable}, the performance of complex encodings was comparably well and converged faster than the simple encodings.

\vspace{-1em}
\subsection{3D: video reconstruction}
As dimensionality of data increases, the faster convergence of the complex encoding becomes more notable. We use the Youtube video~\cite{esteban2017youtube} for our experiments. 
In our experiments, we extracted 256 frames from 5 different videos and rescaled each frame to  $256{\times}256$.
Then, a central $128{\times}128{\times}128$ cube was cropped to create our dataset, since some videos contain borders. 
For simple encoding, a 5 layer MLP with  512-width was used, while for complex encoding, only a single linear layer was used. 
We trained the networks for 500 epochs. 
The training points were regularly sampled from a $64{\times}64{\times}64$ grid and the rest of the points were used for testing. The quantitative results are shown in~\cref{tab:3d-separable}. Qualitative results are illustrated in Fig.\ref{fig:2d_grid_0}.

\section{Conclusion}
 \vspace{-1em}
In this paper, we show that the performance of a positional encoding scheme is mainly governed by the stable rank of the embedding matrix and the distance preservation between the embedded coordinates. In light of this discovery, we propose a novel framework that can incorporate arbitrary continuous signals as potential embedders, under certain constraints. We also propose a positional encoding scheme that enables dramatically faster convergence, allowing a single linear network to encode signals with high fidelity.

\section*{Acknowledgments}
This research was supported by the Australian Research Council under Discovery Project DP220103803.

\newpage
\appendix
\onecolumn
\section*{\Large Appendix}

\setcounter{prop}{0}

\section{Theoretical results}
\label{supp:theoretical_rank}
\begin{prop}
Consider a set of coordinates $\x {=} [x_1, x_2, {\cdots}, x_N]^T$, corresponding outputs $\y {=} [y_1, y_2, \cdots, y_N]^T$, and a $d$ dimensional embedding $\Psi{:}\mathbb{R} {\to} \mathbb{R}^d$. Assuming perfect convergence, the necessary and sufficient condition for a linear model to perfect memorize of the mapping between $\x$ and $\y$ is for $\X {=} [\Psi(x_1), \Psi(x_2), {\dots}, \Psi(x_N)]$ to have full rank.
\end{prop}

\noindent\textbf{Proof:} Let us refer to the row vectors of $\X$ as 
$[\p_{1},{\ldots},\p_{d}]^{T}$. In order to perfectly reconstruct $\y$ using a linear learner with weights $\mathbf{w} {=} [w_1, w_2, {\dots}, w_d]$ as 
\begin{equation}
    \y = \sum_{i=1}^{d} w_{i} \p_{i} + b\:,
\end{equation}
one needs~$\X$~to be of rank~$N$ (since $\y$ needs to completely
span $\{ \p_{i} \}_{i=1}^{d}$). If~$d > N$ then there is no unique
solution to~$\{\mathbf{w}, b\}$ without some regularization. In the unlikely
scenario that the row vectors of $\X$ have zero mean, then $\X$ needs
to be of rank $N-1$ since the bias term~$b$ can account for that
missing linear basis. \qed

\begin{prop}
Let the Gaussian embedder be denoted as $\psi(t,x) {=} \exp \left({-}\frac{\|t{-}x\|^2}{2\sigma^2}\right)$. With a sufficient embedding dimension, the stable rank of the embedding matrix obtained using the Gaussian embedder is $\min\left(N, \frac{1}{2\sqrt{\pi}\sigma}\right)$ where $N$ is the number of embedded coordinates. Under the same conditions, the embedded distance between two coordinates $x_1$ and $x_2$ is $\mathrm{D}(x_1,x_2) {=} \exp \left({-}\frac{\|x_1{-}x_2\|^2}{4\sigma^2}\right)$.
\end{prop}

\noindent\textbf{Proof:} Let us define the Gaussian embedder as $\psi(t,x) {=} \exp \left({-}\frac{\|t{-}x\|^2}{2\sigma^2}\right)$, where $\sigma$ is the standard deviation. Given $d$ samples points  $[t_{1},{\ldots},t_{d}]$ and $N$ input coordinates $[x_{1},{\ldots},x_{N}]$, the elements of the embedding matrix are
\begin{equation}
    \Psi_{i,j} = \psi(t_i,x_j)\:.
\end{equation}

To make sure the stable rank is saturated, we assume that $d$ and $N$ is large enough. Then, $\Psi$ is approximately a circulant matrix. We know that the singular value decomposition of a circulant matrix $C$, whose first row is $c$, can be written as
\begin{equation}
    C =\frac{1}{n}F_n^{-1} diag\left(F_n c\right)F_n \:,
\end{equation}
where $F_n$ is the Fourier transform matrix. This means the singular values of a circulant matrix is the Fourier transform of first row. When $N$ is large enough, we can approximate the first row of $\Psi$ as  a continuous signal, which is $\psi(x, t{=}0) {=} \exp \left({-}\frac{\|x\|^2}{2\sigma^2}\right)$, so the singular values are
\begin{equation}
    s(\xi) =\mathcal{F}\left(\psi(x;t=0)\right) = \sqrt{2\pi}\sigma\exp\left(-2\sigma^2\|\pi\xi\|^2\right) \:.
\end{equation}
Therefore, we can calculate stable rank directly from the definition,
\startcompact{small}
\begin{equation}
    \textrm{Stable Rank}(\Psi) {=} \sum_{i=1}^N{\frac{s_i^2}{s_1^2}} {=} \int_{{-}\infty}^{{+}\infty}\frac{s^2(\xi)}{s^2(0)}d\xi=\int_{-\infty}^{+\infty}\exp\left({-}4\sigma^2\|\pi\xi\|^2\right)d\xi{=}\frac{1}{2\sqrt{\pi}\sigma} \:.
\end{equation}
\stopcompact{small}
Considering the general case, where $N$ might not be large enough, the stable rank will be $\min\left(N, \frac{1}{2\sqrt{\pi}\sigma}\right)$.

The distance (or similarity) between two embedded coordinates can be obtained via the inner product:
\begin{equation}
\begin{aligned}
    \mathrm{D}(x_1,x_2) &= \int_{-\infty}^{+\infty} \psi(t,x_1)\psi(t,x_2)dt
    \\ &=\int_{-\infty}^{+\infty} e^{-\frac{(t-x_1)^2}{2\sigma^2}}e^{-\frac{(t-x_2)^2}{2\sigma^2}}dt
    \\ &=\int_{-\infty}^{+\infty} e^{-\frac{(t-x_1)^2+(t-x_2)^2}{2\sigma^2}}dt
    \\ &=\int_{-\infty}^{+\infty} e^{-\frac{t^2-2x_1t+x_1^2+t^2-2x_2t+x_2^2}{2\sigma^2}}dt
    \\ &=\int_{-\infty}^{+\infty} e^{-\frac{2t^2-2(x_1+x_2)t+\frac{(x_1+x_2)^2}{2}+\frac{(x_1-x_2)^2}{2}}{2\sigma^2}}dt
    \\ &=\int_{-\infty}^{+\infty} e^{-\frac{(t-\frac{x_1+x_2}{2})^2}{\sigma^2}}e^{-\frac{(x_1-x_2)^2}{4\sigma^2}}dt
    \\ &=e^{-\frac{(x_1-x_2)^2}{4\sigma^2}}\int_{-\infty}^{+\infty} e^{-\frac{(t-\frac{x_1+x_2}{2})^2}{\sigma^2}}dt
    \\ &=\sqrt{\pi}\sigma e^{-\frac{(x_1-x_2)^2}{4\sigma^2}}\:.
\end{aligned}
\end{equation}
which is also a Gaussian with a standard deviation ${\sqrt{2}\sigma}$. We can empirically define that the distance between two embedded coordinates $x_1$ and $x_2$ is preserved if $D(x_1,x_2)\ge{10^{-k}}$, for an interval $x_1{-}x_2{\leq} l$, where $k$ is a threshold. In the Gaussian embedder, we can analytically obtain a $\sigma$ for an arbitrary $l$ using the relationship $\sigma {=} \frac{l}{2\sqrt{k{\ln{10}}}}$. \qed

\begin{prop}
Let the RFF embedding be denoted as $\gamma(x){=}[\cos2\pi{\mathbf{b}x},\sin{2\pi\mathbf{b}x}]$, where $\mathbf{b}$ are sampled from a Gaussian distribution. When the embedding dimension is large enough, the stable rank of RFF will be $\min\left(N, \sqrt{2\pi}\sigma\right)$, where $N$ is the numnber of embedded coordinates. Under the same conditions, the embedded distance between two coordinates $x_1$ and $x_2$ is $\mathrm{D}(x_1,x_2) {=} \sum_j \cos2\pi{b_j(x_1{-}x_2)}$.
\end{prop}

\noindent\textbf{Proof:} Given $\frac{d}{2}$ samples for $\mathbf{b}$ as $[b_{1},{\ldots},b_{\frac{d}{2}}]$ from a Gaussian distribution with a standard deviation $\sigma$ and $N$ input coordinates $[x_{1},{\ldots},x_{N}]$,  RFF embedding is defined as $\gamma(x){=}[\cos{2\pi\mathbf{b}x_i},\sin{2\pi\mathbf{b}x_i}]$.

To make sure the stable rank is saturated, we assume that the $d$ and $N$ is large enough. Although RFF embedding matrix is not circulant, it is naturally frequency based so we already know its spectrum, which is its singular value distribution
\begin{equation}
    s(\xi) = \frac{1}{\sqrt{2\pi}\sigma}\exp\left(-\frac{\xi^2}{2\sigma^2}\right)\:.
\end{equation}
Similarly,
\startcompact{small}
\begin{equation}
    \textrm{Stable Rank}(\gamma) = \sum_{i=1}^N{\frac{s_i^2}{s_1^2}} = \int_{-\infty}^{+\infty}\frac{s^2(\xi)}{s^2(0)}d\xi=\int_{-\infty}^{+\infty}\exp\left(-\frac{\xi^2}{2\sigma^2}\right)d\xi=\sqrt{2\pi}\sigma\:,
\end{equation}
\stopcompact{small}
Considering the general case, the stable rank is $\min\left(N, \sqrt{2\pi}\sigma\right)$.

From the basic trigonometry, it can be easily deduced the distance function that $\mathrm{D}(x_1,x_2) {=} \sum_j \cos{2\pi b_j(x_1{-}x_2)}$. 
When $d$ is extremely large it can be considered as $f(\xi) {=} \cos{2\pi\xi(x_1{-}x_2)}$ where $\xi$ is a Gaussian random variable with standard deviation $\sigma$. Then the above sum can be replaced with the integral,
\begin{equation}
\begin{aligned}
    \mathrm{D}(x_1,x_2) &= \int_{-\infty}^{+\infty} e^{-\frac{\xi^2}{2\sigma^2}}\cos{2\pi\xi(x_1-x_2)}d\xi
    \\ &= 2\int_{0}^{+\infty} e^{-\frac{\xi^2}{2\sigma^2}}\cos{2\pi\xi(x_1-x_2)}d\xi
    \\ &= 2\int_{0}^{+\infty} e^{-\frac{\xi^2}{2\sigma^2}}\frac{1}{2}(e^{i2\pi(x_1-x_2)\xi}+e^{-i2\pi(x_1-x_2)\xi})d\xi
    \\ &= \int_{0}^{+\infty} e^{-\frac{\xi^2}{2\sigma^2}+i2\pi(x_1-x_2)\xi}+e^{-\frac{\xi^2}{2\sigma^2}-i2\pi(x_1-x_2)\xi}d\xi\:.
\end{aligned}
\end{equation}
Further,
\startcompact{small}
\begin{equation}
    {\int_{0}^{{+\infty}}} e^{-ax^2+bx}dx \:{=}\: e^{{-}\frac{b^2}{4a}}\int_{0}^{{+}\infty} e^{-a(x-i\frac{b}{2a})^2}dx \:{=}\: \frac{1}{2}\left(1{+}\mathrm{erfi}\left(\frac{b}{2\sqrt{a}}\right)\right)\sqrt{\frac{\pi}{a}}e^{{-}\frac{b^2}{4a}}\:.
\end{equation}
\stopcompact{small}
Let $a {=} \frac{1}{2\sigma^2}$ and $b{=}\pm2\pi(x_1{-}x_2)$. Then, we have
\begin{equation}
    \mathrm{D}(x_1,x_2)=\sqrt{2\pi}\sigma e^{-2\pi^2\sigma^2(x_1-x_2)^2}\:.
\end{equation}\qed

\begin{prop}
Let the Rectangular embedder be denoted as $\psi(t,x) {=} \mbox{rect}\left(\frac{x{-}t}{\sigma}\right) {=} (1-\frac{|x{-}t|}{0.5\sigma}){>}0$. With a sufficient embedding dimension, the stable rank of the embedding matrix obtained using the Rectangular embedder is $\min\left(N, \frac{1}{\sigma}\right)$ where $N$ is the number of embedded coordinates. Under the same conditions, the embedded distance between two coordinates $x_1$ and $x_2$ is $\mathrm{D}\left(x_1,x_2\right) {=} \sigma\mbox{tri}\left(\frac{|x_1-x_2|}{\sigma}\right) {=} \sigma\max(1{-}\frac{|x_1{-}x_2|}{\sigma},0)$.
\end{prop}

\noindent\textbf{Proof:} Let us define the Rectabgular embedder as $\psi(t,x) {=} \mbox{rect}\left(\frac{x{-}t}{\sigma}\right) {=} \left(1{-}\frac{|x{-}t|}{0.5\sigma}\right){>}0$, where $\sigma$ is the width of the rectangle impulse. Given $d$ samples points  $[t_{1},{\ldots},t_{d}]$ and $N$ input coordinates $[x_{1},{\ldots},x_{N}]$, the elements of the embedding matrix are
\begin{equation}
    \Psi_{i,j} = \psi(t_i,x_j)\:.
\end{equation}

To make sure the stable rank is saturated, we assume that $d$ and $N$ are large enough. Then, $\Psi$ is approximately a circulant matrix. We know that the singular value decomposition of a circulant matrix $C$, whose first row is $c$, can be written as
\begin{equation}
    C =\frac{1}{n}F_n^{-1} diag\left(F_n c\right)F_n\:,
\end{equation}
where $F_n$ is the Fourier transform matrix. This means the singular values of a circulant matrix are the Fourier transform of the first row. When $N$ is large enough, we can approximate the first row of $\Psi$ as a continuous signal, which is $\psi(x, t{=}0) {=} \mbox{rect}(\frac{x}{\sigma})$, so the singular values are 
\begin{equation}
    s(\xi) =\mathcal{F}\left(\psi(x;t=0)\right) = \sigma \mbox{sinc}(\sigma\xi)\:,
\end{equation}
where $\mbox{sinc}(\xi){=}\frac{\sin \left(\pi x\right)}{\pi x}$. Therefore, we can compute the stable rank directly from the definition,
\begin{equation}
    \textrm{Stable Rank}(\Psi) = \sum_{i=1}^N{\frac{s_i^2}{s_1^2}} = \int_{-\infty}^{+\infty}\frac{s(\xi)^2}{s(0)^2}d\xi=\int_{-\infty}^{+\infty}\mbox{sinc}^2(\sigma\xi)d\xi=\frac{1}{\sigma}\:.
\end{equation}
Considering the general case, where $N$ might not be large enough, the stable rank will be $\min\left(N, \frac{1}{\sigma}\right)$.

The distance (or similarity) between two embedded coordinates can be obtained via the inner product:
\begin{equation}
\begin{aligned}
    \mathrm{D}(x_1,x_2) &= \int_{-\infty}^{+\infty} \psi(t,x_1)\psi(t,x_2)dt
    \\ &=\int_{-\infty}^{+\infty} \mbox{rect}\left(\frac{x_1-t}{\sigma}\right)\mbox{rect}\left(\frac{x_2-t}{\sigma}\right)dt
    \\ &=\sigma\mbox{tri}\left(\frac{x_1-x_2}{\sigma}\right)\:.
\end{aligned}
\end{equation}
\qed

\begin{prop}
Let the Triangular embedder be $\psi(t,x) {=} \mbox{tri}\left(\frac{x{-}t}{0.5\sigma}\right) {=} \max (1{-}\frac{|x{-}t|}{0.5\sigma},0)$. With a sufficient embedding dimension, the stable rank of the embedding matrix obtained using the Triangular embedder is $\min(N, \frac{4}{3\sigma})$ where $N$ is the number of embedded coordinates. Under the same conditions, the embedded distance between two coordinates $x_1$ and $x_2$ is $\mathrm{D}(x_1,x_2) = \frac{1}{4}\sigma^2\mbox{tri}^2(\frac{|x_1-x_2|}{\sigma}) = \frac{1}{4}\sigma^2\max(1-\frac{|x_1-x_2|}{\sigma},0)^2$.
\end{prop}

\noindent\textbf{Proof:} Let us define the Triangle embedder as $\psi(t,x) {=} \mbox{tri}\left(\frac{x{-}t}{0.5\sigma}\right) {=} \max \left(1{-}\frac{|x{-}t|}{0.5\sigma},0\right)$, where $\sigma$ is the width of the Triangular impulse. Given $d$ samples points $[t_{1},{\ldots},t_{d}]$ and $N$ input coordinates $[x_{1},{\ldots},x_{N}]$, the elements of the embedding matrix are
\begin{equation}
    \Psi_{i,j} = \psi(t_i,x_j)\:.
\end{equation}

To make sure the stable rank is saturated, we assume that $d$ and $N$ are large enough. Then, $\Psi$ is approximately a circulant matrix. We know that the singular value decomposition of a circulant matrix $C$, whose first row is $c$, can be written as
\begin{equation}
    C =\frac{1}{n}F_n^{-1} diag\left(F_n c\right)F_n\:,
\end{equation}
where $F_n$ is the Fourier transform matrix. This means the singular values of a circulant matrix are the Fourier transform of the first row. When $N$ is large enough, we can approximate the first row of $\Psi$ as a continuous signal, which is $\psi(x, t{=}0) {=} \mbox{tri}\left(\frac{x}{\sigma}\right)$, so the singular values are 
\begin{equation}
    s(\xi) =\mathcal{F}\left(\psi(x;t=0)\right) = \frac{\sigma}{2} \mbox{sinc}^2\left(\frac{\sigma}{2}\xi\right)\:,
\end{equation}
where $\mbox{sinc}(\xi){=}\frac{\sin \left(\pi x\right)}{\pi x}$. Therefore, we can compute stable rank directly from the definition as,
\begin{equation}
    \textrm{Stable Rank}(\Psi) = \sum_{i=1}^N{\frac{s_i^2}{s_1^2}} = \int_{-\infty}^{+\infty}\frac{s(\xi)}{s(0)}^2d\xi=\int_{-\infty}^{+\infty} \mbox{sinc}^4\left(\frac{\sigma}{2}\xi\right)d\xi=\frac{4}{3\sigma}\:.
\end{equation}
Considering the general case, where $N$ might not be large enough, the stable rank will be $\min\left(N, \frac{1}{\sigma}\right)$.

The distance (or similarity) between two embedded coordinates can be obtained via the inner product:
\begin{equation}
\begin{aligned}
    \mathrm{D}(x_1,x_2) &= \int_{-\infty}^{+\infty} \psi(t,x_1)\psi(t,x_2)dt
    \\ &=\int_{-\infty}^{+\infty} \mbox{tri}\left(\frac{x-t}{0.5\sigma}\right)\mbox{tri}\left(\frac{x-t}{0.5\sigma}\right)dt
    \\ &=\frac{1}{4}\sigma^2\max\left(1-\frac{|x_1-x_2|}{\sigma},0\right)^2\:.
\end{aligned}
\end{equation}
\qed

\section{2D complex encoding}
\label{supp:2d_complex_encoding}
\subsection{Closed form solution for separable coordinates}
\label{supp:closed_form}
If pixels are sampled on a regular grid formed by samples $\x {=} [x_1,x_2,{\cdots},x_N]^T$ and samples $\y {=} [y_1,y_2,{\cdots},y_M]^T$, then the coordinates of these pixels are separable. 
Let $\mathbf{S} {\in} \mathbb{R}^{M {\times} N}$ be the signal defined as $\mathbf{S}_{i,j}{=}I(x_i,y_j)$, where $i{=}1,2,{\cdots},N$, $j{=}1,2,{\cdots},M$, and $\Psi{:}\mathbb{R}{\to}\mathbb{R}^{K}$ be the 1D encoder. We want to find the weights $\W{\in}\mathbb{R}^{K{\times}K}$ of the linear layer by optimizing the following equation,
\begin{equation}
    \arg\min_{\mathbf{W}} \left\| \mbox{vec}(\mathbf{S}) {-} \left(\Psi(\y)\otimes\Psi(\x)\right)\mbox{vec}\left(\mathbf{W}\right) \right\|_2^2\:,
\end{equation}
where $\Psi(\x){\in}\mathbb{R}^{N{\times}K}$ is the encoding for $\x$, $\Psi(\y){\in}\mathbb{R}^{M{\times}K}$ is the encoding for $\y$. 
This is a linear least squares problem. Based on the properties of the Kronecker product, we find the optimal solution $\mathbf{W}^*$ as,
\begin{equation}
\begin{aligned}
    \mbox{vec}\left(\mathbf{W}^{\ast}\right) &= \arg\min_{\mathbf{W}} \left\| \mbox{vec}\left(\mathbf{S}\right) {-} \left(\Psi(\y)\otimes\Psi(\x)\right)\mbox{vec}\left(\mathbf{W}\right) \right\|_2^2 \\
    &= \left(\left(\Psi(\y)\otimes\Psi(\x)\right)^{T}\left(\Psi(\y)\otimes\Psi(\x)\right)\right)^{-1}\left(\Psi(\y)\otimes\Psi(\x)\right)^{T}\mbox{vec}\left(\mathbf{S}\right) \\
    &= \left(\left(\left(\Psi(\y)^{T}\Psi(\y)\right)^{-1}\Psi(\y)\right)\otimes\left(\left(\Psi(\x)^{T}\Psi(\x)\right)^{-1}\Psi(\x)\right)\right)\mbox{vec}\left(\mathbf{S}\right) \\
    &= \mbox{vec}\left(\left(\left(\Psi(\x)^{T}\Psi(\x)\right)^{-1}\Psi(\x)\right)\mathbf{S}\left(\left(\Psi(\y)^{T}\Psi(\y)\right)^{-1}\Psi(\y)\right)^{T}\right) \\
    &= \mbox{vec}\left(\left(\Psi(\x)^{T}\Psi(\x)\right)^{-1}\Psi(\x)\mathbf{S}\Psi(\y)^{T}\left(\Psi(\y)^{T}\Psi(\y)\right)^{-1}\right)\:,
\end{aligned} 
\end{equation}
which means, 
\begin{equation}
    \mathbf{W}^{\ast} =  \left(\Psi(\x)^{T}\Psi(\x)\right)^{-1}\Psi(\x)\mathbf{S}\Psi(\y)^{T}\left(\Psi(\y)^{T}\Psi(\y)\right)^{-1}\:.
\end{equation}

\subsection{Blending matrix for non-separable coordinates}
\label{supp:blending_matrix}
First, we focus on 1D encoders. 
Given a 1D encoder $\Psi{:}\mathbb{R}{\to}\mathbb{R}^{K}$ and two points $x_0$, $x_1$, we want to express $\Psi(x){\approx} \alpha_0\Psi(x_0){+}\alpha_1\Psi(x_1)$ for $x_0{\leq} x{\leq} x_1$. This problem can be solved by
\begin{equation}
    \arg\min_{\mathbf{\alpha}} \left\|\Psi(x)- \begin{bmatrix}\Psi(x_0)&\Psi(x_1)\end{bmatrix}\mathbf{\alpha} \right\|_2^2\:,
\end{equation}
where $\mathbf{\alpha}{=}{\begin{bmatrix}\alpha_0&\alpha_1\end{bmatrix}}^T$. Note here that $\Psi(x)$, $\Psi(x_0)$, and $\Psi(x_1)$ are $K {\times} 1$ vectors. This is equivalent to a least squared problem, thus, the optimal solution $\mathbf{\alpha}^{\ast}$ can be solved by,
\begin{equation}
\begin{aligned}
    \mathbf{\alpha}^{\ast} &= \arg\min_{\mathbf{\alpha}} \left\|\Psi(x)- \begin{bmatrix}\Psi(x_0)&\Psi(x_1)\end{bmatrix}\mathbf{\alpha} \right\|_2^2 \\
    &= \left(\begin{bmatrix}\Psi(x_0)&\Psi(x_1)\end{bmatrix}^{T}\begin{bmatrix}\Psi(x_0)&\Psi(x_1)\end{bmatrix}\right)^{-1}\begin{bmatrix}\Psi(x_0)&\Psi(x_1)\end{bmatrix}^{T}\Psi(x) \\
    &= \left(\begin{bmatrix}\Psi(x_0)^{T} \\ \Psi(x_1)^{T}\end{bmatrix} \begin{bmatrix}\Psi(x_0)&\Psi(x_1)\end{bmatrix}\right)^{-1}\begin{bmatrix}\Psi(x_0)^{T} \\ \Psi(x_1)^{T}\end{bmatrix}\Psi(x) \\
    &= \begin{bmatrix}\Psi(x_0)^{T}\Psi(x_0)\:\: & \:\:\Psi(x_0)^{T}\Psi(x_1) \\ \Psi(x_1)^{T}\Psi(x_0)\:\: & \:\:\Psi(x_1)^{T}\Psi(x_1) \end{bmatrix}^{-1}\begin{bmatrix}\Psi(x_0)^{T}\Psi(x) \\ \Psi(x_1)^{T}\Psi(x)\end{bmatrix} \:.
\end{aligned} 
\end{equation}
With the definition $D(x_1,x_2)$ in~\cref{supp:theoretical_rank}, this can be written as,
\begin{equation}
    \mathbf{\alpha}^{\ast} = \begin{bmatrix}D(x_0,x_0)\:\: & \:\:D(x_0,x_1) \\ D(x_1,x_0)\:\: & \:\:D(x_1,x_1) \end{bmatrix}^{-1}\begin{bmatrix}D(x_0,x) \\ D(x_1,x)\end{bmatrix} \:.
\end{equation}

Typically, this distance function only depends on the difference of the inputs, as examples shown in~\cref{supp:theoretical_rank}. Therefore, we can have a close form solution for $D{:}\mathbb{R}{\to}\mathbb{R}$. Let $d {=} x_1{-}x_0$, and $x {=} x_0 {+} \beta d$, where $0{\leq}\beta{\leq}1$.  Then, the solution becomes,
\begin{equation}
\begin{aligned}
    \mathbf{\alpha}^{\ast} & = \begin{bmatrix}D(x_0,x_0)\:\: & \:\:D(x_0,x_1) \\ D(x_1,x_0)\:\: & \:\:D(x_1,x_1) \end{bmatrix}^{-1}\begin{bmatrix}D(x_0,x) \\ D(x_1,x)\end{bmatrix} \\
    & = \begin{bmatrix}D(0) & D(d) \\ D(d) & D(0) \end{bmatrix}^{-1}\begin{bmatrix}D(\beta d) \\ D\left(\left(1-\beta\right) d\right)\end{bmatrix} \\
    & = \frac{1}{D^2(0)-D^2(d)}\begin{bmatrix}D(0)&{-}D(d)\\{-}D(d)&D(0)\end{bmatrix}\begin{bmatrix}D(\beta d)\\D\left(\left(1-\beta\right) d\right)\end{bmatrix}
    \:.
\end{aligned}
\end{equation}

Based on the 1D analysis, encoding 2D non-separable points can also be expressed as non-linear interpolation of 2D separable coordinates. Suppose that the settings are the same as in~\cref{supp:closed_form}. The virtual pixels are sampled on a regular grid formed by samples $\x {=} [x_1,x_2,{\cdots},x_N]^T$ and samples $\y {=} [y_1,y_2,{\cdots},y_M]^T$. The query points are randomly sampled in the space as $\mathbf{Q} = [\mathbf{q}_1,\mathbf{q}_2,{\cdots} ,\mathbf{q}_P]^T$, where $P$ is the number of points and each $\mathbf{q}_i{\in} \mathbb{R}^{2\times 1}$ is a random 2D coordinate. Let $\mathbf{s}{\in} \mathbb{R}^{P\times 1}$ be the signal, and $\Psi{:}\mathbb{R}{\to}\mathbb{R}^{K}$ be the 1D encoder. We want to find the weights $\W{\in}\mathbb{R}^{K{\times}K}$ of the linear layer by optimizing the following equation,
\begin{equation}
    \arg\min_{\mathbf{W}} \left\| \mathbf{s} {-} B(\mathbf{Q})\left(\Psi(\y)\otimes\Psi(\x)\right)\mbox{vec}\left(\mathbf{W}\right) \right\|_2^2\:,
\end{equation}
where $B{:}\mathbb{R}^{2}{\to}\mathbb{R}^{MN}$ is the non-linear interpolation coefficients function,~\ie,  $B(\mathbf{Q})\in \mathbb{R}^{P{\times} MN}$ is the blending matrix. Note that although $B$ is large,  it is extremely sparse and only have $4$ non-zero values on each row of $MN$ elements. Consider a certain point $\mathbf{q}_p$ is in the grid whose corner points are $(x_{i},y_{j})$, $(x_{i+1},y_{j})$, $(x_{i},y_{j+1})$, and $(x_{i+1},y_{j+1})$, which means $\x_i{\leq} \mathbf{q}_{p0}{\leq} \x_{i+1}$ and $\y_j{\leq} \mathbf{q}_{p1}{\leq} \y_{j+1}$. Then we can obtain the encoding for $\mathbf{q}_{p0}$ and $\mathbf{q}_{p1}$ as follows,
\begin{equation}
\begin{aligned}
    \Psi(\mathbf{q}_{p0}) & \approx\alpha_0\Psi(x_i)+\alpha_1\Psi(x_{i+1}),\\
    \Psi(\mathbf{q}_{p1}) & \approx\beta_0\Psi(y_j)+\beta_1\Psi(y_{j+1})
    \:.
\end{aligned}
\end{equation}
Then, the 2D encoding for $\mathbf{q}_p$ is,
\begin{equation}
\begin{aligned}
    \Psi(\mathbf{q}) = & \Psi(\mathbf{q}_{p0},\mathbf{q}_{p1}) \\
    = &\Psi(\mathbf{q}_{p1})\otimes\Psi(\mathbf{q}_{p0})\\
    \approx&\left(\beta_0\Psi\left(y_j\right){+}\beta_1\Psi\left(y_{j+1}\right)\right) \otimes \left(\alpha_0\Psi\left(x_i\right){+}\alpha_1\Psi\left(x_{i+1}\right)\right) \\
    = &\alpha_0\beta_0\Psi\left(y_j\right)\otimes\Psi\left(x_i\right) + \alpha_0\beta_1\Psi\left(y_{j+1}\right)\otimes\Psi\left(x_i\right) \\
        & + \alpha_1\beta_0\Psi\left(y_j\right)\otimes\Psi\left(x_{i+1}\right) + \alpha_1\beta_1\Psi\left(y_{i+1}\right)\otimes\Psi\left(x_{i+1}\right) \\
    = &\alpha_0\beta_0\Psi\left(x_i,y_j\right) + \alpha_0\beta_1\Psi\left(x_i,y_{j+1}\right) \\
        & + \alpha_1\beta_0\Psi\left(x_{i+1},y_j\right) + \alpha_1\beta_1\Psi\left(x_{i+1},y_{i+1}\right) \:,
\end{aligned}
\end{equation}
which means $B(\mathbf{q}_p){\in}\mathbb{R}^{1\times MN}$ are all zeros except $\alpha_0\beta_0$ at index $jN{+}i$, $\alpha_0\beta_1$ at index $(j{+}1)N{+}i$, $\alpha_0\beta_0$ at index $jN{+}i{+}1$ and $\alpha_0\beta_0$ at index $(j{+}1)N{+}i{+}1$.

\section{HD complexity}
\label{supp:hd_complexity}
Let $\X{\in}\mathbb{R}^{N^{D}{\times} D}$ be $N^D$ points in $D$ dimensional space, $\Psi{:}\mathbb{R}{\to}\mathbb{R}^{K}$ be the 1D encoder, and we want to know the memory and computational complexity when the encoding multiply a linear layer $\W$.

\noindent\textbf{Simple encoding.}\: The embedding $\Psi(\X) {\in} \mathbb{R}^{N^{D}{\times} DK}$ and the weights $\W{\in}\mathbb{R}^{DK\times 1}$, so the memory complexity is $O(DKN^{D})$ and the computational complexity is $O(D K N^{D})$.

\noindent\textbf{Complex encoding (naive implementation).}\: The embedding $\Psi(\X) {\in} \mathbb{R}^{N^{D}{\times} K^{D}}$ and the weights $\W{\in}\mathbb{R}^{K^{D}{\times} 1}$, so the memory complexity is $O(K^{D}N^{D})$ and the computational complexity is $O(K^{D} N^{D})$.

\noindent\textbf{Complex encoding (separable coordinates).}\: The embedding $\Psi(\X) {\in} \mathbb{R}^{N{\times} K}$ and the weights $\W{\in}\mathbb{R}^{K^{D}}$, so the memory complexity is $O(K^{D}{+}NK)$ and the computational complexity is $\sum_{i{=}1}^D N^{i}K^{D{+}1{-}i} {=} O(NK\frac{N^{D}{-}K^{D}}{N{-}K})$. A special case of $N{=}K$ will be discussed later.

\noindent\textbf{Complex encoding (non-separable coordinates).}\: The embedding $\Psi(\X) {\in} \mathbb{R}^{N{\times} K}$, the weights $\W{\in}\mathbb{R}^{K^{D}}$ and the Blending matrix $B(\X){\in} \mathbb{R}^{N^{D}{\times} {N^{D}}}$(sparse matrix with only $N^D {\times} 2^{D}$ non-zeros values), so the memory complexity is $O(K^{D}{+}NK{+}2^{D}N^{D})$, the computational complexity is $2^{D}N^{D}{+}\sum_{i=1}^D N^{i}K^{D{+}1{-}i} {=} O(2^{D}N^{D} {+} NK\frac{N^{D}{-}K^{D}}{N-K})$.

\noindent\textbf{Special case $N{=}K$.}\: Both simple encoding and separable complex encoding have $O(DN^{D{+}1})$ computational encoding. Memory complexity is $O(DN^{D{+}1})$ for simple encoding while it is $O(N^{D}{+}2N)$ for separable encoding. However, the rank of the latter one is power of $D$ to the first one.

\section{Experiments}
\subsection{Method Notations}

For 1D encoding experiments, we used Fourier-feature-based encodings with linearly, log-linearly, or randomly sampled frequencies, and shifted encodings whose bases are Gaussian or triangle. 
We give a brief introduction to these methods below.

\noindent\textbf{LinF (Fourier feature-based encoding using linearly sampled frequency).}
\startcompact{small}
\begin{equation}
    \phi(x) = \begin{bmatrix}
\cdots, \cos\left( 2\pi\cdot \left(\frac{K-i}{K}2^0+\frac{i}{K}2^{\sigma} \right)x \right), \sin\left( 2\pi\cdot \left(\frac{K-i}{K}2^0+\frac{i}{K}2^{\sigma} \right)x \right),\cdots
\end{bmatrix}^T\:,
\end{equation}
\stopcompact{small}
where $i{=}0,{\dots},K{-}1$ and $\sigma$ is the hyperparameter for the frequency range that sampled linearly from base frequency ($2^0$) to max frequency ($2^\sigma$).

\noindent\textbf{LogF (Fourier feature-based encoding using log-linearly sampled frequency).}
\begin{equation}
    \phi(x) = \begin{bmatrix}
\cdots, \cos \left(2\pi\cdot 2^{\sigma i/K}x \right), \sin \left(2\pi\cdot 2^{\sigma i/K}x \right),\cdots
\end{bmatrix}^T\:,
\end{equation}
where $i{=}0,{\dots},K{-}1$ and $\sigma$ is the hyperparameter for frequency range. The frequency are sampled log-linearly from base frequency ($2^0$) to max frequency ($2^\sigma$).

\noindent\textbf{RFF (Fourier feature-based encoding using randomly sampled frequency)~\cite{tancik2020fourier}.}
\begin{equation}
    \phi(x) = \begin{bmatrix}\cos\left(2\pi\b x\right)^T, \sin\left(2\pi\b x\right)^T
\end{bmatrix}^T\:,
\end{equation}
where $\b{\in}\mathbb{R}^{K{\times}1}$ is random frequencies sampled from $\mathcal{N}(0,\sigma^2)$, where $\sigma$ is the hyperparameter for frequency range.

\noindent\textbf{Tri (shifted triangle encoding).}
\begin{equation}
    \phi(x) = \begin{bmatrix}
\cdots, \max\left(1-\left|\frac{x-i/K}{d}\right|,0\right),\cdots
\end{bmatrix}^T\:,
\end{equation}
where $i{=}0,{\dots},K{-}1$ and $d$ is the hyperparameter for the width of triangle wave.

\noindent\textbf{Gau (shifted Gaussian encoding).}
\begin{equation}
    \phi(x) = \begin{bmatrix}
\cdots, e^{-\frac{x-i/K}{2d^2}},\cdots
\end{bmatrix}^T\:,
\end{equation}
where $i{=}0,{\dots},K{-}1$ and $d$ is the hyperparameter for the width of Gaussian wave.

\subsection{Non-separable 3D video reconstruction}
We used the same Youtube video dataset~\cite{esteban2017youtube} as described in the main paper. 
The only difference is that the training points were \emph{randomly} sampled ($12.5\%$ from the total number of points) of a $64{\times}64{\times}64$ grid, and the rest of the points were used for testing. The results are shown in~\cref{tab:3d-non-separable}.
Similar to our observations in the main paper, complex encodings combined with a single linear layer have comparable performance to simple encodings combined with deep (4 layer MLPs) networks while being 10x faster.
Complex frequency-based encodings (LinF, LogF, RFF) have inferior results than complex shifted-based encodings (Tri, Gau) due to deficient rank.

\begin{table}
    \setlength{\tabcolsep}{10pt}
    \centering
    \caption[]{Performance of video reconstruction with randomly sampled inputs (non-separable coordinates).
    \tikz\draw[taborange,fill=taborange] (0,0) circle (.5ex); are simple positional encodings. \tikz\draw[tabpurple,fill=tabpurple] (0,0) circle (.5ex); are complex positional encodings with stochastic gradient descent using smart indexing. Complex encodings with a single linear network are 10x faster than simple encodings with deep networks.}
    \begin{adjustbox}{width=0.7\linewidth}
    \begin{tabular}{lccccc}
    \toprule
    & \thead{PSNR} & \thead{No. of params (memory)} & \thead{Time (s)}
    \\ \midrule
    \tikz\draw[taborange,fill=taborange] (0,0) circle (.5ex);~LinF
    & $26.15\pm3.33$ & $1,445,891 \:(5.78M)$ & $349.49$ \\  
    \tikz\draw[taborange,fill=taborange] (0,0) circle (.5ex);~LogF
    & $25.63\pm2.93$ & $1,445,891 \:(5.78M)$ & $350.62$ \\ 
    \tikz\draw[taborange,fill=taborange] (0,0) circle (.5ex);~RFF~\cite{tancik2020fourier} 
    & $25.46\pm2.73$ & $1,445,891 \:(5.78M)$ & $350.92$ \\    
    \tikz\draw[taborange,fill=taborange] (0,0) circle (.5ex);~Tri
    & $\mathbf{26.44\pm4.24}$ & $1,445,891 \:(5.78M)$ & $352.66$ \\ 
    \tikz\draw[taborange,fill=taborange] (0,0) circle (.5ex);~Gau
    & $24.58\pm2.76$ & $1,445,891 \:(5.78M)$ & $351.45$ \\
    \cmidrule{1-4}
    \tikz\draw[tabpurple,fill=tabpurple] (0,0) circle (.5ex);~LinF
    & $12.87\pm2.38$ & $786,432 \:(3.15M)$ & $3.37$ \\
    \tikz\draw[tabpurple,fill=tabpurple] (0,0) circle (.5ex);~LogF
    & $16.76\pm2.55$ & $786,432 \:(3.15M)$ & $3.41$ \\ 
    \tikz\draw[tabpurple,fill=tabpurple] (0,0) circle (.5ex);~RFF~\cite{tancik2020fourier} 
    & $21.40\pm2.88$ & $786,432 \:(3.15M)$ & $3.62$ \\
    \tikz\draw[tabpurple,fill=tabpurple] (0,0) circle (.5ex);~Tri
    & $23.91\pm2.81$ & $786,432 \:(3.15M)$ & $3.39$ \\
    \tikz\draw[tabpurple,fill=tabpurple] (0,0) circle (.5ex);~Gau
    & $\mathbf{24.44\pm2.79}$ & $786,432 \:(3.15M)$ & $3.46$ \\
    \arrayrulecolor{black}\bottomrule
    \end{tabular}
    \end{adjustbox}
    \label{tab:3d-non-separable}
\end{table}

\subsection{Visual results for 2D images}
\label{supp:2d_image_visual_results}
\begin{figure}[!ht]
\captionsetup[subfigure]{labelformat=empty}
\centering 

\begin{subfigure}{0.2\textwidth}
\stackinset{c}{}{t}{-1.0\baselineskip}{\textbf{Ground Truth}}{%
\includegraphics[width=\linewidth]{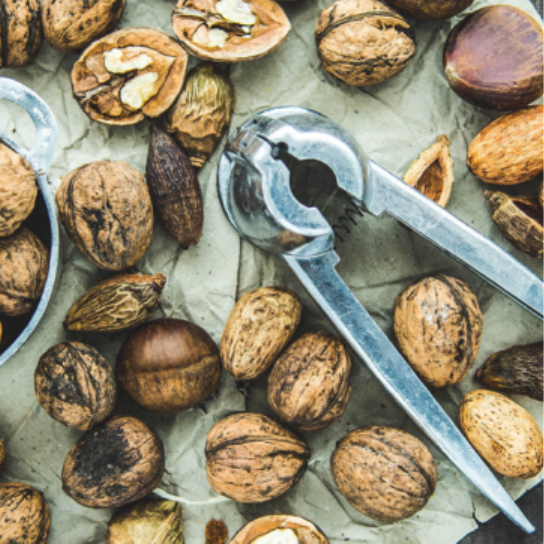}}
\vspace{-0.5\baselineskip}
\end{subfigure}

\begin{subfigure}{0.015\textwidth}\hspace{-2mm}
\rotatebox[origin=c]{90}{\textbf{LinF}}
\end{subfigure}
\begin{subfigure}{0.2\textwidth}
\stackinset{c}{}{t}{-1.0\baselineskip}{\textbf{Simple, Depth 0}}{%
\includegraphics[width=\linewidth]{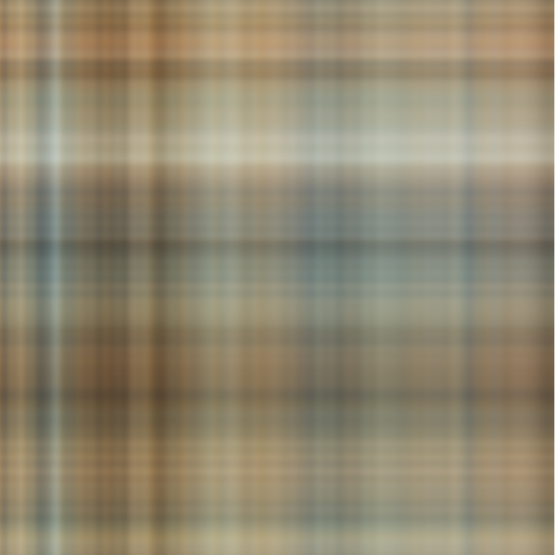}}
\vspace{-1.5\baselineskip}
\caption{PSNR: 12.90}
\end{subfigure} 
\begin{subfigure}{0.2\textwidth}
\stackinset{c}{}{t}{-1.0\baselineskip}{\textbf{Simple, Depth 4}}{%
\includegraphics[width=\linewidth]{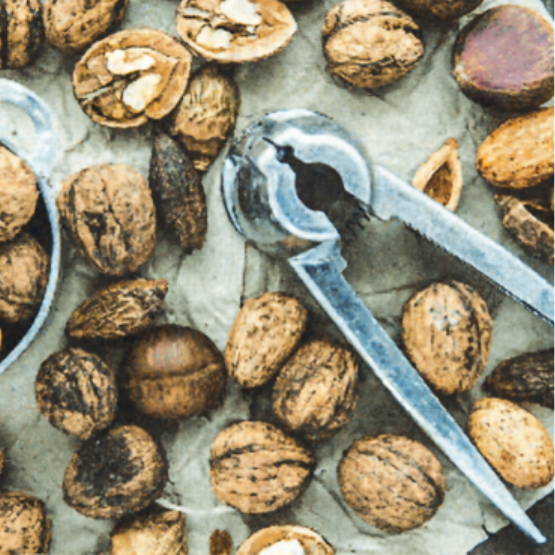} }
\vspace{-1.5\baselineskip}
\caption{PSNR: 21.35}
\end{subfigure}
\begin{subfigure}{0.2\textwidth}
\stackinset{c}{}{t}{-1.0\baselineskip}{\textbf{Complex, Depth 0}}{%
\includegraphics[width=\linewidth]{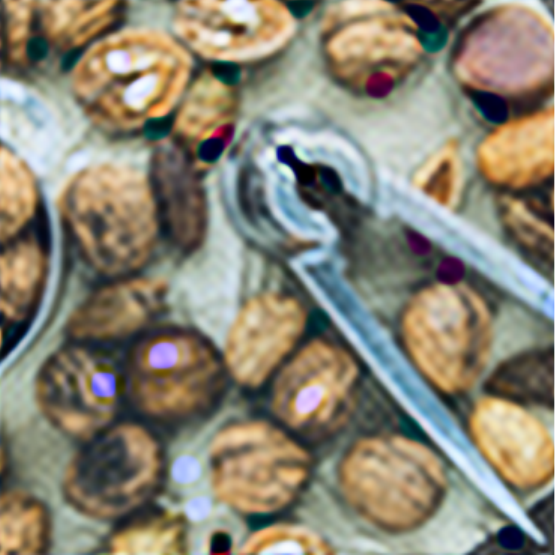}}
\vspace{-1.5\baselineskip}
\caption{PSNR: 18.94}
\end{subfigure}
\begin{subfigure}{0.2\textwidth}
\stackinset{c}{}{t}{-1.0\baselineskip}{\textbf{Complex, Depth 1}}{%
\includegraphics[width=\linewidth]{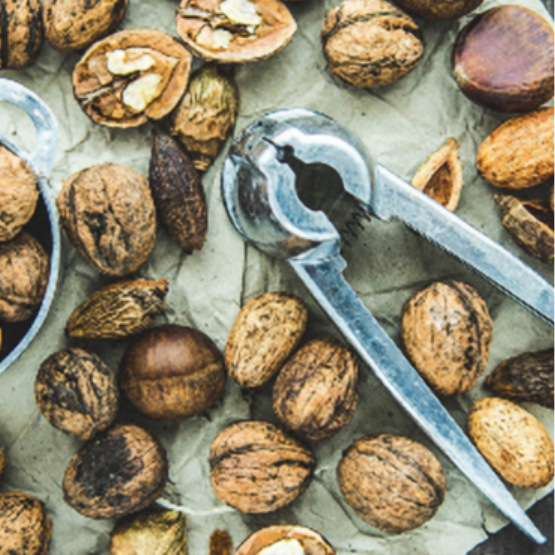}}
\vspace{-1.5\baselineskip}
\caption{PSNR: 23.01}
\end{subfigure}

\begin{subfigure}{0.015\textwidth}\hspace{-2mm}
\rotatebox[origin=c]{90}{\textbf{LogF}}
\end{subfigure}
\begin{subfigure}{0.2\textwidth}
\includegraphics[width=\linewidth]{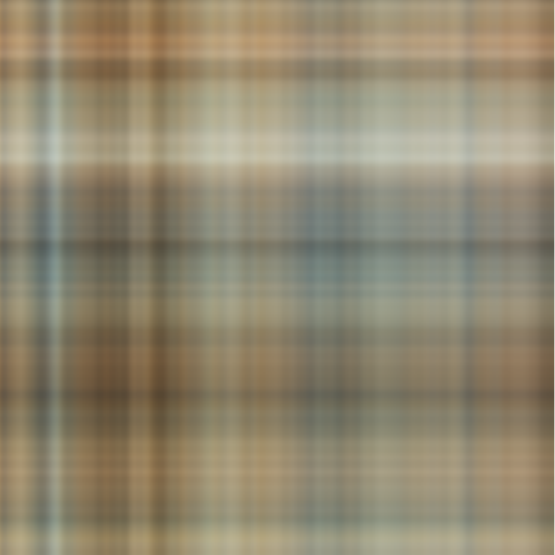} 
\vspace{-1.5\baselineskip}
\caption{PSNR: 12.89}
\end{subfigure} 
\begin{subfigure}{0.2\textwidth}
\includegraphics[width=\linewidth]{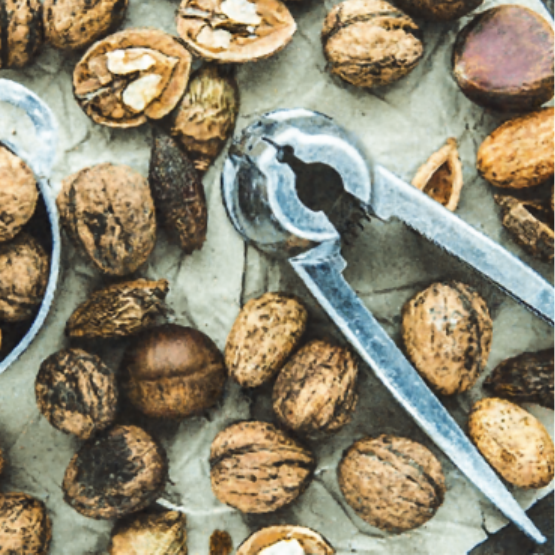} 
\vspace{-1.5\baselineskip}
\caption{PSNR: 21.66}
\end{subfigure} 
\begin{subfigure}{0.2\textwidth}
\includegraphics[width=\linewidth]{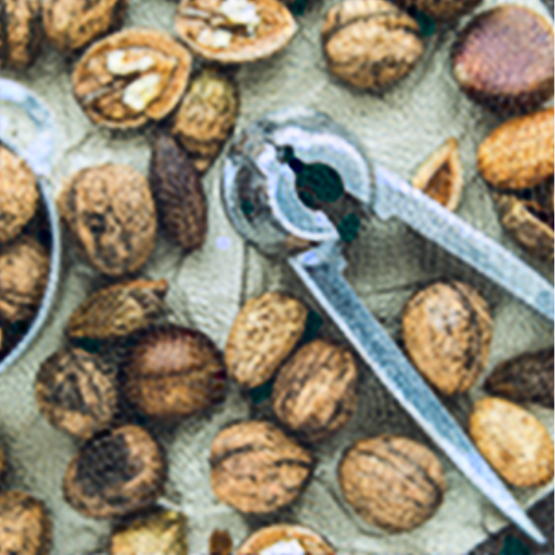} 
\vspace{-1.5\baselineskip}
\caption{PSNR: 20.71}
\end{subfigure}
\begin{subfigure}{0.2\textwidth}
\includegraphics[width=\linewidth]{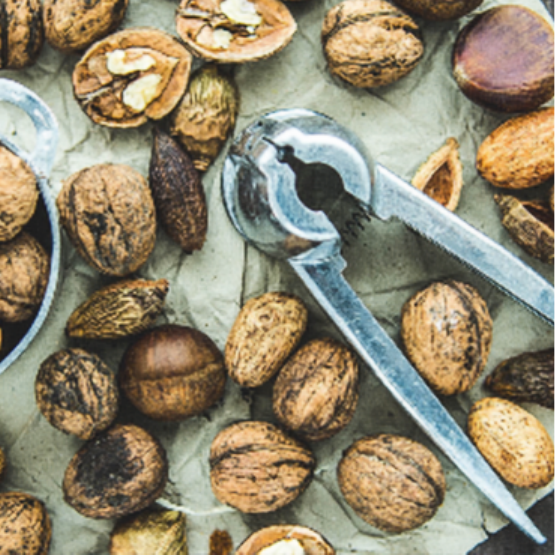} 
\vspace{-1.5\baselineskip}
\caption{PSNR: 22.91}
\end{subfigure}

\begin{subfigure}{0.015\textwidth}\hspace{-2mm}
\rotatebox[origin=c]{90}{\textbf{RFF}}
\end{subfigure}
\begin{subfigure}{0.2\textwidth}
\includegraphics[width=\linewidth]{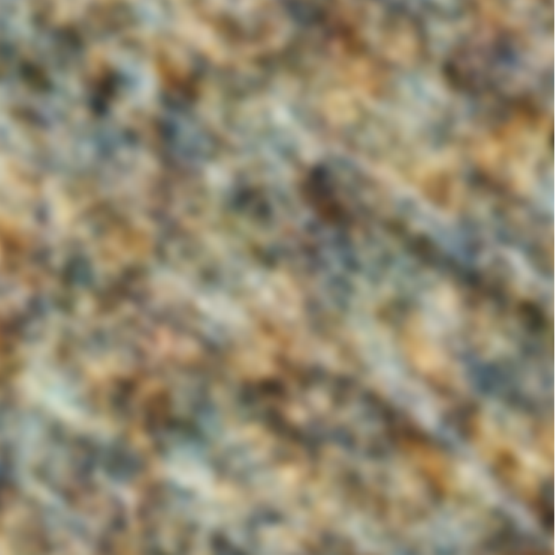} 
\vspace{-1.5\baselineskip}
\caption{PSNR: 13.17}
\end{subfigure}
\begin{subfigure}{0.2\textwidth}
\includegraphics[width=\linewidth]{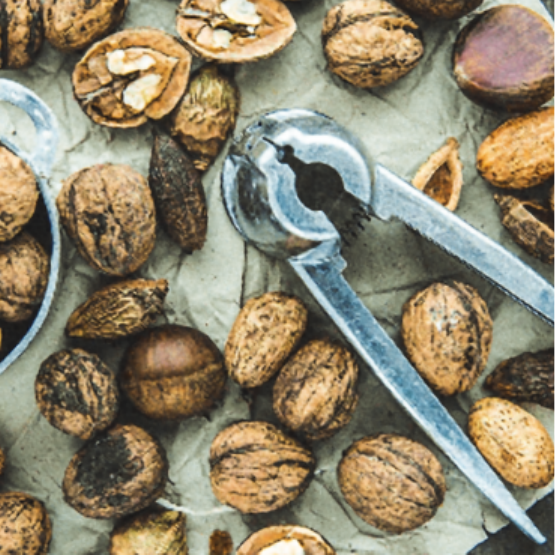} 
\vspace{-1.5\baselineskip}
\caption{PSNR: 22.43}
\end{subfigure} 
\begin{subfigure}{0.2\textwidth}
\includegraphics[width=\linewidth]{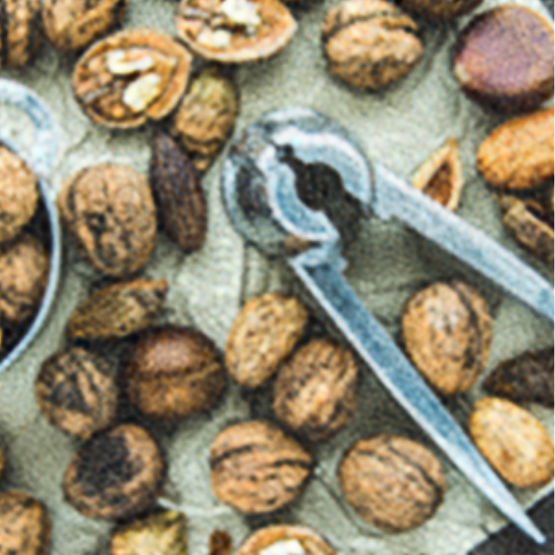} 
\vspace{-1.5\baselineskip}
\caption{PSNR: 20.77}
\end{subfigure}
\begin{subfigure}{0.2\textwidth}
\includegraphics[width=\linewidth]{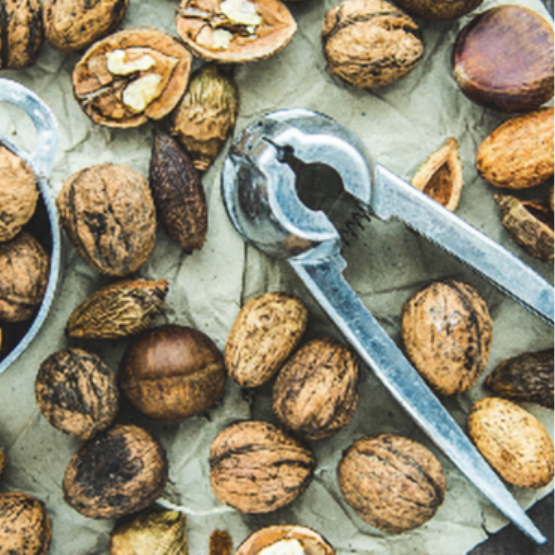} 
\vspace{-1.5\baselineskip}
\caption{PSNR: 22.97}
\end{subfigure}

\begin{subfigure}{0.015\textwidth}\hspace{-2mm}
\rotatebox[origin=c]{90}{\textbf{Tri}}
\end{subfigure}
\begin{subfigure}{0.2\textwidth}
\includegraphics[width=\linewidth]{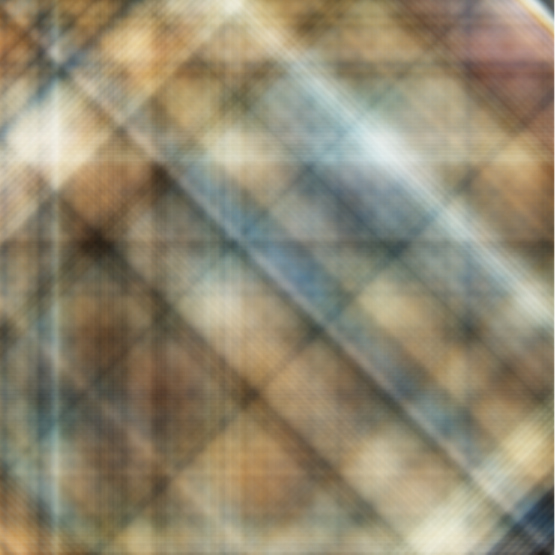} 
\vspace{-1.5\baselineskip}
\caption{PSNR: 13.87}
\end{subfigure} 
\begin{subfigure}{0.2\textwidth}
\includegraphics[width=\linewidth]{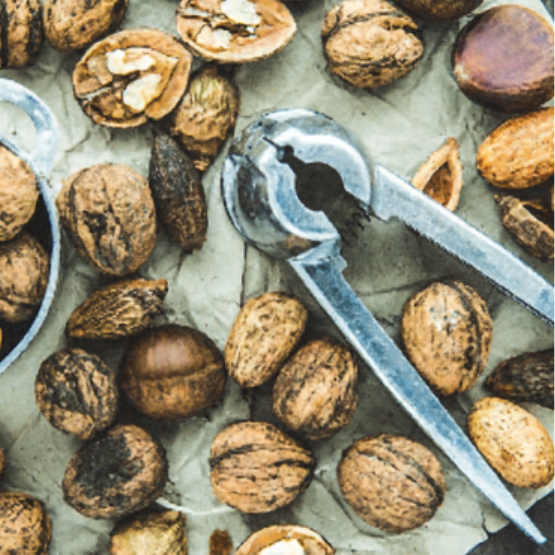} 
\vspace{-1.5\baselineskip}
\caption{PSNR: 21.98}
\end{subfigure}
\begin{subfigure}{0.2\textwidth}
\includegraphics[width=\linewidth]{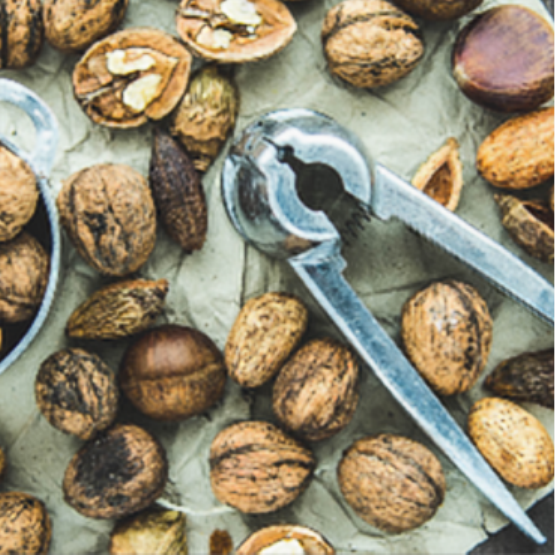} 
\vspace{-1.5\baselineskip}
\caption{PSNR: 22.74}
\end{subfigure} 
\begin{subfigure}{0.2\textwidth}
\includegraphics[width=\linewidth]{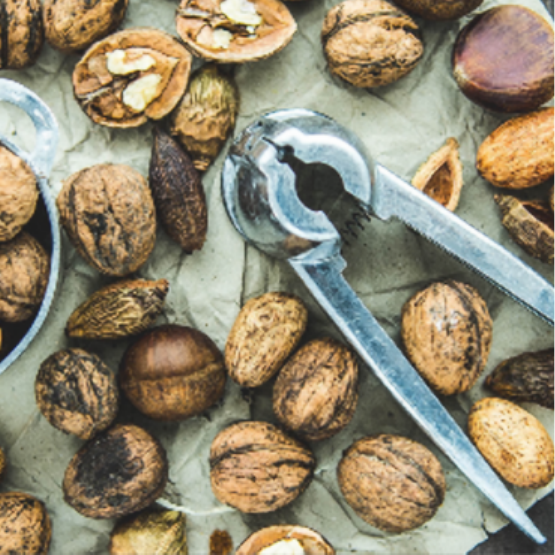} 
\vspace{-1.5\baselineskip}
\caption{PSNR: 23.13}
\end{subfigure}

\begin{subfigure}{0.015\textwidth}\hspace{-2mm}
\rotatebox[origin=c]{90}{\textbf{Gau}}
\end{subfigure}
\begin{subfigure}{0.2\textwidth}
\includegraphics[width=\linewidth]{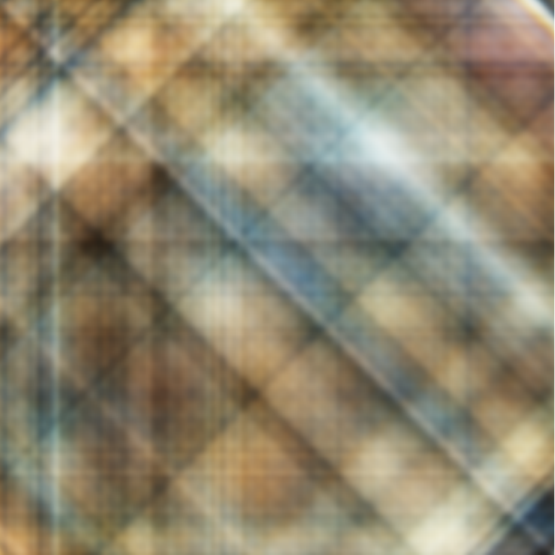}
\vspace{-1.5\baselineskip}
\caption{PSNR: 13.89}
\end{subfigure} 
\begin{subfigure}{0.2\textwidth}
\includegraphics[width=\linewidth]{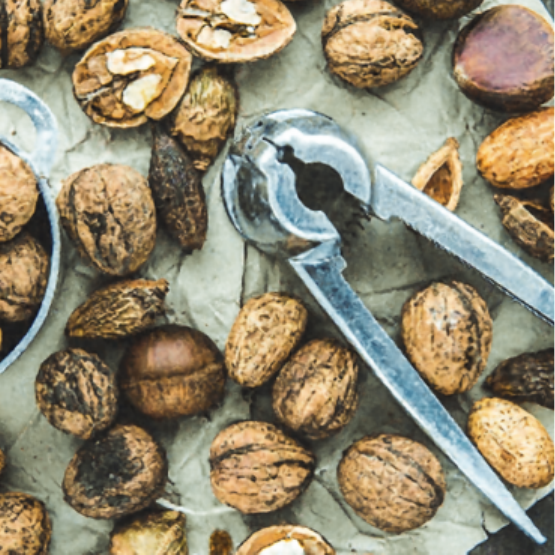} 
\vspace{-1.5\baselineskip}
\caption{PSNR: 22.29}
\end{subfigure}
\begin{subfigure}{0.2\textwidth}
\includegraphics[width=\linewidth]{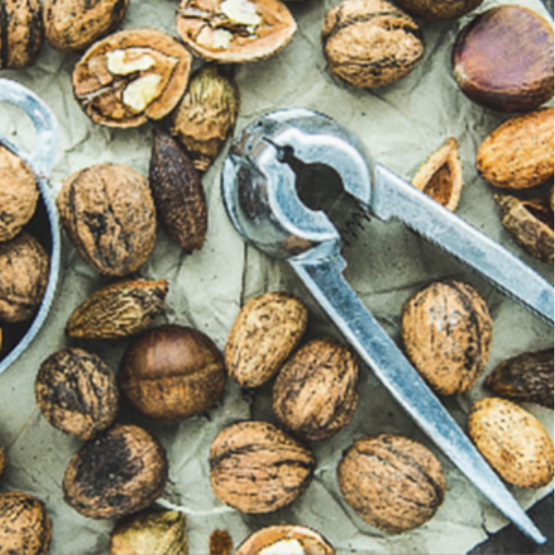}
\vspace{-1.5\baselineskip}
\caption{PSNR: 22.91}
\end{subfigure}
\begin{subfigure}{0.2\textwidth}
\includegraphics[width=\linewidth]{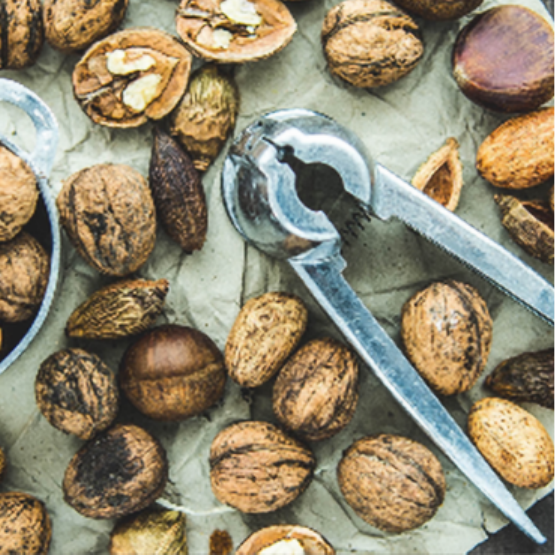}
\vspace{-1.5\baselineskip}
\caption{PSNR: 22.81}
\end{subfigure} 
\vspace{-0.2cm}
\caption{Reconstruction results of a heap of walnuts using separable coordinates (regular-grid sampled training points) with different combinations of simple or complex encodings and network depths.}
\label{fig:2d_grid_2}
\end{figure}

\begin{figure}[ht]
\captionsetup[subfigure]{labelformat=empty}
\centering 

\begin{subfigure}{0.2\textwidth}
\stackinset{c}{}{t}{-1.0\baselineskip}{\textbf{Ground Truth}}{%
\includegraphics[width=\linewidth]{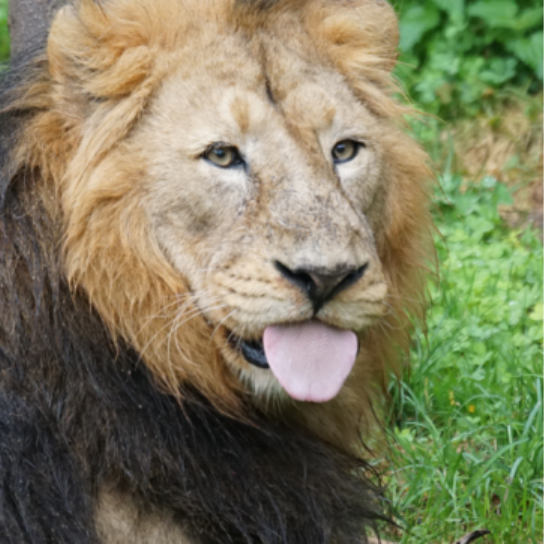}}
\vspace{-0.5\baselineskip}
\end{subfigure}

\begin{subfigure}{0.015\textwidth}\hspace{-2mm}
\rotatebox[origin=c]{90}{\textbf{LinF}}
\end{subfigure}
\begin{subfigure}{0.2\textwidth}
\stackinset{c}{}{t}{-1.0\baselineskip}{\textbf{Simple, Depth 0}}{%
\includegraphics[width=\linewidth]{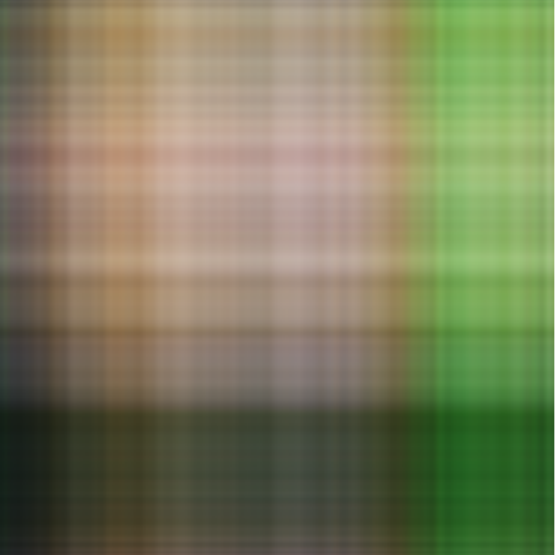}}
\vspace{-1.5\baselineskip}
\caption{PSNR: 16.32}
\end{subfigure} 
\begin{subfigure}{0.2\textwidth}
\stackinset{c}{}{t}{-1.0\baselineskip}{\textbf{Simple, Depth 4}}{%
\includegraphics[width=\linewidth]{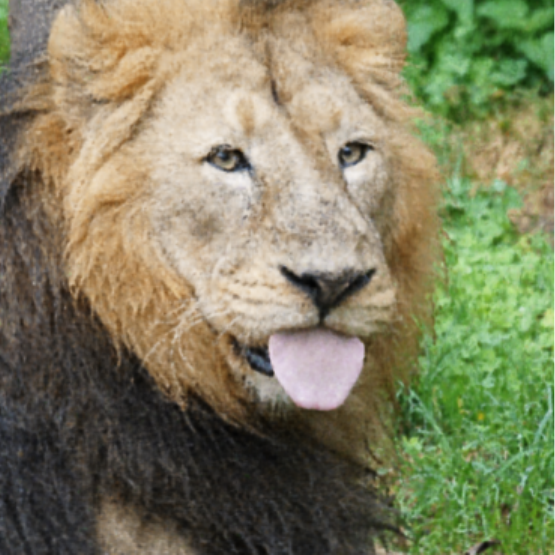} }
\vspace{-1.5\baselineskip}
\caption{PSNR: 26.37}
\end{subfigure}
\begin{subfigure}{0.2\textwidth}
\stackinset{c}{}{t}{-1.0\baselineskip}{\textbf{Complex, Depth 0}}{%
\includegraphics[width=\linewidth]{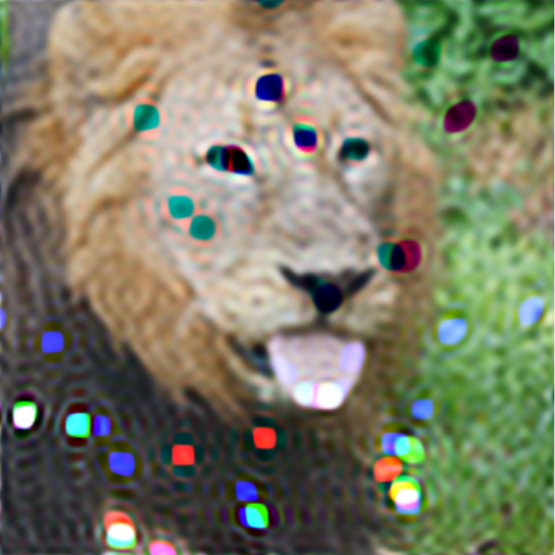}}
\vspace{-1.5\baselineskip}
\caption{PSNR: 18.39}
\end{subfigure}
\begin{subfigure}{0.2\textwidth}
\stackinset{c}{}{t}{-1.0\baselineskip}{\textbf{Complex, Depth 1}}{%
\includegraphics[width=\linewidth]{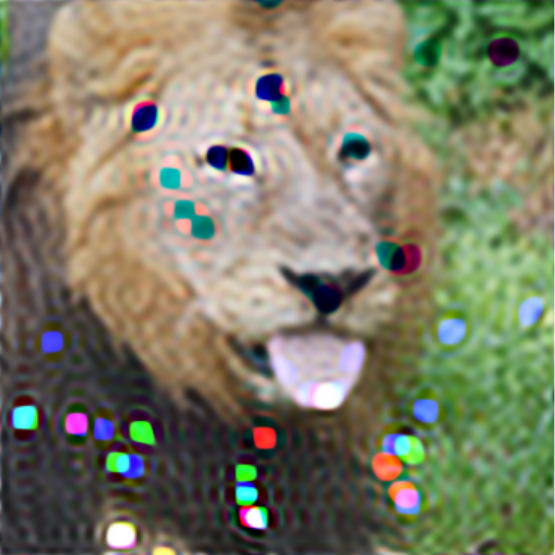}}
\vspace{-1.5\baselineskip}
\caption{PSNR: 18.18}
\end{subfigure}

\begin{subfigure}{0.015\textwidth}\hspace{-2mm}
\rotatebox[origin=c]{90}{\textbf{LogF}}
\end{subfigure}
\begin{subfigure}{0.2\textwidth}
\includegraphics[width=\linewidth]{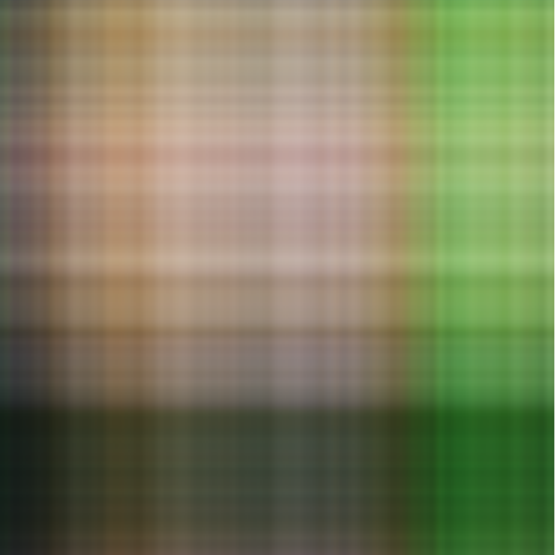} 
\vspace{-1.5\baselineskip}
\caption{PSNR: 16.33}
\end{subfigure} 
\begin{subfigure}{0.2\textwidth}
\includegraphics[width=\linewidth]{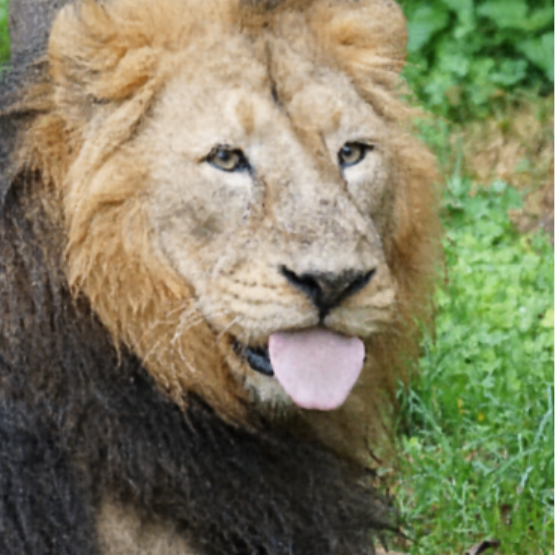} 
\vspace{-1.5\baselineskip}
\caption{PSNR: 27.09}
\end{subfigure} 
\begin{subfigure}{0.2\textwidth}
\includegraphics[width=\linewidth]{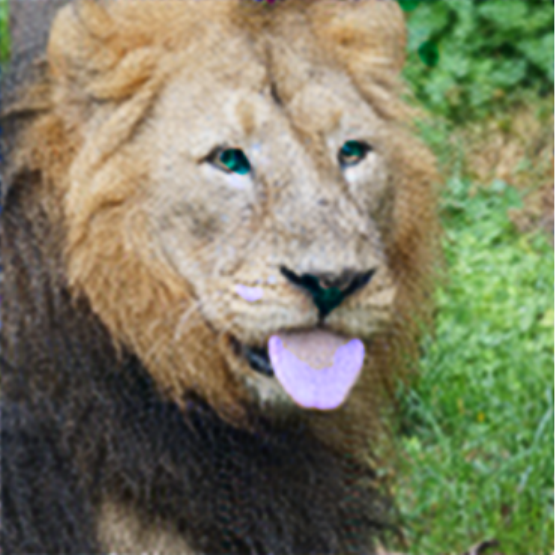} 
\vspace{-1.5\baselineskip}
\caption{PSNR: 25.17}
\end{subfigure}
\begin{subfigure}{0.2\textwidth}
\includegraphics[width=\linewidth]{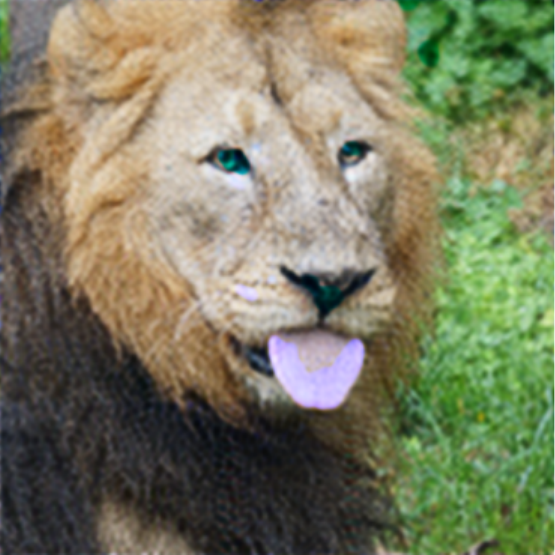} 
\vspace{-1.5\baselineskip}
\caption{PSNR: 25.15}
\end{subfigure}

\begin{subfigure}{0.015\textwidth}\hspace{-2mm}
\rotatebox[origin=c]{90}{\textbf{RFF}}
\end{subfigure}
\begin{subfigure}{0.2\textwidth}
\includegraphics[width=\linewidth]{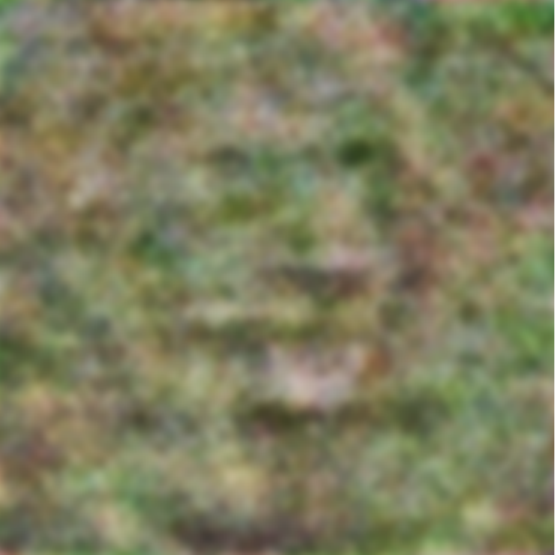} 
\vspace{-1.5\baselineskip}
\caption{PSNR: 13.43}
\end{subfigure}
\begin{subfigure}{0.2\textwidth}
\includegraphics[width=\linewidth]{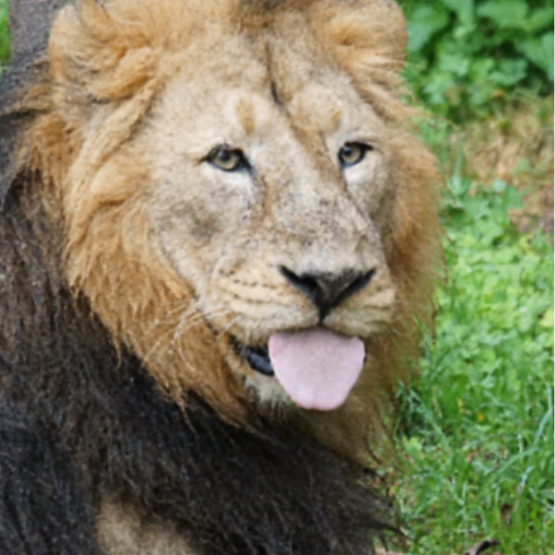} 
\vspace{-1.5\baselineskip}
\caption{PSNR: 27.31}
\end{subfigure} 
\begin{subfigure}{0.2\textwidth}
\includegraphics[width=\linewidth]{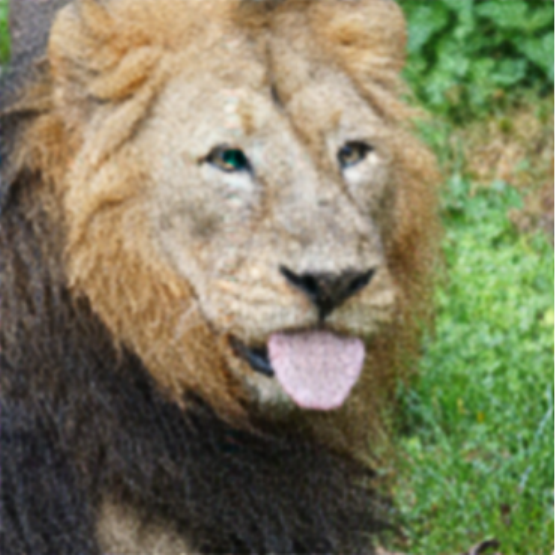} 
\vspace{-1.5\baselineskip}
\caption{PSNR: 25.54}
\end{subfigure}
\begin{subfigure}{0.2\textwidth}
\includegraphics[width=\linewidth]{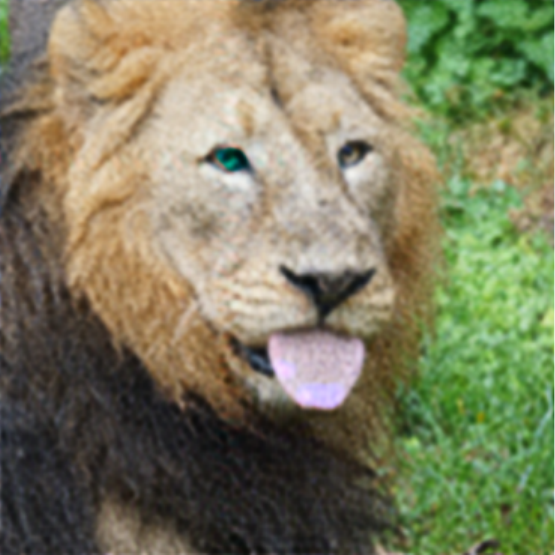} 
\vspace{-1.5\baselineskip}
\caption{PSNR: 25.52}
\end{subfigure}

\begin{subfigure}{0.015\textwidth}\hspace{-2mm}
\rotatebox[origin=c]{90}{\textbf{Tri}}
\end{subfigure}
\begin{subfigure}{0.2\textwidth}
\includegraphics[width=\linewidth]{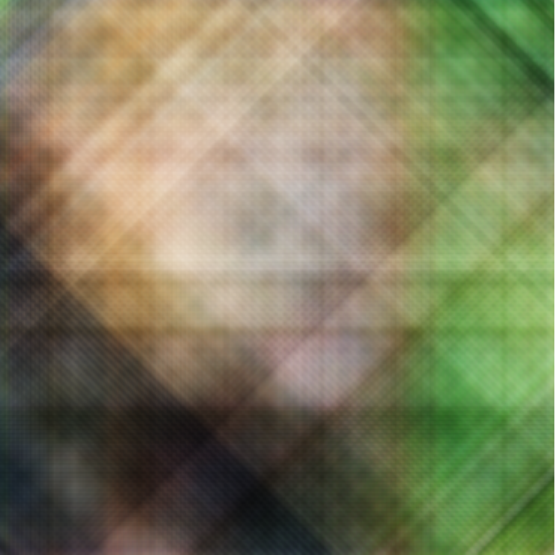} 
\vspace{-1.5\baselineskip}
\caption{PSNR: 18.55}
\end{subfigure} 
\begin{subfigure}{0.2\textwidth}
\includegraphics[width=\linewidth]{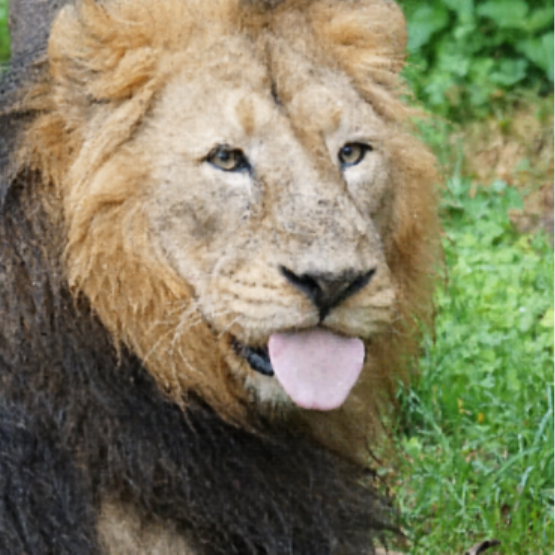} 
\vspace{-1.5\baselineskip}
\caption{PSNR: 26.93}
\end{subfigure}
\begin{subfigure}{0.2\textwidth}
\includegraphics[width=\linewidth]{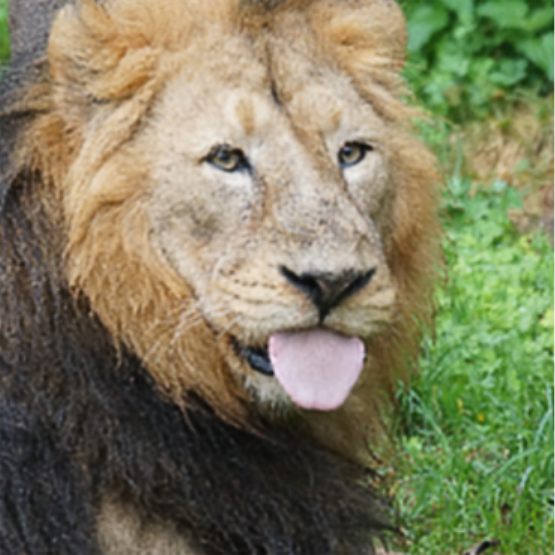} 
\vspace{-1.5\baselineskip}
\caption{PSNR: 27.28}
\end{subfigure} 
\begin{subfigure}{0.2\textwidth}
\includegraphics[width=\linewidth]{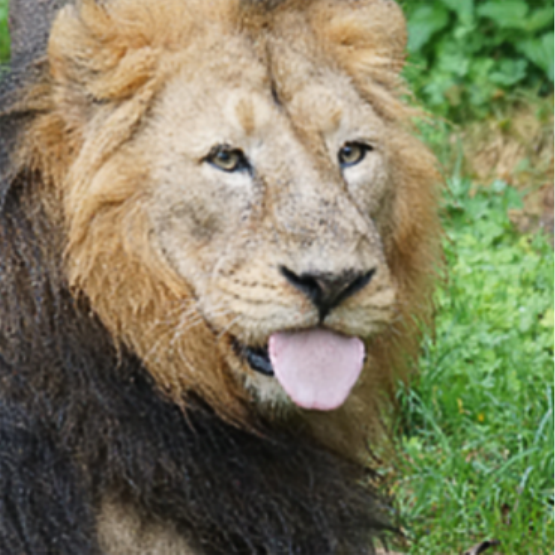} 
\vspace{-1.5\baselineskip}
\caption{PSNR: 27.28}
\end{subfigure}

\begin{subfigure}{0.015\textwidth}\hspace{-2mm}
\rotatebox[origin=c]{90}{\textbf{Gau}}
\end{subfigure}
\begin{subfigure}{0.2\textwidth}
\includegraphics[width=\linewidth]{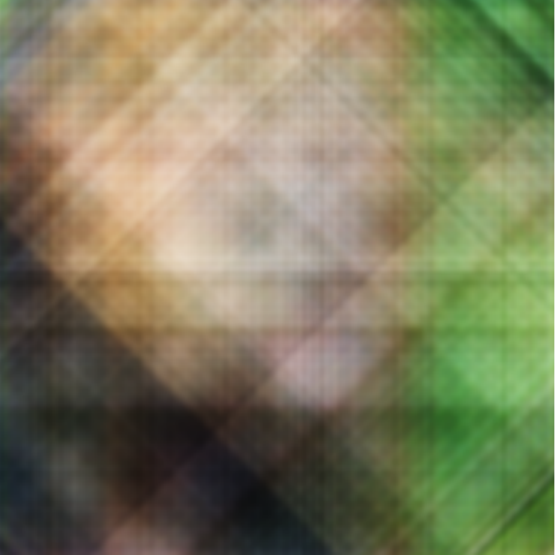}
\vspace{-1.5\baselineskip}
\caption{PSNR: 18.59}
\end{subfigure} 
\begin{subfigure}{0.2\textwidth}
\includegraphics[width=\linewidth]{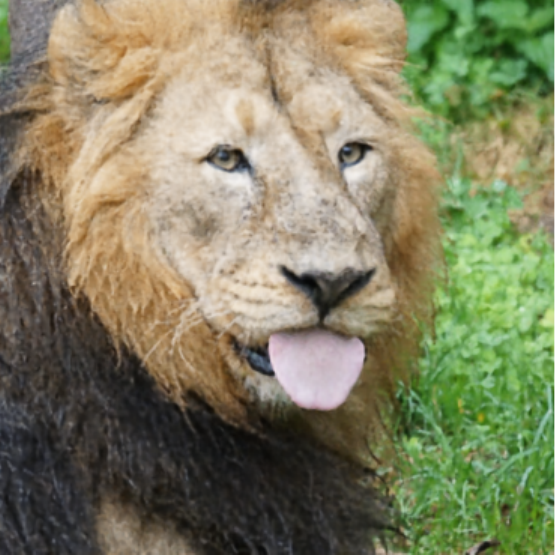} 
\vspace{-1.5\baselineskip}
\caption{PSNR: 26.88}
\end{subfigure}
\begin{subfigure}{0.2\textwidth}
\includegraphics[width=\linewidth]{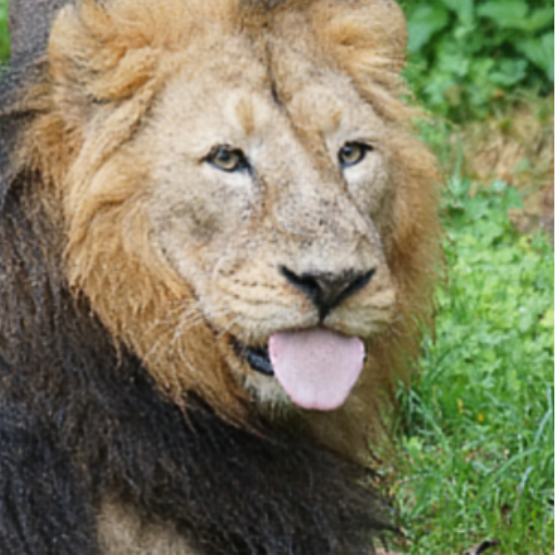}
\vspace{-1.5\baselineskip}
\caption{PSNR: 27.23}
\end{subfigure}
\begin{subfigure}{0.2\textwidth}
\includegraphics[width=\linewidth]{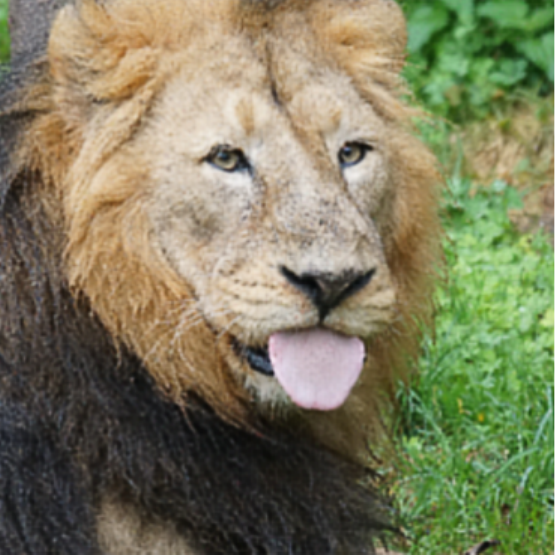}
\vspace{-1.5\baselineskip}
\caption{PSNR: 27.23}
\end{subfigure} 
\vspace{-0.2cm}
\caption{Reconstruction results of a lion using non-separable coordinates (randomly sampled training points) with different combinations of simple or complex encodings and network depths.}
\label{fig:2d_random_3}
\end{figure}

\begin{figure}[!ht]
\captionsetup[subfigure]{labelformat=empty}
\centering 

\begin{subfigure}{0.2\textwidth}
\stackinset{c}{}{t}{-1.0\baselineskip}{\textbf{Ground Truth}}{%
\includegraphics[width=\linewidth]{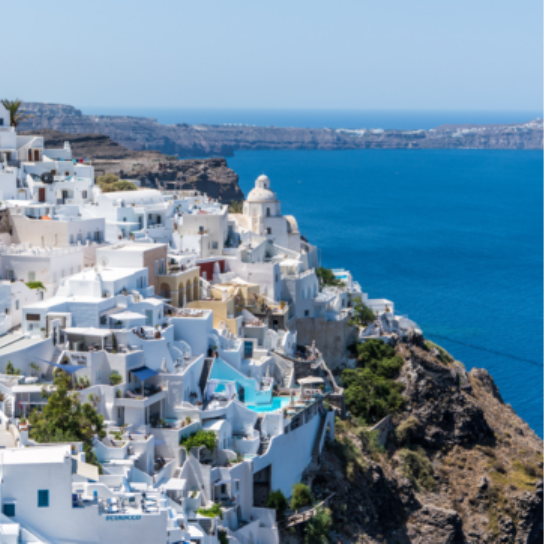}}
\vspace{-0.5\baselineskip}
\end{subfigure}

\begin{subfigure}{0.015\textwidth}\hspace{-2mm}
\rotatebox[origin=c]{90}{\textbf{LinF}}
\end{subfigure}
\begin{subfigure}{0.2\textwidth}
\stackinset{c}{}{t}{-1.0\baselineskip}{\textbf{Simple, Depth 0}}{%
\includegraphics[width=\linewidth]{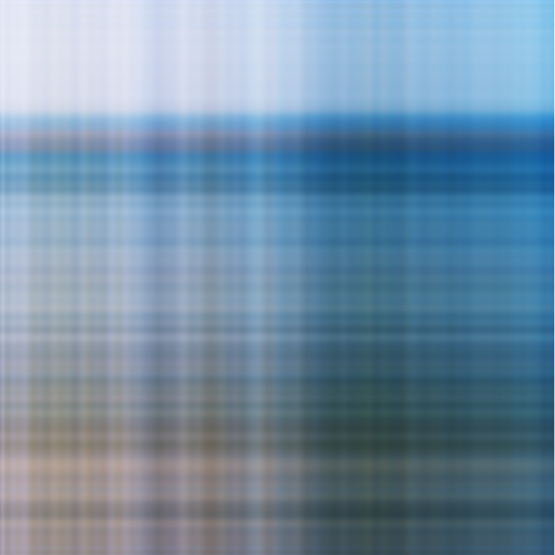}}
\vspace{-1.5\baselineskip}
\caption{PSNR: 14.99}
\end{subfigure} 
\begin{subfigure}{0.2\textwidth}
\stackinset{c}{}{t}{-1.0\baselineskip}{\textbf{Simple, Depth 4}}{%
\includegraphics[width=\linewidth]{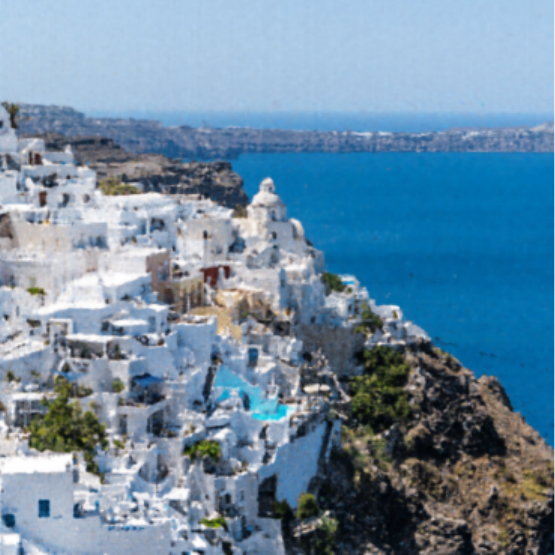} }
\vspace{-1.5\baselineskip}
\caption{PSNR: 22.68}
\end{subfigure}
\begin{subfigure}{0.2\textwidth}
\stackinset{c}{}{t}{-1.0\baselineskip}{\textbf{Complex, Depth 0}}{%
\includegraphics[width=\linewidth]{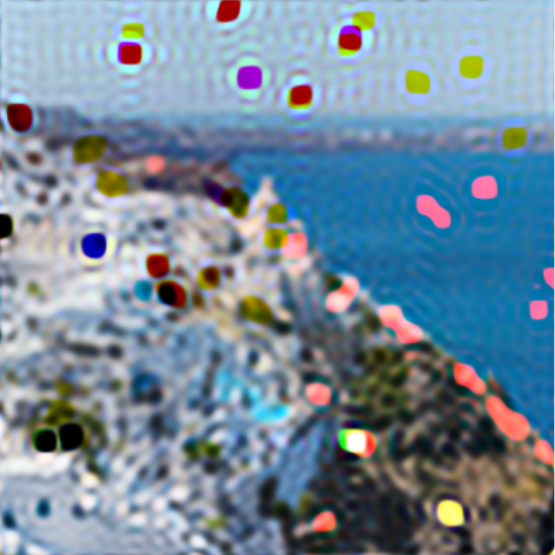}}
\vspace{-1.5\baselineskip}
\caption{PSNR: 16.62}
\end{subfigure}
\begin{subfigure}{0.2\textwidth}
\stackinset{c}{}{t}{-1.0\baselineskip}{\textbf{Complex, Depth 1}}{%
\includegraphics[width=\linewidth]{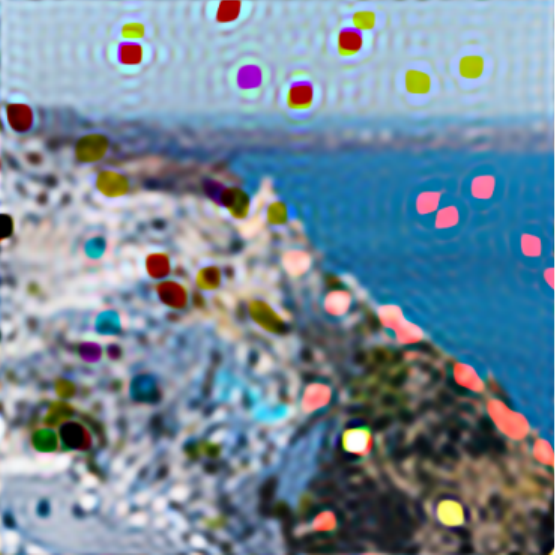}}
\vspace{-1.5\baselineskip}
\caption{PSNR: 16.71}
\end{subfigure}

\begin{subfigure}{0.015\textwidth}\hspace{-2mm}
\rotatebox[origin=c]{90}{\textbf{LogF}}
\end{subfigure}
\begin{subfigure}{0.2\textwidth}
\includegraphics[width=\linewidth]{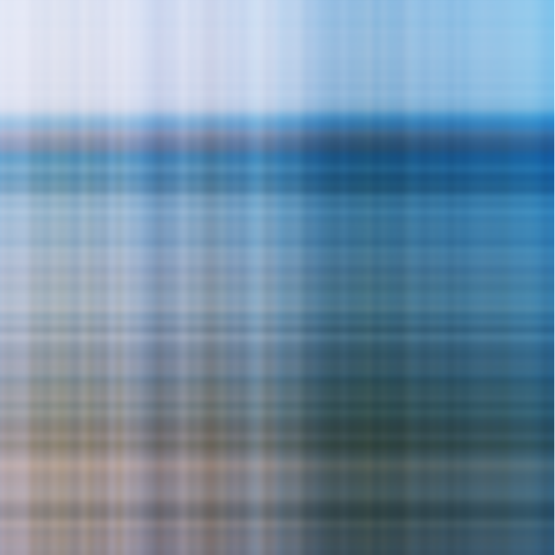} 
\vspace{-1.5\baselineskip}
\caption{PSNR: 14.98}
\end{subfigure} 
\begin{subfigure}{0.2\textwidth}
\includegraphics[width=\linewidth]{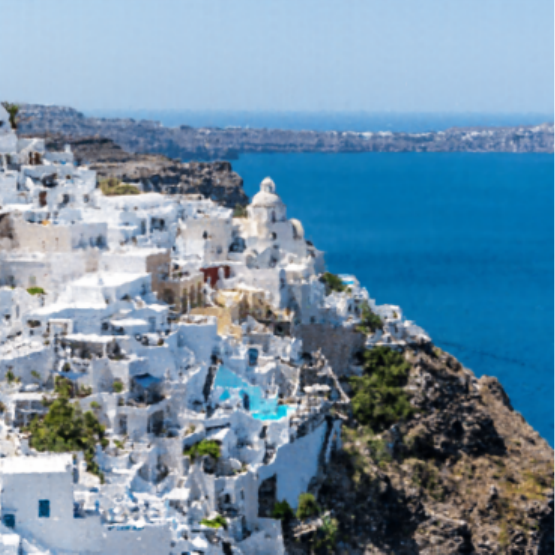} 
\vspace{-1.5\baselineskip}
\caption{PSNR: 22.72}
\end{subfigure} 
\begin{subfigure}{0.2\textwidth}
\includegraphics[width=\linewidth]{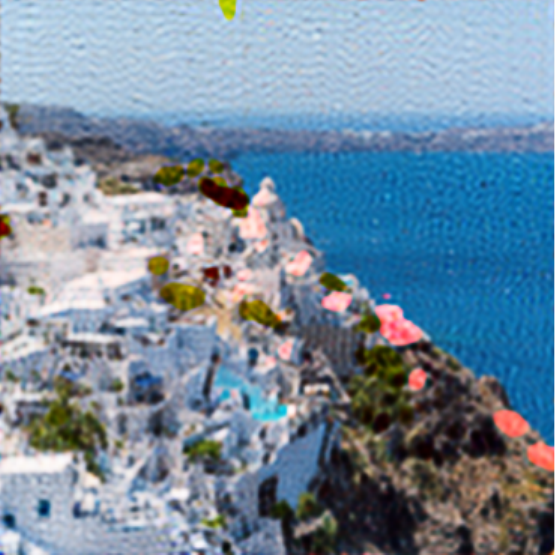} 
\vspace{-1.5\baselineskip}
\caption{PSNR: 20.48}
\end{subfigure}
\begin{subfigure}{0.2\textwidth}
\includegraphics[width=\linewidth]{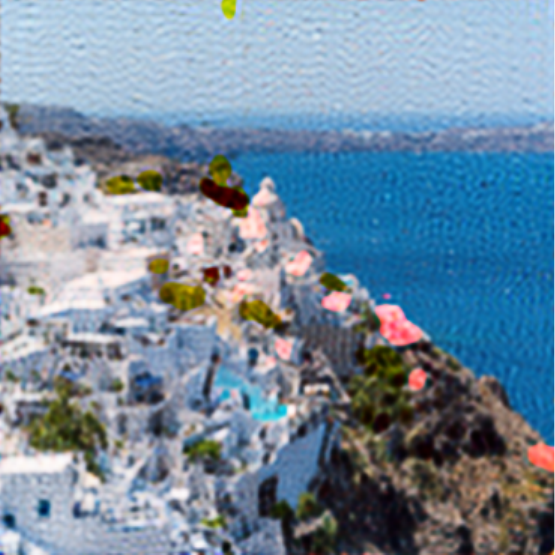} 
\vspace{-1.5\baselineskip}
\caption{PSNR: 20.55}
\end{subfigure}

\begin{subfigure}{0.015\textwidth}\hspace{-2mm}
\rotatebox[origin=c]{90}{\textbf{RFF}}
\end{subfigure}
\begin{subfigure}{0.2\textwidth}
\includegraphics[width=\linewidth]{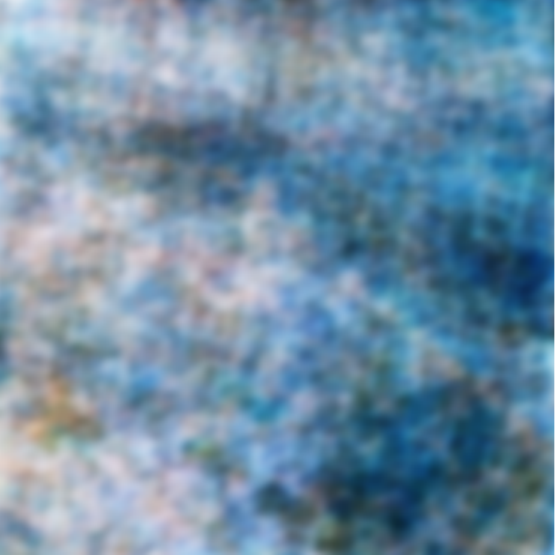} 
\vspace{-1.5\baselineskip}
\caption{PSNR: 14.00}
\end{subfigure}
\begin{subfigure}{0.2\textwidth}
\includegraphics[width=\linewidth]{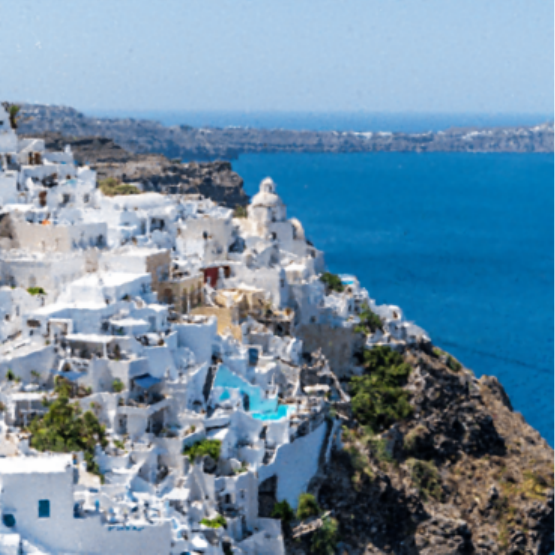} 
\vspace{-1.5\baselineskip}
\caption{PSNR: 23.25}
\end{subfigure} 
\begin{subfigure}{0.2\textwidth}
\includegraphics[width=\linewidth]{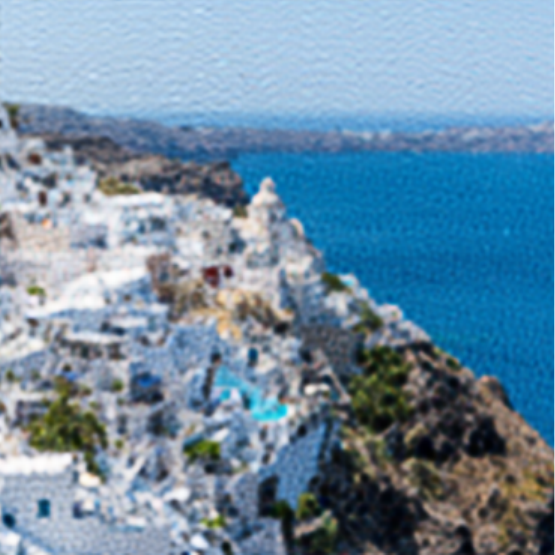} 
\vspace{-1.5\baselineskip}
\caption{PSNR: 21.69}
\end{subfigure}
\begin{subfigure}{0.2\textwidth}
\includegraphics[width=\linewidth]{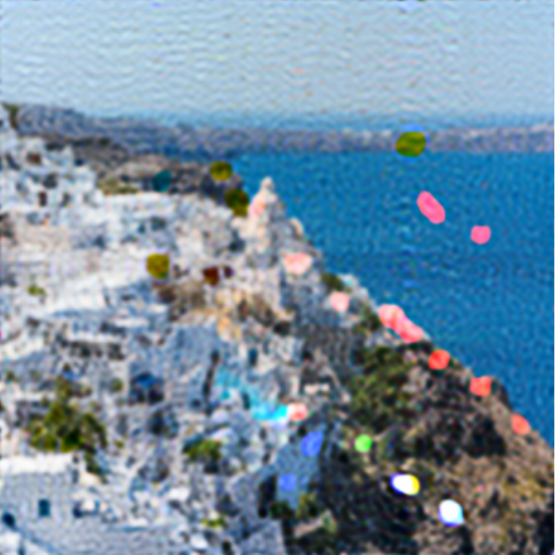} 
\vspace{-1.5\baselineskip}
\caption{PSNR: 19.51}
\end{subfigure}

\begin{subfigure}{0.015\textwidth}\hspace{-2mm}
\rotatebox[origin=c]{90}{\textbf{Tri}}
\end{subfigure}
\begin{subfigure}{0.2\textwidth}
\includegraphics[width=\linewidth]{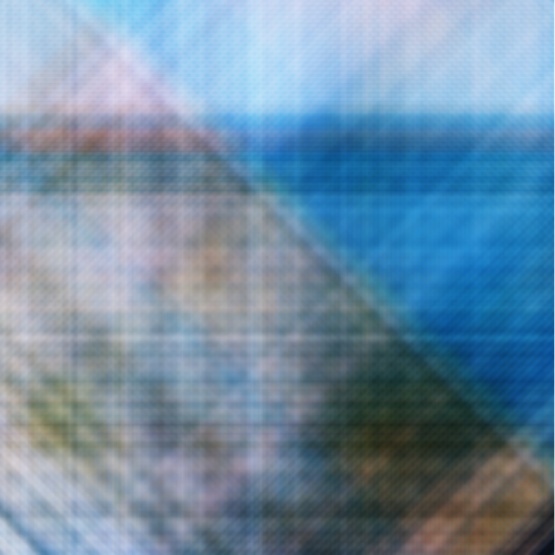} 
\vspace{-1.5\baselineskip}
\caption{PSNR: 16.31}
\end{subfigure} 
\begin{subfigure}{0.2\textwidth}
\includegraphics[width=\linewidth]{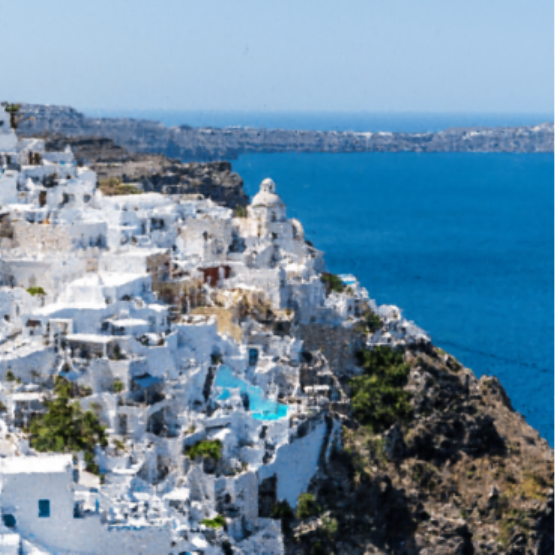} 
\vspace{-1.5\baselineskip}
\caption{PSNR: 21.85}
\end{subfigure}
\begin{subfigure}{0.2\textwidth}
\includegraphics[width=\linewidth]{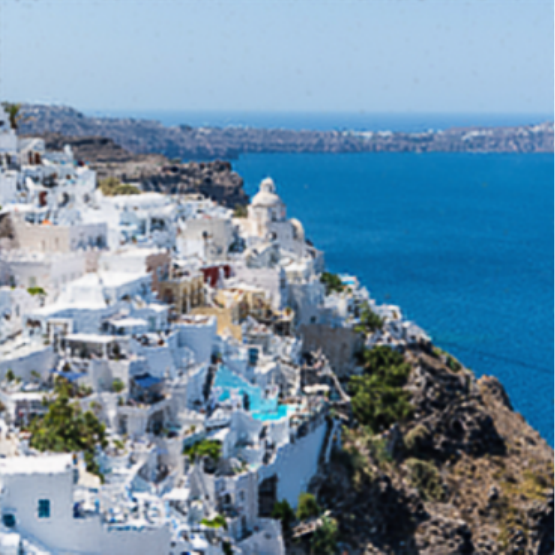} 
\vspace{-1.5\baselineskip}
\caption{PSNR: 22.74}
\end{subfigure} 
\begin{subfigure}{0.2\textwidth}
\includegraphics[width=\linewidth]{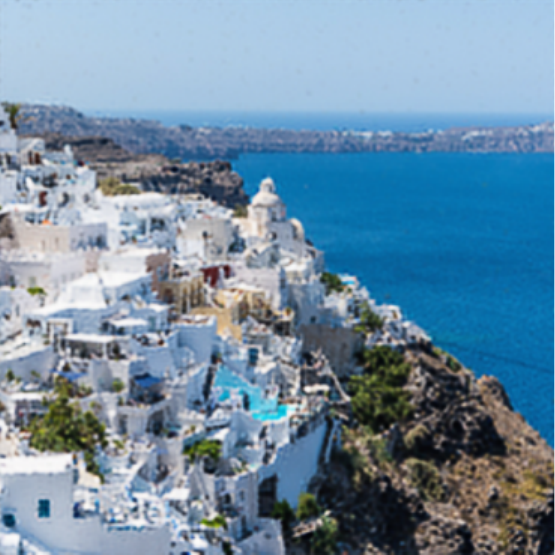} 
\vspace{-1.5\baselineskip}
\caption{PSNR: 22.73}
\end{subfigure}

\begin{subfigure}{0.015\textwidth}\hspace{-2mm}
\rotatebox[origin=c]{90}{\textbf{Gau}}
\end{subfigure}
\begin{subfigure}{0.2\textwidth}
\includegraphics[width=\linewidth]{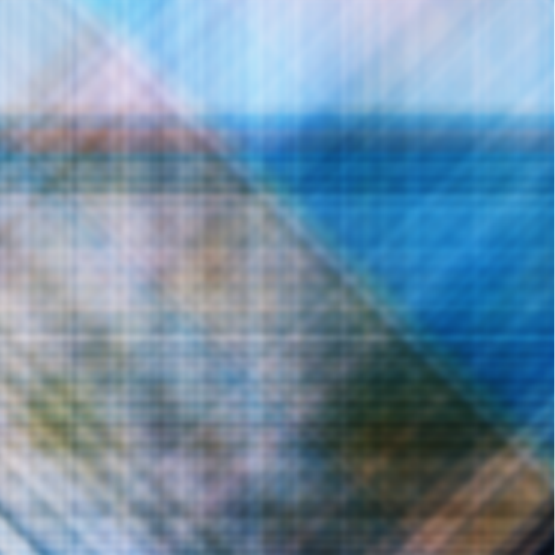}
\vspace{-1.5\baselineskip}
\caption{PSNR: 16.33}
\end{subfigure} 
\begin{subfigure}{0.2\textwidth}
\includegraphics[width=\linewidth]{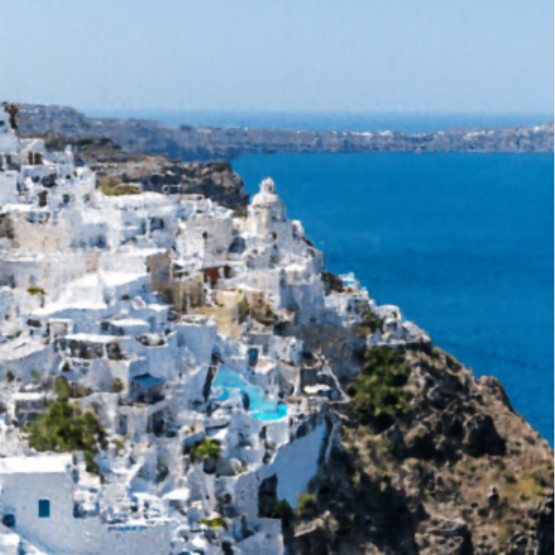} 
\vspace{-1.5\baselineskip}
\caption{PSNR: 22.18}
\end{subfigure}
\begin{subfigure}{0.2\textwidth}
\includegraphics[width=\linewidth]{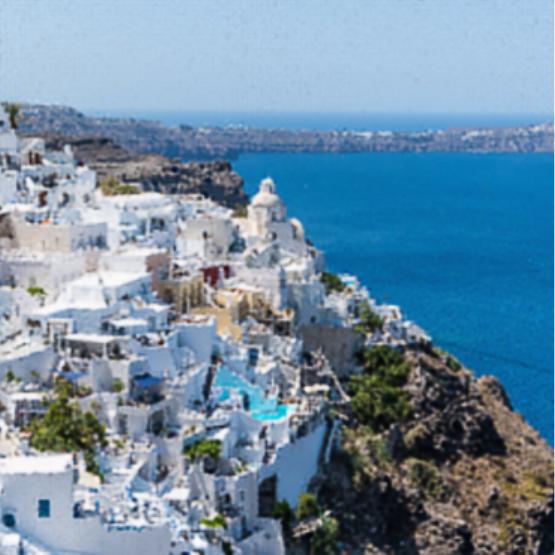}
\vspace{-1.5\baselineskip}
\caption{PSNR: 22.83}
\end{subfigure}
\begin{subfigure}{0.2\textwidth}
\includegraphics[width=\linewidth]{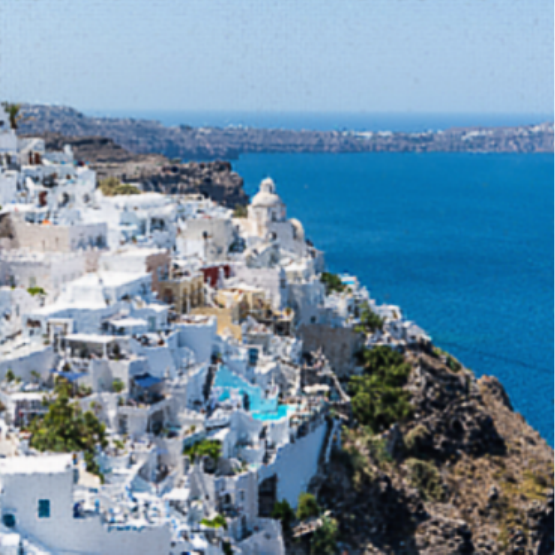}
\vspace{-1.5\baselineskip}
\caption{PSNR: 22.83}
\end{subfigure} 
\vspace{-0.2cm}
\caption{Reconstruction results of a seaside residential area using non-separable coordinates (randomly sampled training points) with different combinations of simple or complex encodings and network depths.}
\label{fig:2d_random_1}
\end{figure}

Here we show 2D image visual results for separable coordinates in~\cref{fig:2d_grid_0,fig:2d_grid_2}, and non-separable coordinates in~\cref{fig:2d_random_1,fig:2d_random_3}.
For simple encoding, five aforementioned encoders were tested with 256 width MLP of 0 and 4 hidden ReLU layers (0 means only a linear layer). 
For complex encoding, the same five encoders were tested with 0 and 1 hidden ReLU MLPs.

\begin{wrapfigure}[24]{r}{0.5\textwidth}
\centering
\vspace{-0.6cm}
\includegraphics[width=0.5\textwidth]{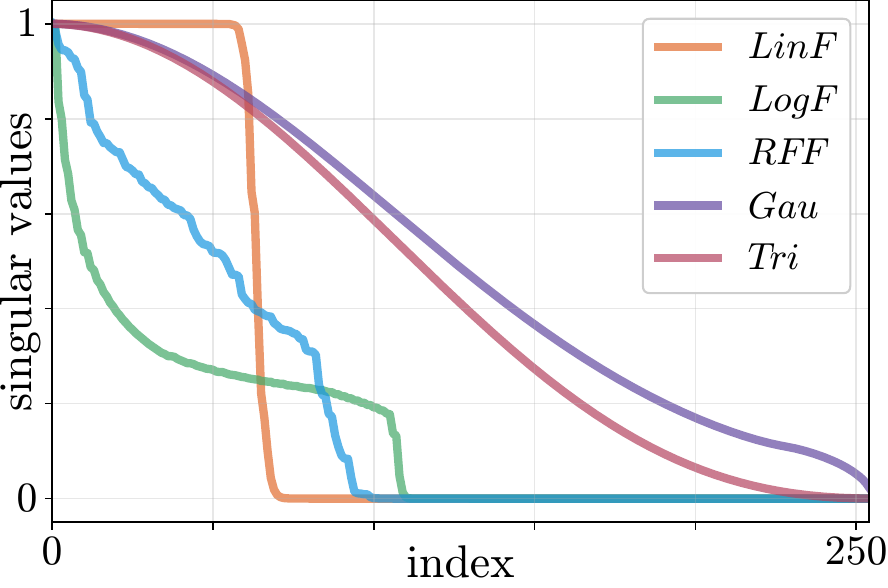}
\caption{\label{fig:svd} The normalized singular values of different 1D embeddings $\Psi(\x)\in\mathbb{R}^{N\times K}$. Here $N{=}K{=}256$ and $\x$ is sampled equally spaced from $0$ to $1$.
Fourier feature-based encodings (LinF, LogF, RFF) tend to have much fewerr non-zero singular values, which results in low rank. While shifted encodings (Tri, Gau) usually have sufficient non-zero singular values, which leads to a high rank. When $\x$ is randomly sampled, the rank deficiency in Fourier feature-based encodings becomes worse.}
\end{wrapfigure}

As shown in column 1 of these figures, when we used simple encodings and the network only had a single linear layer (0 hidden layers), the reconstructed images are of low quality, showing low-resolution color grids (LinF, LogF), cross strip colors (Tri, Gau), or random color blobs (RFF). 
The results clearly support our claim that a linear network can only reconstruct a 2D image signal with at most rank $2$.
When we introduced non-linear layers and increased the hidden layer depth (depth 4, column 2), the reconstruction quality improves, leading to a better PSNR.

On the contrary, even with a single linear layer (depth 0, column 3), our complex encoding methods can achieve comparable results with methods that used a simple encoding combined with deeper non-linear networks.
Note that Fourier feature-based (frequency-based) complex encodings (LinF, LogF, RFF) performed worse than shifted-based complex encodings (Tri, Gau) when there was only one single linear layer due to the deficiency of the embedding rank (shown in~\cref{fig:svd}).
Adding an extra non-linear layer (depth 1, column 4) did not substantially improve the performance of shifted-based complex encodings while adding more details for frequency-based complex encodings.

\clearpage
\bibliographystyle{splncs04}
\bibliography{egbib}

\end{document}